\documentclass{article}
\PassOptionsToPackage{numbers,compress}{natbib}
\usepackage[preprint]{neurips_2026}

\usepackage[utf8]{inputenc}
\usepackage[T1]{fontenc}
\usepackage{amsmath,amssymb,amsthm,mathtools,bm}
\usepackage{booktabs,multirow,makecell,xcolor,array}
\usepackage{soul}
\usepackage{float}
\usepackage{placeins}
\usepackage{graphicx,subcaption,wrapfig,float}
\usepackage{titletoc}
\usepackage{hyperref}
\usepackage[capitalise,noabbrev]{cleveref}  % "Figure 1", "Table 2", "Section 3"
\hypersetup{colorlinks=true,urlcolor=blue,linkcolor=purple,citecolor=violet}

\usepackage[normalem]{ulem}

\theoremstyle{definition}
\theoremstyle{remark}

\newcommand{\R}{\mathbb{R}}
\DeclareRobustCommand{\method}{\texorpdfstring{\textup{\textsc{\textcolor{dialRampA}{D}\textcolor{dialRampB}{I}\textcolor{dialRampC}{A}\textcolor{dialRampD}{L}}}}{DIAL}} % consistent title/body treatment
\DeclareRobustCommand{\methodLite}{\texorpdfstring{\textup{\textsc{\textcolor{dialRampA}{D}\textcolor{dialRampB}{I}\textcolor{dialRampC}{A}\textcolor{dialRampD}{L-IF}}\textcolor{dialRampD}{-Lite}}}{DIAL-IF-Lite}} % plain module (no R+Z read-out) on SteerIF, 4.08M
\DeclareRobustCommand{\methodL}{\texorpdfstring{\textup{\textsc{\textcolor{dialRampA}{D}\textcolor{dialRampB}{I}\textcolor{dialRampC}{A}\textcolor{dialRampD}{L-IF}}}}{DIAL-IF}} % the general instruction-following module (mixed pool)
\newcommand{\ttt}{\texttt} % match BarrierSteer: baselines, datasets, and benchmarks in monospace
\newcommand{\nrmm}[1]{\lVert #1\rVert}            % norm
\theoremstyle{plain}\theoremstyle{remark}   % theory appendix; restores remark style
\crefname{assumption}{Assumption}{Assumptions}\crefname{conjecture}{Conjecture}{Conjectures}   % cleveref has no default for these
\crefname{equation}{Eq.}{Eqs.}\Crefname{equation}{Eq.}{Eqs.}   % "Eq. (1)"; other types spelled out (cleveref noabbrev)
\newcommand{\expY}{\ensuremath{\checkmark}}      % table: yes
\newcommand{\expN}{\ensuremath{\times}}          % table: no
\newcommand{\expPub}{\textsuperscript{\dag}}     % published number, copied from the cited table
\RequirePackage{tikz}
\RequirePackage{pgfplots}
\pgfplotsset{compat=1.18}
\usetikzlibrary{arrows.meta,positioning,calc,fit,backgrounds,decorations.pathreplacing}
\definecolor{figBlue}{HTML}{0072B2}
\definecolor{figOrange}{HTML}{E69F00}
\definecolor{figRed}{HTML}{D55E00}
\definecolor{figGreen}{HTML}{009E73}
\definecolor{figGray}{HTML}{7F7F7F}
\definecolor{dialBlue}{HTML}{0072B2} % Okabe--Ito blue: visible on white and color-blind-safe
\definecolor{dialRampA}{HTML}{7FAAE0}\definecolor{dialRampB}{HTML}{4F88D0}
\definecolor{dialRampC}{HTML}{2369C2}\definecolor{dialRampD}{HTML}{0057B8}
\newcommand{\dialicon}{\tikz[baseline=0.0ex,x=0.80em,y=0.80em,line cap=round]{%
  \draw[line width=0.24em,black!15] (-1,0) arc[start angle=180,end angle=0,radius=1];%
  \foreach \a in {157.5,112.5,67.5,22.5}{\draw[line width=0.03em,black!45] (\a:1.22) -- (\a:1.42);}%
  \draw[line width=0.13em,dialRampA] (0,0) -- (157.5:0.82);%
  \draw[line width=0.13em,dialRampB] (0,0) -- (112.5:0.82);%
  \draw[line width=0.13em,dialRampC] (0,0) -- (67.5:0.82);%
  \draw[line width=0.13em,dialRampD] (0,0) -- (22.5:0.82);%
  \fill[black!85] (0,0) circle[radius=0.12];%
}}
\newcommand{\dialramp}{\texorpdfstring{\textup{\textsc{\textcolor{dialRampA}{D}\textcolor{dialRampB}{I}\textcolor{dialRampC}{A}\textcolor{dialRampD}{L}}}}{DIAL}}

\title{
    \vspace{-3mm}
    \texorpdfstring{\dialicon\hspace{0.3em}}{}
    \dialramp{}: Steerable Language Models with\\
    \textcolor{dialRampA}{D}istilled \textcolor{dialRampB}{I}nstruction-\textcolor{dialRampC}{A}dherence \textcolor{dialRampD}{L}evels
}

\author{
    Thanh Q. Tran\textsuperscript{1}\thanks{Equal contribution. Correspondence to \texttt{thanhtran@u.nus.edu}.}\quad
    John Russell Himawan\textsuperscript{1}\footnotemark[1]\quad
    Arun Verma\textsuperscript{2}\footnotemark[1]\\[2pt] 
    \textbf{Kiwan Wong}\textsuperscript{3}\quad
    \textbf{Wei Xiao}\textsuperscript{4,2,3}\quad
    \textbf{Daniela Rus}\textsuperscript{3,2}\quad
    \textbf{Bryan Kian Hsiang Low}\textsuperscript{1,2}\\[5pt]
    \textsuperscript{1}School of Computing, National University of Singapore, Singapore\\
    \textsuperscript{2}Singapore-MIT Alliance for Research and Technology Centre, Singapore\\
    \textsuperscript{3}CSAIL, Massachusetts Institute of Technology, USA\\
    \textsuperscript{4}School of Electrical and Electronic Engineering, Nanyang Technological University, Singapore \\
}

\begin{document}
    % Keep main-paper paragraph and float spacing natural rather than stretching each page.
    \raggedbottom
    \setlength{\floatsep}{6pt}
    \setlength{\textfloatsep}{6pt}
    \setlength{\intextsep}{6pt}
    \setlength{\abovecaptionskip}{4pt}
    \maketitle
    %!TEX root =  main.tex

\begin{abstract}
    When a user says  %instructs 
    ``\emph{Keep it short}'', how short should the LLM's response be?\\
    More generally, users and applications require different levels of instruction adherence, but prompting does not reliably provide fine-grained control over them. % these levels. 
    We therefore study ``\emph{steerable language models}'' that provide fine-grained numerical control over the degree to which models follow %adherence levels for 
    general instructions specified in text. 
    However, existing approaches do not jointly offer \emph{fine-grained control}, \emph{efficiency}, and \emph{scalability} across diverse instructions. %are not simultaneously fine-grained, efficient, and scalable across general instructions. 
    To address this gap, we introduce \method{} (\textcolor{dialRampA}{\textbf{D}}istilled \textcolor{dialRampB}{\textbf{I}}nstruction-\textcolor{dialRampC}{\textbf{A}}dherence \textcolor{dialRampD}{\textbf{L}}evels) to achieve all three properties. %address these three challenges. 
    For \emph{fine-grained control}, we perform classifier-free guidance (CFG) with %on 
    a frozen LLM to provide training signals for on-policy distillation of instruction adherence at different levels. 
    For \emph{efficiency}, we design a lightweight steering adapter that uses a classifier over the LLM's hidden states to compute %based on the LLM's hidden-state classifier that yields %classifier on hidden states with 
    a closed-form steering edit. 
    The classifier adaptively sets the steering magnitude at each token based on the target adherence level, 
    while accounting for the geometry of the LLM's hidden states to preserve response coherence. 
    For \emph{scalability}, the classifier is text-conditioned, allowing a shared adapter to support new instructions without per-instruction fitting. 
    Training requires only instruction and request text, without labeled responses, external generators, or judges. Evaluation across multiple LLM families shows that \method{} achieves a competitive balance between concept expression and response fluency, with concept scores increasing monotonically across adherence levels. 
    We introduce \ttt{SteerIF}, an instruction-only dataset of $98,625$ concept, persona, verifiable, and descriptive instructions. %instructions including concept, persona, and verifiable instructions. 
    When trained on \ttt{SteerIF}, \method{} provides fine-grained control over personas and, when combined with prompting, improves verifiable instruction following. % when combined with prompting. 
    In \ttt{vLLM}, the LLM steered by \method{} %'s steering adapter 
    retains 96--99\% of the base LLM throughput, with %while its
    per-token inference latency approximately matching that of the base LLM. 
    % We believe \method{} opens new possibilities for LLM adaptation with adjustable adherence levels. %% Let's push it to conclsuion
\end{abstract}

    %!TEX root =  main.tex

\section{Introduction}
\label{sec:intro}

% \begin{figure}[t]
%   \centering
%   \begin{subfigure}{\linewidth}
%     \includegraphics[width=\linewidth]{figs/adherence_length.pdf}
%     \phantomsubcaption\label{fig:length-choice}
%   \end{subfigure}
%   \par\vspace{2pt}
%   \begin{subfigure}{\linewidth}
%     \includegraphics[width=\linewidth]{figs/adherence_roles.pdf}
%     \phantomsubcaption\label{fig:roles}\label{fig:teaser}
%   \end{subfigure}
%   \vspace{-6pt}
%   \caption{\textbf{Steerable Language Models with \method{} and Applications.}
%   (a) \method{} provides fine-grained control over instruction adherence, allowing users to choose among responses
%   generated at different adherence levels. (b) The ability to control adherence levels for general instructions with
%   \method{} facilitates personalization, parental control, and adherence to company specifications.}
%   \label{fig:steerability}
%   \vspace{-6pt}
% \end{figure}
\begin{figure}[t]
    \centering
    \includegraphics[width=\linewidth]{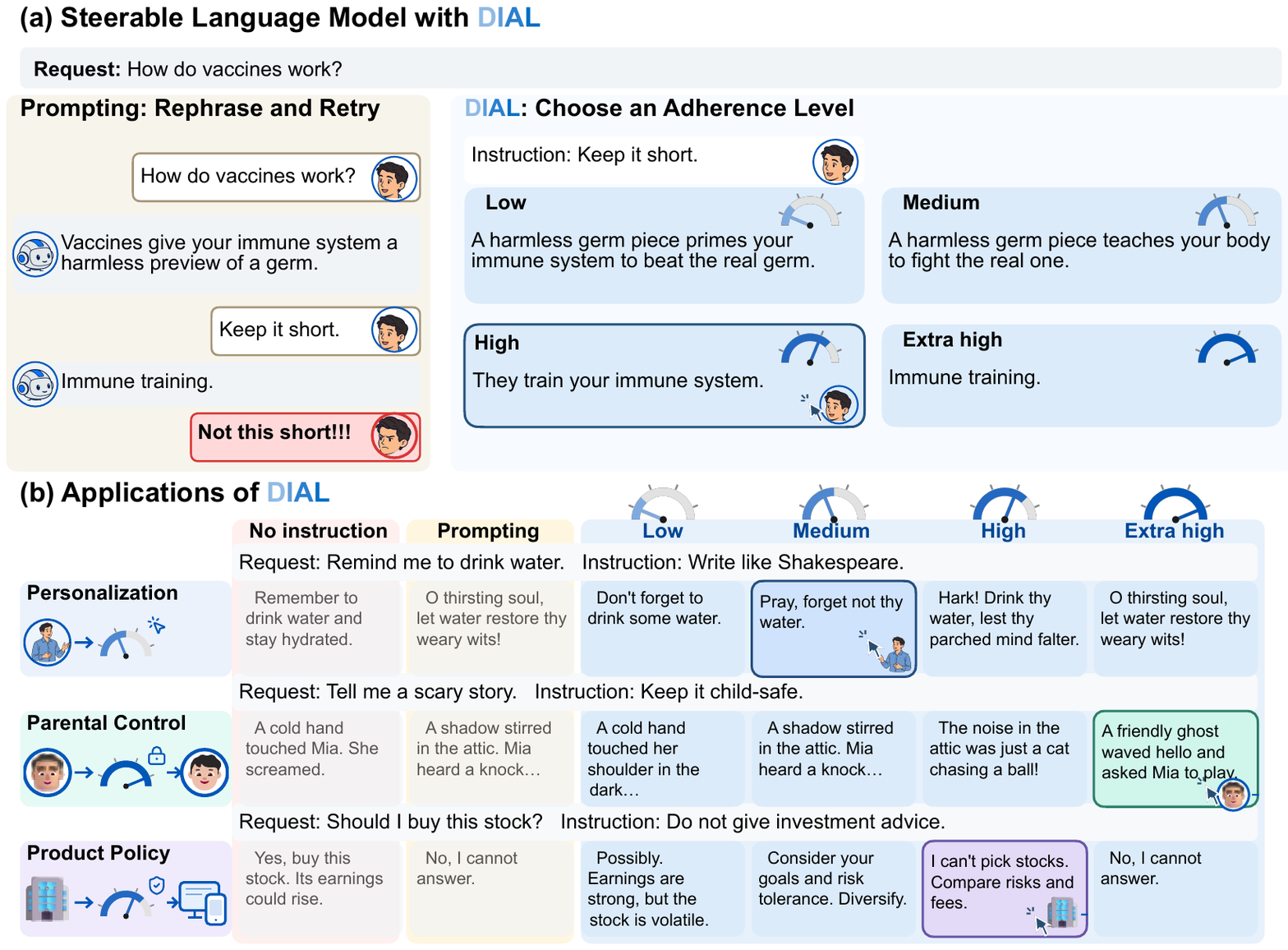}
    \caption{\textbf{Steerable Language Models with \method{} and Applications.}
  (a) \method{} provides fine-grained control over instruction adherence, allowing users to choose among responses
  generated at different adherence levels. (b) Illustrative examples of controlling adherence levels for general instructions with
  \method{} in personalization, parental control, and adherence to company specifications.}
    \label{fig:steerability}
    % \label{fig:length-choice}\label{fig:roles}\label{fig:teaser}
    \vspace{-1.5mm}
\end{figure}

Large language models (LLMs) are deployed across users and applications with different requirements for instruction-following~\citep{ouyang2022training}.
Meeting these requirements involves choosing which instructions an LLM should follow and the appropriate
\emph{level of adherence} for each instruction. Without fine-grained control over adherence levels, users may need to revise an instruction and regenerate responses repeatedly to achieve the desired level of adherence. %obtain the instruction adherence level they want. 
As an illustrative example, \cref{fig:steerability}{\color{purple}(a)} shows a user asking for a shorter answer, but the LLM's response is too short for the user. 
Further prompting is then needed to obtain the desired response length, creating a frustrating user experience. 
Ideally, users could %A desirable capability is to let the user 
view responses generated at different adherence levels and select one %the response 
with the desired length. 
More generally, fine-grained adherence control %control over the adherence level 
for different instructions would support a range of applications (\cref{fig:steerability}{\color{purple}(b)}). For personalization, a user could select the adherence level for writing-style or persona instructions according to their preferences~\citep{salemi2024lamp,sorensen2024pluralistic}. 
For parental control, parents could set the adherence level for child-safety instructions to help safeguard children~\citep{openai2025parentalcontrols}. 
A company could configure the adherence level for product-policy instructions to reflect its %its company 
specifications~\citep{openai2026modelspec,anthropic2026constitution}. Motivated by these applications, we %We therefore 
study ``\emph{steerable language models}''~\citep{dong2023steerlmattributeconditionedsft,wang2024conditionallanguagepolicy} that provide fine-grained numerical control over adherence levels for general text instructions.

Existing approaches for controlling adherence levels in LLMs are not jointly % simultaneously
\emph{fine-grained}, \emph{efficient}, and \emph{scalable} across general instructions. The most straightforward approach is prompting, but it does not reliably provide fine-grained control over the adherence level~\citep{zhou2023instruction,pyatkin2025ifbench}.
In contrast, classifier-free guidance (CFG)~\citep{ho2022cfg} enables fine-grained control by computing two next-token distributions: one conditioned on the instruction and one unconditioned. It then combines their log probabilities to achieve the desired adherence level~\citep{sanchez2023cfg,dong2026systemprompt}. However, this requires two forward passes per generated token, increasing inference cost.
% In contrast, classifier-free guidance (CFG)~\citep{ho2022cfg} provides fine-grained control by computing two next-token distributions, one with and one without the instruction, and combining their log probabilities according to the adherence level~\citep{sanchez2023cfg,dong2026systemprompt}. However, each generated token requires two inference passes, which increases inference cost. 
Activation steering provides a more efficient approach by editing the hidden states of a frozen LLM within a single inference pass~\citep{turner2023steering,rimsky2024caa,li2023iti,arditi2024refusal}. 
Adjustable steering strength offers %The ability to adjust steering strength provides 
a promising way to adapt an LLM at inference time~\citep{ostermann2026steering}. 
However, conventional steering methods are not fine-grained because they are fitted to labeled examples that indicate whether an instruction is followed, without supervision at intermediate adherence levels. 
Increasing steering strength therefore does not reliably increase % instruction 
adherence and can reduce fluency and coherence~\citep{wu2025axbench}. 
Moreover, these methods require fitting for each instruction, limiting scalability. 
Text-conditioned steering methods learn to steer an LLM from instruction text and support instructions unseen during training~\citep{sun2025hypersteer,jin2026flas,shi2026unisteer}. 
However, they are trained on labeled responses without supervision at intermediate adherence levels, and flow-based variants add per-token inference overhead.
% However, the additional computation for text conditioning and steering introduces inference overhead and can reduce efficiency.

\textbf{Contributions.} We introduce \method{} (\textcolor{dialRampA}{\textbf{D}}istilled \textcolor{dialRampB}{\textbf{I}}nstruction-\textcolor{dialRampC}{\textbf{A}}dherence \textcolor{dialRampD}{\textbf{L}}evels) to jointly achieve three key properties: %address the three challenges of 
\emph{fine-grained control}, \emph{efficiency}, and \emph{scalability}. 
For \emph{fine-grained control}, we perform CFG on a frozen LLM to provide training signals for on-policy distillation of instruction adherence at different levels (\cref{sec:method-train}). 
For \emph{efficiency}, we design a lightweight steering adapter~\citep{krasheninnikov2024steering} that uses a classifier over the LLM's hidden states to compute a closed-form steering edit.
% For efficiency, we design a lightweight \emph{steering adapter}~\citep{krasheninnikov2024steering} based on a classifier on hidden states that yields a closed-form edit. 
The classifier adaptively sets the steering magnitude at each token according to the target adherence level, while accounting for the geometry of the LLM's hidden states to preserve the base LLM's response coherence (\cref{sec:method-classifier}). 
For \emph{scalability}, the classifier is text-conditioned, allowing 
a shared steering adapter to support previously unseen instructions without per-instruction fitting.
Furthermore, CFG provides training signals without requiring labeled responses, external generators, or judges, allowing training to extend to new instruction types using only instruction and request text.
% Furthermore, CFG provides training signals without labelled responses, external generators, or judges, so training can extend to new instruction types using only instruction text.

Empirical evaluation across multiple LLM families demonstrates the advantages of \method{}. When steering LLM generation toward concepts such as ``Shakespeare,'' \method{} achieves a better balance between concept expression and response fluency than existing methods. 
Importantly, \method{} provides fine-grained control over concept expression, with the concept score increasing monotonically across the range of adherence levels. 
To evaluate %extend the evaluation to 
general instruction following, we construct \ttt{SteerIF}, an instruction-only dataset comprising $98,625$ instructions, including concept, % instructions, 
persona, %instructions,
and verifiable instructions.
We train a shared steering adapter for each base LLM and evaluate it on persona instructions, such as ``Be humorous,'' and verifiable instructions, such as ``Respond in JSON.'' 
Persona scores increase with the adherence level across LLM families, and combining steering with prompting improves verifiable instruction following. 
In terms of efficiency, \method{}'s $20.6$M-parameter steering adapter retains 96--99\% of base-LLM throughput in \ttt{vLLM}~\citep{kwon2023efficient} on Gemma-2-9B~\citep{riviere2024gemma2}, while per-token inference latency approximately matches that of the base LLM.

    \section{Related Work}
\label{sec:related}
We review three lines of work on steering LLMs. Guided inference is fine-grained but incurs substantial inference overhead; %not efficient, 
activation steering is efficient, but neither fine-grained nor scalable; and text-conditioned control generalizes to unseen instructions, but lacks fine-grained control and requires labeled responses, limiting scalability. See \cref{app:related} for details.

% In this section, we review existing work on controlling a language model's generation. Guided decoding sets how strongly
% an instruction is followed at a fine-grained level, but it is slow. Activation steering can adapt a model at different
% strengths in one pass, but it is fitted to one concept at a time and its strength has no defined meaning. Text-conditioned
% methods remove the per-concept fit by conditioning on text, but they learn from labelled data that does not scale to
% general instructions, offer no defined levels, and are often slow. \Cref{app:related} further extends this review.

\textbf{Guided Inference.}
In diffusion models~\citep{sohl2015deep,ho2020ddpm,song2021sde}, guided inference controls how strongly a condition, such as a text prompt, influences the generated image. Given the prompt ``a cat in the style of Van Gogh,'' a low guidance level gives an image that only hints at the style, and a high level gives one that is unmistakably Van Gogh. 
Classifier guidance pushes each sampling step along the gradient of a separately trained classifier~\citep{dhariwal2021diffusion}, while classifier-free guidance (CFG) removes this classifier by using the diffusion model itself as an implicit classifier~\citep{ho2022cfg}. In LLMs, CFG combines the next-token predictions with and without the instruction~\citep{sanchez2023cfg,dong2026systemprompt}. 
A low guidance (i.e., adherence) level weakens the instruction relative to prompting, and a high level strengthens it. Related contrastive inference methods for LLMs improve faithfulness, instruction following, and open-ended generation~\citep{shi2024cad,kim2024instructive,li2023contrastive,liu2021dexperts}. 
Because every guidance level defines a next-token distribution for any instruction, guided inference is \emph{fine-grained}. However, CFG is not \emph{efficient}, because every generated token needs two forward passes. %inference passes. 
This motivates guidance distillation in diffusion models, which trains a single-pass student  %that runs in a single pass and is 
conditioned on the guidance level, thereby recovering CFG at multiple levels~\citep{meng2023guided}.
% This motivates guidance distillation in diffusion models, which trains a student that runs in a single pass but is conditioned on the guidance level, so that it recovers CFG at multiple levels~\citep{meng2023guided}.
In autoregressive models, CFG has been distilled into the weights only at a single
level~\citep{cideron2024diversity}. 
% \del{Distilling CFG at multiple levels is difficult for LLMs because,
% unlike a diffusion model, an LLM offers no natural input through which to condition the student on a guidance level.}
% \del{Unlike a diffusion model, an LLM has no dedicated input for the guidance level, so a student must receive the level through its prompt or through additional parameters. \method{} conditions a lightweight steering adapter on the level, leaving the LLM's weights and prompt unchanged.}
Unlike a diffusion model, %whose denoiser already takes continuous conditioning inputs (e.g., the noise level) to which a guidance-level embedding can be added, 
an LLM takes only tokens and has no natural input for the guidance level. %an LLM has no dedicated input for the guidance level. 
\method{} serves as a lightweight steering adapter that conditions the LLM on the adherence level, approximating CFG at multiple levels in a single pass without changing the LLM's weights or prompt.

\textbf{Activation Steering.}
Activation steering offers a promising way to adapt an LLM cheaply at multiple levels~\citep{ostermann2026steering}.
It edits the hidden states of a frozen LLM in a single forward pass while leaving the weights unchanged, allowing the steering level to be set for each input at inference time. A common intervention is to use a steering vector obtained from contrastive hidden states~\citep{turner2023steering,rimsky2024caa}, which has been used to control truthfulness~\citep{li2023iti}, refusal~\citep{arditi2024refusal}, and instruction following~\citep{stolfo2025instruction}. 
Classifier-guided steering methods %instead 
project the hidden state onto the half-space of a trained classifier in closed form~\citep{cheng2024liseco,xu2024scav,zhang2025gcav}, or onto the constraint set given by a first-order condition on %first-order condition of 
a learned nonlinear safety score~\citep{tran2026barriersteer}. Geometry-aware methods rotate the state rather than translate it, preserving the hidden state's norm %so that the hidden state's norm is preserved~
\citep{pham2024hpr,vu2025angular,you2026spherical}. 
These interventions are \emph{efficient}, and their steering level can be varied continuously. However, they are fitted to labeled examples that either follow the instruction or do not, with no training at %so they are never trained for 
intermediate levels of adherence, and are therefore not \emph{fine-grained}. Increasing the steering level therefore does not reliably yield a corresponding increase in adherence, and high levels often reduce the output fluency and coherence %of the output~
\citep{wu2025axbench}.
Moreover, each intervention is fitted to a single instruction, such as one concept or behavior, so each new instruction requires a new intervention.

\textbf{Text-Conditioned Control.}
To avoid refitting for every instruction, text-conditioned methods train a single network that maps a text description to an intervention. 
\ttt{HyperSteer} maps a concept description to a steering vector~\citep{sun2025hypersteer}, \ttt{FLAS} integrates a description-conditioned flow~\citep{jin2026flas}, and \ttt{UniSteer} trains a text-conditioned flow-matching network that also covers personas and multi-instruction following %following multiple instructions~
\citep{shi2026unisteer}. 
Their training, however, suffers from the same two problems as activation steering. 
They are fitted to labeled examples, here concept-aligned completions written by an external LLM or labeled datasets for each behavior, so \emph{scaling} to a new kind of instruction requires collecting new data. 
These examples also either follow the instruction or do not, so the methods are never trained for intermediate levels of adherence and are not \emph{fine-grained}. 
\ttt{Persona Dosing} partly addresses this for personas by fine-tuning \ttt{FLAS} on persona-labeled data and calibrating its flow time against measured persona expression~\citep{jin2026personadosing}, which requires measuring each persona, whereas \method{} learns its levels during training and needs no such measurement for a new instruction. 
Moreover, flow-based methods integrate a flow over several steps per token, which adds latency and limits practicality.

To address these limitations, \method{} distills the frozen LLM's own 
CFG distributions at multiple adherence levels %at multiple levels 
into a single steering adapter. 
Since the teacher requires only instruction and request text, training can scale to new types of instructions without collecting labeled responses. 
% so training scales to new kinds of instruction without collecting responses. 
The adapter learns to approximate the CFG-guided distribution at each target level, enabling fine-grained control while keeping %Each level is trained toward the CFG distribution at that level, and the adapter keeps the 
inference cost close to that of the base LLM. 
\method{} thus provides a fine-grained, efficient, and scalable way to condition an LLM on
general instructions and adherence levels.
    \section{Method: \method{} (\textcolor{dialRampA}{\textbf{D}}istilled \textcolor{dialRampB}{\textbf{I}}nstruction-\textcolor{dialRampC}{\textbf{A}}dherence \textcolor{dialRampD}{\textbf{L}}evels)}
\label{sec:method}

\method{} uses a text-conditioned steering adapter to control instruction adherence in a frozen LLM at different levels. 
We first define adherence levels via %through
CFG (\cref{sec:method-problem}), and then describe how on-policy distillation trains the adapter at different adherence levels (\cref{sec:method-train}). 
Finally, we introduce the adapter's lightweight classifier-based architecture and closed-form steering edit (\cref{sec:method-classifier}).

\subsection{Instruction-Adherence Levels via Classifier-Free Guidance}
\label{sec:method-problem}
Let $x$ denote a request (e.g., ``How do vaccines work?''), $c$ an instruction (e.g., ``Keep it short''), and $\alpha>0$ an adherence level. 
Given a response prefix $y_{<t}$, the frozen LLM yields two next-token distributions: $\pi_{\mathrm{base}}$, conditioned on the request alone, and $\pi_{\mathrm{prompted}}$, conditioned on both the request and the instruction. 
For brevity, we write $\pi_{\cdot}(\cdot\mid y_{<t})$ and leave the conditioning on $x$ and $c$ implicit.
Following CFG~\citep{sanchez2023cfg,dong2026systemprompt}, we define the teacher distribution at level $\alpha$ as
\begin{equation}
  \pi^T_\alpha(\cdot\mid y_{<t})\ \propto\
  \pi_{\mathrm{base}}(\cdot\mid y_{<t})^{1-\alpha}\,\pi_{\mathrm{prompted}}(\cdot\mid y_{<t})^{\alpha}.
  \label{eq:guided}
\end{equation}
For $\alpha=1$, the teacher recovers $\pi_{\mathrm{prompted}}$. For $0<\alpha<1$, it interpolates toward $\pi_{\mathrm{base}}$, weakening the instruction's influence. 
For $\alpha>1$, it extrapolates beyond $\pi_{\mathrm{prompted}}$, strengthening the instruction's influence. 
Thus, larger $\alpha$ %naturally 
corresponds to stronger instruction adherence, and 
CFG provides a next-token distribution at each adherence level for any instruction text, without labeled responses, external generators, or judges.
The prompt format for each instruction type is given in \cref{app:optimization}.

\begin{figure}[t]
  \centering
  \includegraphics[width=\linewidth,trim=0 48bp 0 36bp,clip]{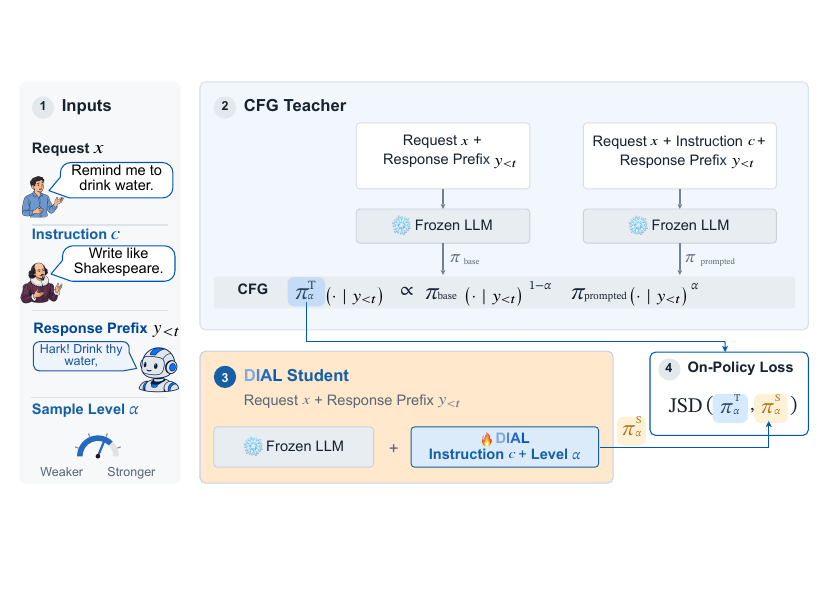}
  \caption{\textbf{\method{}'s Training Pipeline.} 
  Given a request $x$ and an instruction $c$, we sample adherence levels. CFG defines the teacher distributions on prefixes sampled from the student (on-policy), and the frozen LLM steered by \method{} defines the student distributions. \method{}'s steering adapter is trained with a Jensen--Shannon divergence (JSD) loss across adherence levels.
  % Given a Request $x$ and Instruction $c$, we sample adherence levels  $\alpha$. CFG defines teacher on-policy distributions, and the frozen LLM steered by \method{} defines student distributions. \method{}'s steering adapter is trained using JSD loss across adherence levels.
  }
  \label{fig:method}
  \vspace{-6pt}
\end{figure}

\subsection{On-Policy Self-Distillation across Adherence Levels}
\label{sec:method-train}

\textbf{Training across Adherence Levels.}
We train \method{}'s steering adapter to match the CFG teacher distributions at different adherence levels. 
The student $\pi^S_\alpha$ is the frozen LLM with generation steered by \method{} at layer $\ell$. 
The request $x$ is passed to the frozen LLM, while the instruction $c$ and adherence level $\alpha$ are passed to \method{}'s steering adapter. 
At each prompt and generated position, the adapter replaces the residual hidden state $h_t\in\R^{d_{\mathrm{LLM}}}$ with an edited state $h'_t$.
As in %Similar to 
guidance distillation in diffusion models~\citep{meng2023guided}, a single adapter is trained to match the CFG distributions across levels.

For each request--instruction pair, we sample $\alpha\sim\operatorname{LogUniform}[0.5,2]$, a distribution that is symmetric %which is symmetric 
in $\log\alpha$ about $\alpha=1$, where the CFG teacher recovers the prompted distribution. 
This range covers weaker and stronger instruction adherence than prompting. For $0.5\le\alpha<1$, the teacher weakens the instruction's influence relative to prompting, providing fine-grained intermediate training signals between unprompted and prompted generation. 
Beyond $\alpha=2$, the CFG teacher's coherence degrades (\cref{app:optimization}), consistent with prior work~\citep{sanchez2023cfg}.
% We found that {$\alpha>2$ begins to degrade the CFG teacher's coherence}, consistent with prior work~\citep{sanchez2023cfg}.
We empirically found that sampling 24 adherence levels %$\alpha$  
per instruction--request pair worked well for % gives good results in 
training \method{} across LLM families. 
We also verify that sampling $\alpha$ is important for fine-grained instruction adherence levels, and that training on more diverse instructions works better than training longer~(see \cref{app:abl-training}).

\textbf{On-Policy Self-Distillation.}
The student's response distribution varies with adherence level and can differ from the prompted LLM's.
We therefore use on-policy self-distillation~\citep{agarwal2024gkd,gu2024minillm}, sampling student responses and computing the CFG teacher distribution at each response prefix.
We minimize the Jensen--Shannon divergence (JSD) between teacher and student distributions per response token:
\begin{equation}
  \mathcal L
  =\mathbb E_{(x,c)\sim\mathcal D,\, \alpha,\,y\sim\pi^S_\alpha}\!\left[\frac{1}{|y|}\sum_{t}
  \operatorname{JSD}\!\left(\pi^T_\alpha(\cdot\mid y_{<t}),\,\pi^S_\alpha(\cdot\mid y_{<t})\right)\right],
  \label{eq:loss}
\end{equation}
where $\mathcal D$ is the distribution of training request--instruction pairs. 
Both distributions are restricted to the union of the base, prompted, and student top-64 tokens.
We also add a %small 
regularization term to encourage stable training and generalization by penalizing large relative edits (\cref{app:optimization}).
Our loss ablation in \cref{app:abl-training} suggests that JSD performs best, with forward KL performing substantially worse and reverse KL
performing slightly worse.
We refer to \cref{app:optimization,app:pseudocode} for training details and pseudocode of the full on-policy training loop across adherence levels.

\subsection{Lightweight and Adaptive Classifier-Based Steering Adapter}
\label{sec:method-arch}\label{sec:method-classifier}\label{sec:method-geometry}\label{sec:method-props}

\textbf{Instruction and Memory Conditioning.}
We design the adapter to condition each steering edit on the instruction and the response generated so far. 
We encode the instruction through the frozen LLM and cache its layer-$\ell$ hidden states $\mathcal I$. 
At each position, the current hidden state $h_t$ attends to $\mathcal I$ and a memory of the 16 most recent edited states, $\mathcal M_t=\{h'_{t-16},\dots,h'_{t-1}\}$:
\begin{equation}
  z_t=\operatorname{Attn}_{\ell}\!\left(\mathrm{query}=h_t;[\mathcal I;\mathcal M_t]\right),\qquad
  f_t=\operatorname{GELU}\!\left(\operatorname{LayerNorm}(\operatorname{Linear}(z_t))\right).
  \label{eq:attn}
\end{equation}
Here, $\operatorname{Attn}_\ell$ reuses the frozen layer's normalization and query, key, and value projections.
The current state is the query, while the instruction and memory states supply the keys and values. 
Since these states come from the same LLM layer, the adapter does not require a separate text encoder for the instructions. 
Attention combines instruction and memory states, followed by a linear map, LayerNorm, and GELU to produce the 512-dimensional feature $f_t$. 
In our experiments, a memory window of the 16 most recent states performed well while keeping attention cost low.
% We empirically found that a memory window of the 16 most recent states performs well while keeping attention cost low.

\textbf{Adaptive Classifier-Based Steering.}
From $f_t$, the adapter predicts a unit direction $u_t\in\R^{d_{\mathrm{LLM}}}$, a threshold $\tau_t(\alpha)\in\R$, and a norm weight $\lambda_t(\alpha)\in[0,1]$. 
The direction and threshold define a linear classifier with half-space $\{h:\langle u_t,h\rangle\ge\tau_t(\alpha)\}$. Unlike classifiers fitted separately for each instruction, \method{}'s classifier is generated from instruction and memory information at each token. 
Rather than linearly scaling a fixed steering edit, \method{} learns the threshold and norm weight as functions of $\alpha$. 
On-policy distillation provides rich supervision from the CFG teacher for learning steering magnitude and geometry at different adherence levels.

The classifier $(u_t,\tau_t(\alpha))$ determines whether and how much to steer the current hidden state. 
If $h_t$ lies outside the half-space, we project it onto the boundary with a minimum-norm step, following classifier-based steering~\citep{cheng2024liseco,xu2024scav,zhang2025gcav}. 
We then adjust the projected state's norm using the norm weight $\lambda_t(\alpha)$ while retaining its direction. 
Both operations have closed-form expressions:
\begin{equation}
  h^P_t=h_t+\big[\tau_t(\alpha)-\langle u_t,h_t\rangle\big]_+\,u_t,\qquad
  h'_t=\Bigl[\lambda_t(\alpha)+\bigl(1-\lambda_t(\alpha)\bigr)\frac{\nrmm{h_t}}{\nrmm{h^P_t}}\Bigr]\,h^P_t,
  \label{eq:projrot}
\end{equation}
where $[a]_+=\max(a,0)$. 
The gap $[\tau_t(\alpha)-\langle u_t,h_t\rangle]_+$ sets the projection magnitude.
Thus, the edit depends on the current state as well as the generated classifier, rather than applying a fixed steering magnitude to every token. 
If the state already lies inside the half-space, the edit leaves $h_t$ unchanged.

The norm weight lets the adapter interpolate %choose 
between two steering geometries, which we visualize in \cref{app:projrot}. 
At $\lambda_t=1$, the edited state is the minimum-norm projection $h'_t=h^P_t$~\citep{cheng2024liseco,xu2024scav,zhang2025gcav}. 
At $\lambda_t=0$, it preserves the original norm, $h'_t=\nrmm{h_t}\,h^P_t/\nrmm{h^P_t}$, giving a rotation in the span of $h_t$ and $u_t$~\citep{pham2024hpr,vu2025angular,you2026spherical}.
Intermediate values preserve the projected direction with norm $(1-\lambda_t)\nrmm{h_t}+\lambda_t\nrmm{h^P_t}$. 
Learning this weight allows the geometry to adapt at each token. 
We found that adapters trained for different LLM families learn different norm weights and edit magnitudes, %thresholds and norm weights, 
suggesting that appropriate steering magnitudes and geometries depend on the base LLM (see \cref{app:abl-geometry}). 
\method{}'s steering adapter has 20.6M trainable parameters on Gemma-2-9B (see \cref{app:arch}).

\textbf{Efficient Inference.}
During inference, instruction and memory keys and values can be cached, so each token requires only a
single-layer attention computation and a closed-form steering edit.

    \section{Experiments}
\label{sec:experiments}
% Keep float-to-text gaps fixed instead of stretching them to fill the page.
\setlength{\textfloatsep}{10pt}

We evaluate \method{} when the instruction is to express a concept, adopt a persona, or satisfy a verifiable requirement.
We also compare its training and inference efficiency against baselines.

\textbf{Base LLMs.}
We train \method{}'s adapters for instruction-tuned LLMs: Gemma-2-9B~\citep{riviere2024gemma2}, Qwen2.5-7B~\citep{yang2025qwen25}, Llama-3.1-8B~\citep{grattafiori2024llama3}, Phi-4-14B~\citep{abdin2024phi4}, LFM2.5-1.2B~\citep{amini2025lfm2}, and Ministral-8B~\citep{mistral2024ministraux}.
These cover six LLM families with 1.2--14B parameters, including Transformer and hybrid architectures.

\textbf{Training Setup.}
\label{sec:steerif}
We train \method{} on $14,880$ concept descriptions from \ttt{Concept16k}~\citep{wu2025axbench}.
We also construct \ttt{SteerIF}, an instruction-only dataset of $98,625$ instructions covering concepts, personas, and verifiable and descriptive requirements, given as single or composite instructions.
% texts covering concept, persona, and single and composite instructions, including verifiable and descriptive instructions.
We train one shared adapter per LLM on \ttt{SteerIF}, denoted \methodL{}.
For both datasets, we remove exact and near matches to held-out evaluation instructions to assess generalization.
We select a steering layer for each LLM and train its adapter at that layer.
See \cref{app:training-data} for more details on dataset construction and to \cref{app:optimization} for the steered layer chosen for each LLM.

\subsection{Controlling Concept Expression Levels}
\label{sec:exp-axbench}
\label{sec:exp-training}\label{sec:exp-level}\label{sec:exp-geometry}
\begin{wraptable}[26]{R}{0.42\linewidth}
  \vspace{-9pt}
  \renewcommand{\expPub}{\textsuperscript{+}}
  \captionsetup{justification=raggedright,singlelinecheck=false,font=small}
  \centering\footnotesize
  \setlength{\tabcolsep}{2pt}\renewcommand{\arraystretch}{0.95}
  \caption{\textbf{Concept Steering on AxBench.} Overall scores on Gemma-2-9B with \ttt{gpt-4o-mini} as judge.
  \expPub\ denotes results adapted from~\citep{wu2025axbench,wu2025reps,heyman2026psr,sun2025hypersteer}.}
  \label{tab:main-axb-9b}\label{tab:flas-axb-reference}
  % \vspace{1.5pt}
  \begin{tabular}{@{}lcc@{}}
    \toprule
    Method & Held-In & Held-Out \\
    \midrule
    Prompting & 1.049 & 1.060 \\
    CFG ($\alpha{=}2$) & -- & 1.115 \\
    \midrule
    \multicolumn{3}{@{}l@{}}{\textit{One Adapter per Concept}} \\
    \ttt{LoReFT}~\citep{wu2025axbench} & 0.777\expPub & -- \\
    \ttt{LoRA}~\citep{wu2025axbench} & 0.602\expPub & -- \\
    S-\ttt{PSR}~\citep{heyman2026psr} & 0.900\expPub & -- \\
    SV + \ttt{RePS}~\citep{wu2025reps} & 0.892\expPub & -- \\
    \ttt{ReFT-r1}~\citep{wu2025axbench} & 0.630\expPub & -- \\
    \ttt{DiffMean}~\citep{wu2025axbench} & 0.322\expPub & -- \\
    \ttt{SAE}~\citep{wu2025axbench} & 0.191\expPub & -- \\
    \midrule
    \multicolumn{3}{@{}l@{}}{\textit{One Adapter for All Concepts}} \\
    \ttt{HyperSteer}~\citep{sun2025hypersteer} & 1.091\expPub & 0.934\expPub \\
    \ttt{FLAS}~\citep{jin2026flas} ($T{=}2$) & 0.997 & 0.987 \\
    \midrule
    \multicolumn{3}{@{}l@{}}{\textnormal{\method{}}\textit{ trained on \ttt{Concept16k}}} \\
    \method{} ($\alpha{=}1$) & 0.757 & 0.765 \\
    \method{} ($\alpha{=}2$) & 1.059 & 1.036 \\
    Prompt + \method{} ($\alpha{=}0.5$) & 1.108 & 1.111 \\
    Prompt + \method{} ($\alpha{=}1$) & \textbf{1.198} & \textbf{1.218} \\
    \midrule
    \multicolumn{3}{@{}l@{}}{\textnormal{\method{}}\textit{ trained on \ttt{SteerIF}}} \\
    \methodL{} ($\alpha{=}1$) & 0.704 & 0.672 \\
    \methodL{} ($\alpha{=}2$) & 1.039 & 0.993 \\
    Prompt + \methodL{} ($\alpha{=}0.5$) & 1.118 & 1.130 \\
    Prompt + \methodL{} ($\alpha{=}1$) & 1.196 & 1.188 \\
    \bottomrule
  \end{tabular}
\end{wraptable}

\textbf{Training and Baselines.}
We train \method{} for one epoch on 14{,}880 concept descriptions from \ttt{Concept16k}~\citep{wu2025axbench}.
We use \ttt{FLAS}'s released Gemma-2-9B checkpoint, trained on \ttt{Concept16k} with LLM-generated labeled responses~\citep{jin2026flas}, whereas \method{} uses only concept descriptions.
We compare with prompting, per-concept methods (\ttt{LoReFT}, \ttt{LoRA}, \ttt{ReFT-r1}, \ttt{DiffMean}, and \ttt{SAE}~\citep{wu2025axbench}, SV + \ttt{RePS}~\citep{wu2025reps}, and S-\ttt{PSR}~\citep{heyman2026psr}), and text-conditioned methods (\ttt{HyperSteer} and \ttt{FLAS}).
We also include CFG~\citep{sanchez2023cfg,dong2026systemprompt} for fine-grained control over concept expression, %
% \del{, and with prompting w/ level, which states the adherence level in the prompt as a number out of 2}
and prompting w/ level~\citep{zhou2024smooth}, which uses the same prompt as prompting but also asks the LLM to follow it at adherence level $\alpha$, on a scale from 0 (ignore) to 2 (follow strongly).
See \cref{app:results-concept} for implementation details.

\begin{figure}[t]
  \centering
  \includegraphics[width=\linewidth,trim=0 10bp 0 0,clip]{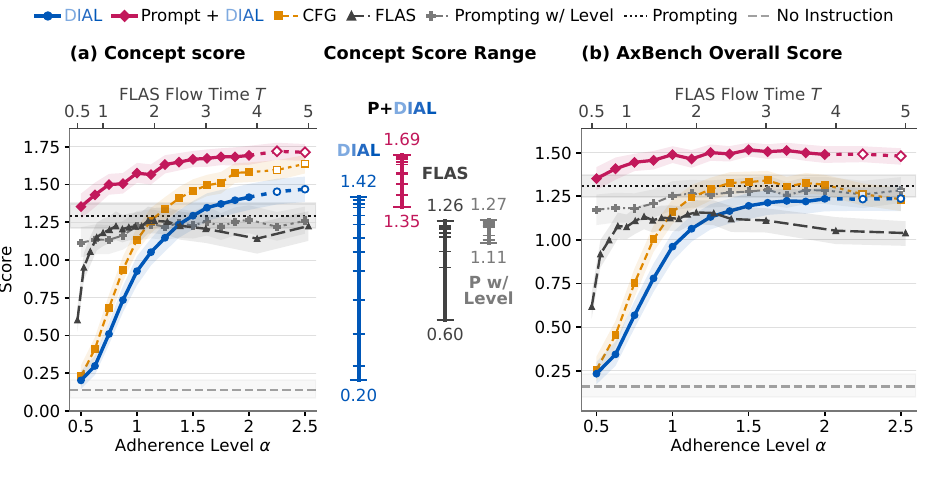}
  \caption{\textbf{Fine-Grained Concept Expression Levels.} (a) \method{} smoothly increases concept expression with increasing $\alpha$ (\cref{sec:method-problem}), covering a wide range of intermediate levels and surpassing prompting at $\alpha=2$.
  \textbf{P+\method{}} further raises expression.
  (b) \method{} approaches prompting's overall score, enabling adjustable adherence levels while maintaining response quality.
  Results reported are on 100 held-out concepts judged by median of \ttt{Phi-4-14B}, \ttt{Qwen3.8-27B}, and \ttt{gpt-4o-mini}.}
  \label{fig:level-dial}
\end{figure}

\textbf{Evaluation Protocol.}
We evaluate on \ttt{AxBench}'s \ttt{Concept500} (Held-In)
% \new{for \ttt{FLAS}, which trains on 480 of its 500 concepts}
and \ttt{FLAS}'s 100 held-out concepts (Held-Out).
\ttt{HyperSteer}'s published Held-Out score uses its own held-out concepts.
Neither \ttt{Concept500} nor the held-out concepts appear in \method{}'s training pool.
Following the \ttt{AxBench} protocol, we limit responses to 128 tokens and steer \method{} at layer 20 for Gemma-2-9B.
Per-concept methods select a steering level for each concept, while \method{} uses a fixed adherence level $\alpha$ across concepts.
\ttt{FLAS} controls concept expression through its flow time $T$, with larger $T$ applying the learned flow for longer~\citep{jin2026flas}.
We use $T=2$ as in \ttt{FLAS}'s reported evaluation and vary $T$ in \cref{fig:level-dial}.
Protocols, baselines, and further results are in \cref{app:results-concept}.

\textbf{Evaluation Metrics.}
Each response receives three scores from 0 to 2: concept expression, request following %following the original request
(\ttt{AxBench}'s instruction score), and fluency.
\ttt{AxBench} defines the overall score as the harmonic mean of the three scores.
We use \ttt{gpt-4o-mini}~\citep{openai2024gpt4o} as a judge to compare with prior work (\cref{tab:main-axb-9b}) and the median of three judges for the curves in \cref{fig:level-dial}.

\textbf{Concept Steering Performance.}
\method{} achieves overall scores of 1.059 and 1.036 on Held-In and Held-Out concepts, compared with 0.997 and 0.987 for \ttt{FLAS} (\cref{tab:main-axb-9b}).
Combining \method{} with prompting raises the scores to 1.198 and 1.218, compared with prompting's 1.049 and 1.060.
\methodL{}, trained on \ttt{SteerIF}, steers persona and verifiable instructions and also performs well on \ttt{AxBench}, scoring 1.039 and 0.993.

\textbf{Fine-Grained Control.}
\method{}'s concept score increases smoothly with the adherence level (\cref{fig:level-dial}).
From $\alpha=0.5$ to 2, it rises from 0.20 to 1.42, covering a range of 1.21 points compared with 0.66 for \ttt{FLAS} over flow times 0.5--2.
% \del{Stating the adherence level in the prompt does not provide this control: prompting w/ level stays between 1.11 and 1.27 from $\alpha{=}0.5$ to 2, below prompting's 1.29 at every level.}
We also find that prompting w/ level does not provide fine-grained control, consistent with prior work~\citep{zhou2024smooth}.
At $\alpha=1.5$, \method{} matches prompting's concept score of 1.29 and surpasses it at $\alpha=2$.
Combining prompting with \method{} further raises concept expression, with scores from 1.35 to 1.69.
The \method{} adapter thus covers intermediate expression levels below and above prompting, mirroring CFG.
Although trained only on single-turn responses, \method{} also shows potential for multi-turn steering.
Since an instruction given once in the context tends to lose its effect as the conversation grows~\citep{li2024instability}, whereas \method{} steers the hidden state at every token during generation.
\cref{app:abl-multiturn} demonstrates this advantage in multi-turn settings.
\par
% Add clearance only if Table 1 is still wrapping this paragraph.
\ifnum\value{WF@wrappedlines}>0
  \vspace{12pt}
\fi
\WFclear

\textbf{Inference Efficiency.}
We measure inference efficiency on Gemma-2-9B on an H200 at batch sizes of 1, 32, 128, and 256, generating 128 tokens per response.
\Cref{tab:cost} demonstrates that \method{} retains most of the frozen LLM's inference efficiency.
In \ttt{vLLM}~\citep{kwon2023efficient}, it retains 96--99\% of the base LLM's throughput, compared with 41--86\% for CFG and 84--86\% for \ttt{FLAS}.
Its per-token inference latency approximately matches the base LLM's.
Processing the instruction before generation increases time to first token and contributes to the remaining throughput overhead.
Similar trends hold for Hugging~Face.
We refer to \cref{app:results-cost} for absolute values and time to first token.

\begin{table}[!ht]
  % \vspace{-1mm}
  \centering\footnotesize
  \setlength{\tabcolsep}{4pt}
  \caption{\textbf{Inference Efficiency of \method{}.} Latency and throughput relative to the base LLM in \ttt{vLLM} and Hugging~Face across batch sizes ($\mathrm{B}$).
  CFG runs out of memory in Hugging~Face at $\mathrm{B}=256$.}
  \label{tab:cost}\label{tab:cost-infer-main}
  \vspace{2pt}
  \begin{tabular}{@{}llcccccccc@{}}
    \toprule
    & & \multicolumn{4}{c}{\ttt{vLLM}} & \multicolumn{4}{c}{Hugging~Face} \\
    \cmidrule(lr){3-6}\cmidrule(l){7-10}
    & Method & B=1 & B=32 & B=128 & B=256 & B=1 & B=32 & B=128 & B=256 \\
    \midrule
    \multirow{3}{*}{Throughput $\uparrow$}
      & CFG & 0.86 & 0.66 & 0.47 & 0.41 & 0.89 & 0.73 & 0.51 & \textit{OOM} \\
      & \ttt{FLAS} & 0.85 & 0.84 & 0.85 & 0.86 & 0.77 & 0.79 & 0.91 & 0.91 \\
      & \method{} & \textbf{0.99} & \textbf{0.96} & \textbf{0.98} & \textbf{0.97} & \textbf{0.95} & \textbf{0.95} & \textbf{0.99} & \textbf{0.99} \\
    \midrule
    \multirow{3}{*}{Per-token latency $\downarrow$}
      & CFG & 1.15 & 1.45 & 2.06 & 2.40 & 1.13 & 1.30 & 1.90 & \textit{OOM} \\
      & \ttt{FLAS} & 1.17 & 1.19 & 1.15 & 1.12 & 1.30 & 1.27 & 1.10 & 1.09 \\
      & \method{} & \textbf{1.00} & \textbf{1.00} & \textbf{0.99} & \textbf{1.00} & \textbf{1.04} & \textbf{1.03} & \textbf{1.00} & \textbf{1.00} \\
    \bottomrule
  \end{tabular}
  \vspace{4pt}
\end{table}

\begin{wraptable}{r}{0.35\linewidth}
  \vspace{-6pt}
  \captionsetup{justification=raggedright,singlelinecheck=false,font=small}
  \centering\small
  \setlength{\tabcolsep}{2pt}
  \caption{\textbf{Training Cost Comparison.}}
  \label{tab:cost-train-main}
  \vspace{2pt}
  \begin{tabular}{@{}lcc@{}}
    \toprule
    & Params $\downarrow$ & GPU-h $\downarrow$ \\
    \midrule
    \ttt{HyperSteer} & ${\sim}$9.2B & 102.39 \\
    \ttt{FLAS} & 255.6M & 18.89 \\
    \method{} & \textbf{20.6M} & \textbf{4.06} \\
    \bottomrule
  \end{tabular}
\end{wraptable}

\textbf{Training Scalability.}
\label{sec:exp-cost}
For Gemma-2-9B in H200 GPU-hours (\cref{tab:cost-train-main}), \method{} uses $4.65\times$ fewer GPU-hours than \ttt{FLAS} and $25\times$ fewer than \ttt{HyperSteer}, with $12.4\times$ fewer trainable parameters than \ttt{FLAS}.
\method{} trains using only instruction text, whereas \ttt{FLAS} and \ttt{HyperSteer} require labelled responses.
With on-policy supervision across adherence levels, we found training for more than one epoch gives no consistent gain.
\WFclear

\subsection{Controlling Persona Expression Levels}
\label{sec:exp-persona}
\textbf{Evaluation Setup.}
We evaluate \methodL{} on seven personas excluded from training, following the evaluation protocol in~\citep{chen2025persona}.
We compare \methodL{}, prompting, prompt+\methodL{}, prompting w/ level, and CFG.
Three judges score persona expression from 0 to 100, and we use the median rating for each response.
We refer to \cref{app:persona} for generation, judge details, and additional results.

\begin{figure}[t]
  \centering
  \includegraphics[width=0.92\linewidth]{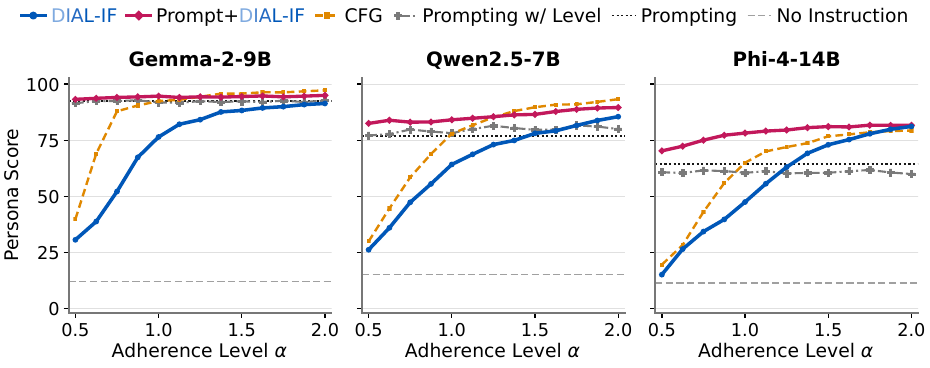}
  \vspace{2pt}
  \caption{\textbf{Fine-Grained Persona Expression Levels.} \methodL{} controls intermediate persona expression levels as $\alpha$ increases.
  Combining prompting with \methodL{} further raises persona scores.}
  \label{fig:persona-dial}
\end{figure}

\textbf{Fine-Grained Persona Expression.}
\Cref{fig:persona-dial} shows that \methodL{}'s persona score increases with the adherence level, more linearly than CFG, using one shared adapter per LLM.
On Gemma, it rises from 12.0 without instructions to 76.4 at $\alpha=1$ and 91.4 at $\alpha=2$, close to prompting's 92.5.
The adapter thus covers intermediate persona expression levels below, and on Qwen2.5-7B and Phi-4-14B above, prompting.
% Stating the adherence level in the prompt does not reach these levels: prompting w/ level stays within 6.2 points of prompting's persona score on every LLM (\cref{app:persona}).
Combining prompting with \methodL{} further raises persona expression.
On Qwen, the persona score rises from prompting's 77.0 to 82.5, 84.1, and 89.6 at $\alpha=0.5$, 1, and 2.
{On Phi-4-14B, the score rises from prompting's 64.3 to 70.2 and 78.2 at $\alpha=0.5$ and 1.}

\suppressfloats[t]
\begin{table}[t]
  \centering\small
  \setlength{\tabcolsep}{4pt}\renewcommand{\arraystretch}{0.95}
  \caption{\textbf{Verifiable Instruction Following on \ttt{IFEval}.} Prompt-level strict accuracy (\%) across six LLM families.
  Bold and underline indicate the best and second-best results per LLM.}
  \label{tab:if-main-dial}
  \vspace{1pt}
  \resizebox{\linewidth}{!}{%
  \begin{tabular}{lccccccc}
    \toprule
    Method & $\alpha$ & Gemma-2-9B & Qwen2.5-7B & Llama-3.1-8B & Phi-4-14B & LFM2.5-1.2B & Ministral-8B \\
    \midrule
    No instruction & -- & 9.6 & 9.2 & 9.8 & 10.0 & 9.8 & 8.5 \\
    Prompting & -- & 72.8 & \textbf{72.3} & 73.6 & 61.7 & 83.4 & 53.4 \\
    \midrule
    \multirow{4}{*}{CFG} & 0.5 & 21.4 & 30.5 & 21.3 & 19.0 & 50.1 & 14.0 \\
     & 1 & 72.5 & 71.2 & 74.7 & 64.1 & 83.7 & 54.5 \\
     & 1.5 & \textbf{78.0} & \textbf{72.3} & \underline{77.1} & \underline{72.1} & 81.3 & 64.7 \\
     & 2 & 76.2 & \textbf{72.3} & 75.4 & \textbf{72.5} & 81.1 & \underline{67.1} \\
    \midrule
    \multirow{4}{*}{\methodL{}} & 0.5 & 19.0 & 24.0 & 17.7 & 13.7 & 37.9 & 9.4 \\
     & 1 & 44.4 & 48.1 & 41.6 & 33.3 & 53.0 & 29.6 \\
     & 1.5 & 50.8 & 50.5 & 45.8 & 39.9 & 54.3 & 42.3 \\
     & 2 & 53.8 & 49.9 & 47.5 & 43.3 & 57.1 & 40.9 \\
    \midrule
    \multirow{4}{*}{P+\methodL{}} & 0.5 & 77.1 & \underline{71.9} & \textbf{78.0} & 63.2 & \underline{85.2} & 58.0 \\
     & 1 & 76.9 & \underline{71.9} & 75.6 & 66.7 & \textbf{85.6} & 65.1 \\
     & 1.5 & 77.1 & 70.6 & 73.8 & 68.2 & 83.7 & \textbf{67.5} \\
     & 2 & \underline{77.4} & 70.1 & 71.2 & 67.5 & 81.3 & 66.4 \\
    \bottomrule
  \end{tabular}}
  \vspace{0pt}
\end{table}

\subsection{Verifiable Instruction Following}
\label{sec:exp-if}
\textbf{Evaluation Setup.}
We evaluate \methodL{} on 541 \ttt{IFEval} prompts, which contain verifiable instructions on format, lexical content, and language~\citep{zhou2023instruction}.
We note that all 25 instruction types in \ttt{IFEval} are covered by \ttt{SteerIF}, though matching instruction texts are removed and 376 of its requests occur in training.
The official checker assigns a binary pass/fail result to each instruction.
A prompt passes only when all its instructions are followed.
Unlike concept and persona expression scores, these binary checks do not measure intermediate adherence levels per response.
We compare prompt-level strict accuracy at four adherence levels ($\alpha=0.5, 1, 1.5, 2$).

\textbf{Instruction Following Performance.}
\methodL{} alone generally improves instruction following as the adherence level increases (\cref{tab:if-main-dial}).
On Gemma-2-9B, accuracy rises from 9.6 without instructions to 53.8 at $\alpha=2$.
At $\alpha=1$, combining prompting with \methodL{} improves over prompting on five of the six evaluated LLMs, while Qwen scores 71.9 compared with prompting's 72.3.
At their best levels, CFG and prompt+\methodL{} are within two points of each other on five of the six LLMs, and CFG is higher on Phi-4-14B (72.5 against 68.2), while CFG requires two LLM passes per token.
On Gemma, prompt+\methodL{} reaches 76.9, compared with prompting's 72.8.
Llama reaches 75.6, compared with 73.6.
On Phi-4-14B, LFM2.5, and Ministral-8B, it reaches 66.7, 85.6, and 65.1, versus 61.7, 83.4, and 53.4.
Thus, \method{}'s adapter carries information from unseen instruction texts.
We observe similar gains on the \ttt{IFEval} questions that do not appear in training (\cref{app:results-if}).

\textbf{Discussion.}
We differentiate instructions that apply to the whole response, such as its length or format (global), from instructions that specify particular words or positions (local).
On Qwen2.5-7B, prompt+\methodL{} at $\alpha=2$ follows local instructions better than prompting but global instructions worse, which helps explain why its overall accuracy stays below prompting's on this LLM (\cref{tab:if-main-dial}).
\Cref{app:results-if} covers evaluation details, category analysis, and unseen instruction types.

    \section{Conclusion}
\label{sec:conclusion}
\label{sec:limitations}
We introduce \method{} to build ``\emph{steerable language models}'' that provide fine-grained control
over instruction-adherence levels while keeping the LLM frozen.
On-policy distillation from CFG trains a lightweight, text-conditioned steering adapter. \method{} provides fine-grained control over concept and persona
expression, and combining it with prompting can improve verifiable instruction following. Training is scalable, using only instruction text and a shared adapter without labelled responses or
per-instruction fitting. However, high adherence levels can still reduce response
coherence, and \method{} does not improve every verifiable instruction type. Future work should explore better
classifier architectures and training losses, and combine on-policy distillation with verifiable rewards or
judge feedback to filter training rollouts. Future work could also explore training at negative $\alpha$, matching CFG with negative weights to steer
away from specific styles, topics, or vocabulary. 
% We hope \method{} inspires further work on fine-grained LLM adaptation and its practical applications.
% We also encourage exploring steerable language models in use cases where users need to precisely adjust instruction-adherence levels.

    % \clearpage
    \bibliographystyle{unsrt}
    \begingroup\hypersetup{urlcolor=black}\urlstyle{same}
    \bibliography{refs}
    \endgroup
    
    % Appendix
    \clearpage
    \appendix
    \crefalias{section}{appendix}\crefalias{subsection}{appendix}  % cleveref: "Appendix B", "Appendix E.3"
    \raggedbottom
    % Appendix floats may use available top/bottom space and start at the top of float pages.
    \setcounter{topnumber}{5}\setcounter{bottomnumber}{3}\setcounter{totalnumber}{6}
    \renewcommand{\topfraction}{0.95}\renewcommand{\bottomfraction}{0.9}
    \renewcommand{\textfraction}{0.05}\renewcommand{\floatpagefraction}{0.7}
    \setlength{\abovecaptionskip}{10pt}
    \setlength{\textfloatsep}{12pt plus 2pt minus 2pt}
    \setlength{\floatsep}{12pt plus 2pt minus 2pt}
    \setlength{\intextsep}{12pt plus 2pt minus 2pt}
    \makeatletter
    \setlength{\@fptop}{0pt}\setlength{\@fpsep}{12pt}
    \setlength{\@fpbot}{0pt plus 1fil}
    \makeatother
    \begin{center}
      {\bfseries\Large Appendix of ``\method{}: Steerable Language Models with\\
      \textcolor{dialRampA}{D}istilled \textcolor{dialRampB}{I}nstruction-\textcolor{dialRampC}{A}dherence \textcolor{dialRampD}{L}evels''}
    \end{center}
    \startcontents
    \printcontents{}{1}{\textbf{Table of Contents}\vskip3pt\hrule\vskip5pt}
    \vskip3pt\hrule\vskip5pt
    \clearpage
    % DIAL extended related work.
\section{Extended Related Work}
\label{app:related}
This appendix expands on \cref{sec:related} and adds the distillation work our training builds on.
\Cref{tab:method-comparison} compares methods on the three properties from \cref{sec:intro}: \emph{fine-grained}, \emph{efficient}, and \emph{scalable}.

\begin{table}[h]
  \centering
  \footnotesize
  \setlength{\tabcolsep}{4pt}
  \caption{Steering methods compared on the three desired properties.
  \emph{Adherence Level} means the adherence level for an instruction can be adjusted.
  \emph{Cost} is the inference cost per generated token.
  \emph{No labelled data} means no target completions, contrastive pairs, external judges, or labelled datasets are needed.
  \emph{Zero-shot} means a new target needs no re-fitting.
  \emph{General instr.} means the method is evaluated on general instructions.
  \emph{Per-token} means the edit is recomputed at each generated token.}
  \label{tab:method-comparison}
\vspace{6pt}
\resizebox{\linewidth}{!}{%
  \begin{tabular}{lcccccc}
    \toprule
    & Fine-grained & Efficient & \multicolumn{3}{c}{Scalable} & \\
    \cmidrule(lr){2-2}\cmidrule(lr){3-3}\cmidrule(lr){4-6}
    Method & Adherence Level & Cost & No labelled data & Zero-shot & General instr. & Per-token \\
    \midrule
    Prompting & \expN & Low & \expY & \expY & \expY & -- \\
    CFG~\citep{sanchez2023cfg,dong2026systemprompt} & \expY & High & \expY & \expY & \expY & -- \\
    \midrule
    DiffMean / CAA~\citep{rimsky2024caa} & \expN & Low & \expN & \expN & \expN & \expN \\
    Instruction vectors~\citep{stolfo2025instruction} & \expN & Low & \expY & \expN & \expY & \expN \\
    \ttt{PSR}~\citep{heyman2026psr} & \expN & Low & \expY & \expN & \expY & \expY \\
    \ttt{HyperSteer}~\citep{sun2025hypersteer} & \expN & Low & \expN & \expY & \expN & \expN \\
    \ttt{FLAS}~\citep{jin2026flas} & \expN & Medium & \expN & \expY & \expN & \expY \\
    \ttt{UniSteer}~\citep{shi2026unisteer} & \expN & High & \expN & \expY & \expY & \expY \\
    \midrule
    \method{} (ours) & \expY & Low & \expY & \expY & \expY & \expY \\
    \bottomrule
  \end{tabular}}

\end{table}

\textbf{Steerable Language Models.}
\ttt{SteerLM} conditions supervised finetuning on labelled attribute levels that users select at inference~\citep{dong2023steerlmattributeconditionedsft}, and Conditional Language Policy learns policies conditioned on weights over multiple reward objectives~\citep{wang2024conditionallanguagepolicy}.
Both finetune the whole LLM for numerical control over predefined attributes or objectives, using labelled data or a reward model.
\method{} keeps the LLM frozen and conditions a lightweight adapter on free-form instruction text, controlling adherence levels for new instructions without labelled responses, reward models, or retraining.

\textbf{Guided Inference and Distillation.}
Besides CFG for language models~\citep{sanchez2023cfg,dong2026systemprompt}, proxy-tuning combines the logits of several language models~\citep{liu2024proxytuning}.
Our training also draws on distillation.
Context distillation trains a language model to behave as if it had been prompted without seeing the prompt~\citep{askell2021general,snell2022context}, and prompt baking internalizes a prompt into the language model's weights~\citep{bhargava2024baking}.
In on-policy distillation (OPD), the student is trained on its own samples with token-level feedback from a teacher (typically forward KL, reverse KL, or JSD)~\citep{agarwal2024gkd,gu2024minillm}.
This approach has been used with a copy of the model that sees extra context as the teacher~\citep{ye2026opcd}, and for self-distillation of a reasoning model given privileged context~\citep{zhao2026opsd}.
\ttt{ExOPD} distills a sequence-level geometric mixture~\citep{yang2026exopd}, while \method{}'s mixture is defined separately at each token, so the two sequence distributions generally differ.
Guidance itself has also been distilled.
In diffusion, a student conditioned on the guidance weight learns every strength~\citep{meng2023guided}, and \ttt{PDM} supervises the conditional branch and the guided combination separately, after finding that matching only the combination corrupts the conditional branch~\citep{li2026pdm}.
Diversity-rewarded CFG distillation~\citep{cideron2024diversity} distills CFG into all weights of an autoregressive music model at one fixed scale with a negative prompt, training on the student's own samples, and adding a diversity reward because distillation alone reduces diversity.
\method{} uses the same ingredients, namely the language model as its own teacher with privileged context, on-policy rollouts, and a JSD loss that is an instance of \ttt{GKD}, but it distills every adherence level into one lightweight, text-conditioned adapter rather than one fixed scale into all weights.

\textbf{Activation Steering.}
Steering vectors can be extracted by contrasting activations on paired inputs~\citep{rimsky2024caa} or the outputs of attention heads~\citep{li2023iti}.
They can also be trained for each concept.
For instance, \ttt{RePS} trains them with a preference-style objective~\citep{wu2025reps}.
\ttt{AxBench} compares these steering methods against prompting and finds that it is difficult for them to beat prompting~\citep{wu2025axbench}.
Several methods decide when to apply the intervention.
\ttt{CAST} turns a vector on only when it detects a condition~\citep{lee2025cast}, \ttt{MAT-Steer} learns per-token gates over a fixed set of attributes~\citep{nguyen2025matsteer}, and \ttt{DSAS} learns a gate for each layer~\citep{dsas2025}.
\ttt{StTP} edits only the tokens whose projection falls below a certain boundary~\citep{herbster2026sttp}, and \ttt{GCAD} gates each token by its attention to system-prompt positions~\citep{kang2026gcad}.
Instruction-vector steering instead moves the projection onto an instruction vector to its average value under instructed inputs~\citep{stolfo2025instruction}.
Other methods learn directly from prompted behavior.
\ttt{PSR} trains a per-token gate on a fixed direction to imitate prompting~\citep{heyman2026psr}, and we adopt its \ttt{IFEval} and persona protocols.
\ttt{CDAS} matches the output of an interchange intervention to the model's output on an input that elicits the concept~\citep{bao2026cdas}, and a system prompt alone can yield a difference-of-means vector~\citep{blank2026subliminal}.
Rather than fixing the strength, some methods sample it during training~\citep{wu2025reps} or learn it for each concept~\citep{prosv2026}.
In every case, the parameters are fitted to one concept, attribute, or instruction.

\textit{Classifier-guided and geometric edits.} Projecting a hidden state onto the half-space of a probe with the smallest possible change has often been used in steering~\citep{cheng2024liseco,xu2024scav,zhang2025gcav}, where the probe is fitted for each attribute.
A similar projection under a state-dependent linear constraint is used in safe control~\citep{dalal2018safe}.
\ttt{SaP} iterates toward a learned safety polytope~\citep{ICML2025_chenlearning}, and \ttt{BarrierSteer} projects onto a first-order condition of a nonlinear safety score~\citep{tran2026barriersteer}.
Other projections account for the norm of the hidden state.
These include Householder pseudo-rotation~\citep{pham2024hpr}, angular steering in a fixed plane~\citep{vu2025angular}, spherical steering with a per-token confidence gate~\citep{you2026spherical}, selective steering~\citep{dang2026selective}, and Riemannian steps~\citep{ngo2026geosteer}.
One method solves a minimum-displacement problem under a unit-norm constraint~\citep{nguyen2026coast}, and an angle--norm analysis finds that a modest norm increase lowers perplexity at a fixed angle~\citep{aparin2026anglenorm}.
Affine minimal interventions on distributions have closed-form solutions~\citep{singh2024mimic}, and \ttt{FishBack} derives edits that minimize the off-target KL divergence under a pullback Fisher metric~\citep{fishback2026}.
\method{}'s projection builds on these methods, but generates its half-space and norm weight from the instruction and level at every token, and learns both from the guided distributions.

\textbf{Text-Conditioned Control.}
\ttt{HyperSteer} maps a steering prompt to a single vector that is added at every position, and its best variant cross-attends to the base model's residual stream~\citep{sun2025hypersteer}.
\ttt{FLAS} learns a velocity field at one layer of Gemma-2 that is conditioned on the description and integrated over several steps, and cross-attends to the description encoded by the frozen model's first layers.
It has about 255M trainable parameters and learns from concept-aligned completions, and on Gemma-2 2B its scaling curve crosses the prompting baseline only at 1{,}900--5{,}500 training concepts~\citep{jin2026flas}.
Persona Dosing fine-tunes \ttt{FLAS} on persona-labelled data and then calibrates its flow time against measured trait expression~\citep{jin2026personadosing}.
\method{} instead defines every level by the CFG teacher for any instruction.
\ttt{UniSteer} trains a text-conditioned flow with an external text encoder~\citep{shi2026unisteer}.
\ttt{Text-to-LoRA}, \ttt{HyperDAS}, and \ttt{Doc-to-LoRA} amortize other interventions over text~\citep{text2lora2025,hyperdas2025,charakorn2026doc2lora}.
\method{} is also text-conditioned, but outputs a half-space and a norm weight rather than a vector, velocity, or adapter, and learns from the frozen model's guided distributions rather than completions.

    \FloatBarrier
    \section{Implementation Details}
\label{app:setup}\label{app:implementation}

\subsection{Steering Adapter Architecture}
\label{app:arch}

\textbf{Frozen and Trainable Components.}
We apply \method{} at one layer $\ell$ of the frozen LLM.
The adapter reuses the layer's input normalization and query, key, and value projections, but not its output projection.
We train the 512-dimensional feature layer, LayerNorm, prediction heads, and read-out paths.
The adapter's feature computation at each token is given in \cref{eq:attn}.

\textbf{Adherence-Level Conditioning.}
We condition the threshold and norm weight on $\alpha$ through a learned shift to each base head:
\[
  \tau_t(\alpha)=\tau_0(f_t)+\Delta_t(\alpha),\qquad
  \lambda_t(\alpha)=1-\sigma\bigl(r(f_t)+\Delta^R_t(\alpha)+b_R\bigr).
\]
Here, $\tau_0$ and $r$ are linear maps, and
\[
  \Delta_t(\alpha)=\phi\bigl(\operatorname{stopgrad}(f_t),\log\alpha\bigr)
  -\phi\bigl(\operatorname{stopgrad}(f_t),0\bigr).
\]
The MLP $\phi$ has one 128-unit GELU hidden layer and a zero-initialized output layer.
A separate MLP defines $\Delta^R_t$ in the same way.
Both shifts are zero at $\alpha=1$.
The bias $b_R=\operatorname{logit}(0.05)$ initializes the norm weight near 0.95, making the initial edit close to the projection geometry.

\textbf{Attention and Memory.}
The current state $h_t$ provides the attention query.
Instruction and memory states provide the keys and values.
We do not apply positional rotation to instruction keys.
Memory keys use the frozen layer's rotary embedding relative to the current token position.

During inference, we cache instruction keys and values once per request.
A 16-entry ring buffer stores the memory keys and values of edited states.
During training, the memory instead uses the unedited states $h_{t-16},\dots,h_{t-1}$ from the student's rollout.
These states are available from one LLM inference pass, allowing us to compute the adapter's edits at all positions in parallel (\cref{alg:train}).
Training on edited memory states, which matches inference exactly, does not improve results (\cref{app:abl-arch}).

\textbf{Read-Out Paths and the Lighter Variant.}
We add two read-out paths to \method{} and \methodL{}.
A linear skip $Sz_t$ passes the attention context to the direction head.
The second path adds $W_r\operatorname{RMSNorm}(h_t)$ to the feature layer before LayerNorm.
The resulting computations are
\begin{gather*}
  f_t=\operatorname{GELU}\!\left(\operatorname{LayerNorm}\!\left(W_z z_t+b_z+
  W_r\operatorname{RMSNorm}(h_t)\right)\right),\\
  \widetilde u_t=W_u f_t+b_u+Sz_t,\qquad
  u_t=\frac{\widetilde u_t}{\nrmm{\widetilde u_t}}.
\end{gather*}
Both paths are zero-initialized and use a $0.1\times$ learning-rate multiplier.
\methodLite{} omits them.
On Gemma-2-9B, \method{} and \methodL{} have 20.6M trainable parameters, while \methodLite{} has 4.08M.
All variants keep the base LLM frozen throughout training and inference.

\subsection{Geometry of the Steering Edit}
\label{app:projrot}

\textbf{Minimum-Norm Projection.}
The adapter predicts $u_t$, $\tau_t(\alpha)$, and $\lambda_t(\alpha)$ from the current state and context, then holds them fixed while editing the state.
The classifier score is $s_t(v)=\langle u_t,v\rangle-\tau_t(\alpha)$.
The projection in \cref{eq:projrot} solves
\[
  \delta_t=\arg\min_\delta\ \tfrac12\nrmm{\delta}^2
  \quad\text{subject to}\quad \langle u_t,h_t+\delta\rangle\ge\tau_t(\alpha),
  \qquad h^P_t=h_t+\delta_t.
\]
Because $u_t$ has unit norm, a state outside the half-space takes a minimum-norm step along $u_t$ with magnitude $\tau_t(\alpha)-\langle u_t,h_t\rangle$.
A state inside the half-space remains unchanged.

\begin{figure}[htbp]
  \centering
  \includegraphics[width=0.48\linewidth]{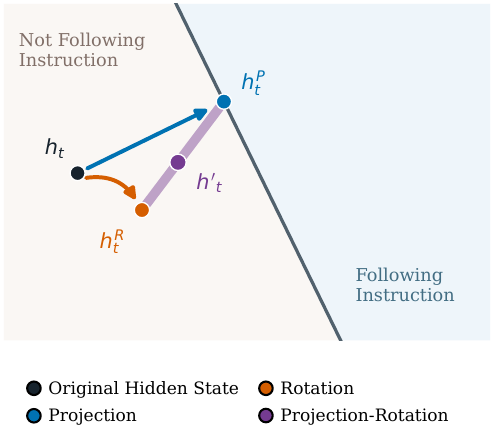}
  \caption{\textbf{Geometry of the Steering Edit.} The norm weight $\lambda_t$ interpolates between norm-preserving rotation ($\lambda_t=0$) and minimum-norm projection ($\lambda_t=1$).}
  \label{fig:projection-rotation}
\end{figure}

\textbf{Norm Adjustment.}
For nonzero $h_t$ and $h^P_t$, the state with the projected direction and original norm is
\begin{equation}
  h^R_t=\nrmm{h_t}\,\frac{h^P_t}{\nrmm{h^P_t}}.
  \label{eq:rotation-state}
\end{equation}
This preserves the original norm and rotates the state within $\operatorname{span}\{h_t,u_t\}$ without a rotation matrix.
The edited state is the interpolation $h'_t=(1-\lambda_t)h^R_t+\lambda_t h^P_t$, with norm $(1-\lambda_t)\nrmm{h_t}+\lambda_t\nrmm{h^P_t}$ (\cref{fig:projection-rotation}).
At $\lambda_t=1$, the state lies in the classifier's half-space.
Norm adjustment at $\lambda_t<1$ need not preserve that membership.
Neither property guarantees instruction adherence in the response.
The adapter generates a new classifier at each token and handles degenerate normalization cases numerically.

\subsection{Training Data}
\label{app:training-data}
We use separate training runs for \method{} on \ttt{Concept16k} and \methodL{} on \ttt{SteerIF}.
\method{} uses concept instructions, while \methodL{} covers concept steering, persona steering, and verifiable instruction following.

\textbf{Concept Instructions.}
The concept-steering pool contains 14{,}880 descriptions from \ttt{Concept16k}.
We remove all exact-match and near-duplicate descriptions of \ttt{Concept500} and \ttt{FLAS}'s 100 held-out concepts.

\textbf{SteerIF Construction.}
\ttt{SteerIF} contains 98{,}625 instruction texts in five categories (\cref{tab:steerif}).
The dataset includes the same 14{,}880 concept descriptions and 584 one-sentence persona instructions.
Of these, 57 are author-generated, kept after filtering out the evaluation personas of~\citep{chen2025persona} and related ones.
The other 527 are derived from persona adjectives: 510 occur only in \ttt{TDA}, 14 only in \ttt{MWE}, and 3 in both sources~\citep{condon2022tda,perez2022discovering}.
We count shared adjectives once.
We convert each adjective into a one-sentence persona description using the generation prompt in~\citep{chen2025persona}.
We remove matches to held-out persona groups using WordNet relations~\citep{miller1995wordnet}, embedding similarity with \ttt{all-mpnet-base-v2}~\citep{reimers2019sbert}, and a pairwise comparison with \ttt{Phi-4-14B}.

Single instructions specify verifiable or descriptive response requirements.
We collect 23{,}480 from \ttt{IF\_multi}~\citep{pyatkin2025ifbench}, 25{,}248 from \ttt{VerInstruct}~\citep{peng2025verif}, and 44 from the \ttt{IFEvalG} instruction registry~\citep{lambert2024tulu3}.
\ttt{VerInstruct} covers 13 categories, including literary devices, morphology, item listing, grammatical structure, and multilingual output.
We also generate 2{,}942 instructions by writing alternative phrasings and wrappers for types represented by only one template in the original sources.
These author-generated variants retain the original response requirement.

We construct 31{,}109 composite instructions by combining two (85\%) or three (15\%) compatible single instructions.
Each combination contains at most one requirement on length, case, language, structure, or punctuation.
We also include 338 \ttt{IFEval}-type instructions.
\ttt{SteerIF} therefore contains only instruction text, without responses or LLM outputs in any category.

\textbf{Request Pairing and Verification.}
We pair verifiable and descriptive instructions with 16--32 requests drawn from 376 instruction-free \ttt{IFEval} questions~\citep{zhou2023instruction,stolfo2025instruction}.
Concept and persona instructions use \ttt{AxBench}'s shared request corpus.
Code can verify 36{,}014 of the 82{,}823 verifiable and descriptive instructions.
The other 46{,}809, including all \ttt{VerInstruct} texts, receive supervision from the CFG teacher without a response checker.
All 25 \ttt{IFEval} instruction types appear in training, with at least 340 training instructions for every type in the dataset.

\textbf{Positive and Negative Instructions.}
We group 82{,}823 single and composite instructions into positive, negative, and other instructions (\cref{tab:steerif-polarity}).
The analysis excludes concept and persona instructions.
Positive instructions account for 42.4\% of this subset, negative instructions for 29.9\%, and the remaining 27.7\% specify lower, upper, range, or exact requirements.
Composite instructions contain a larger share of negative instructions (44.5\%) than single verifiable (14.4\%) or descriptive instructions (28.3\%).
These negative instructions specify what to avoid and are distinct from using a negative adherence level $\alpha$.

\begin{table}[htbp]
  \centering\small
  \setlength{\tabcolsep}{4pt}\renewcommand{\arraystretch}{1.1}
  \caption{\textbf{Positive and Negative Instructions in \ttt{SteerIF}.} Counts and percentages for single and composite instructions.
  Concepts and personas are excluded.}
  \label{tab:steerif-polarity}
  \vspace{6pt}
  \begin{tabular}{@{}lrccc@{}}
    \toprule
    Instructions & Count & Positive & Negative & Others \\
    \midrule
    Single, verifiable & 26{,}467 & 12{,}779 (48.3\%) & 3{,}814 (14.4\%) & 9{,}874 (37.3\%) \\
    Single, descriptive & 25{,}247 & 15{,}332 (60.7\%) & 7{,}141 (28.3\%) & 2{,}774 (11.0\%) \\
    Composite & 31{,}109 & 7{,}001 (22.5\%) & 13{,}835 (44.5\%) & 10{,}273 (33.0\%) \\
    \midrule
    Total & 82{,}823 & 35{,}112 (42.4\%) & 24{,}790 (29.9\%) & 22{,}921 (27.7\%) \\
    \bottomrule
  \end{tabular}
\end{table}

\begin{table}[htbp]
  \centering\small
  \setlength{\tabcolsep}{4pt}\renewcommand{\arraystretch}{1.1}
  \caption{\textbf{Lower, Upper, Range, and Exact Requirements.} Counts and examples for the 22{,}921 instructions grouped as Others in \cref{tab:steerif-polarity}.}
  \label{tab:steerif-bounds}
  \vspace{6pt}
  \begin{tabular}{@{}lr>{\raggedright\arraybackslash}p{0.67\linewidth}@{}}
    \toprule
    Requirement & Count & Example \\
    \midrule
    Lower & 6{,}416 & The letter k should appear at least 16 times. \\
    Upper & 5{,}777 & The response must not exceed 270 words. \\
    Range & 1{,}399 & Use at least 450 words and at most 480 words. \\
    Exact & 9{,}329 & There should be 7 paragraphs. \\
    \bottomrule
  \end{tabular}
\end{table}

\textbf{Evaluation Filtering.}
We remove training texts that match instructions from \ttt{IFEval}, \ttt{IFBench}, \ttt{FollowBench}~\citep{jiang2024followbench}, \ttt{InFoBench}~\citep{qin2024infobench}, \ttt{ComplexBench}~\citep{wen2024complexbench}, persona questions, or \ttt{AlpacaEval}~\citep{dubois2024lcalpacaeval}.
Matches are identical sentences, word-Jaccard similarity $\ge0.8$, or shared spans of at least six words.

\begin{table}[H]
  \centering\small
  \caption{\textbf{\ttt{SteerIF} Composition.} Training instruction counts by source.}
  \label{tab:steerif}
  \setlength{\tabcolsep}{4pt}\renewcommand{\arraystretch}{1.12}
  \vspace{6pt}
  \begin{tabular}{@{}>{\raggedright\arraybackslash}p{0.22\linewidth}>{\raggedright\arraybackslash}p{0.65\linewidth}r@{}}
    \toprule
    Category & Source & Count \\
    \midrule
    Concepts & \ttt{Concept16k}~\citep{wu2025axbench} & 14{,}880 \\
    Personas & 57 author-generated personas & 57 \\
     & \ttt{TDA}~\citep{condon2022tda} (510), \ttt{MWE}~\citep{perez2022discovering} (14), 3 shared between \ttt{TDA} and \ttt{MWE} & 527 \\
    Base instructions & \ttt{IFEval}-type instructions & 338 \\
    Single instructions & \ttt{IF\_multi} annotations & 23{,}480 \\
     & \ttt{VerInstruct} instruction texts & 25{,}248 \\
     & \ttt{IFEvalG} registry grid & 44 \\
     & Author-generated phrasings and wrappers & 2{,}942 \\
    Composite instructions & Author-generated combinations of 2 or 3 single instructions & 31{,}109 \\
    \midrule
    Total & & 98{,}625 \\
    \bottomrule
  \end{tabular}
\end{table}

\subsection{Training Configuration}
\label{app:optimization}\label{app:checkpoints}
\textbf{Concept16k Training.}
For the \ttt{AxBench} experiments, we train \method{} on 14{,}880 \ttt{Concept16k} descriptions.
We train the Gemma-2-9B adapter from scratch for one epoch on H200 GPUs, using 4.06 GPU-hours.
The adapter edits layer 20, counting from 0 as in \ttt{AxBench}, and uses the architecture and read-out paths in \cref{app:arch}.

\textbf{SteerIF Training.}
In separate runs, we train \methodL{} and \methodLite{} for one epoch on \ttt{SteerIF}.
Each adapter is shared across concept, persona, and verifiable instructions, and is also evaluated on \ttt{AxBench}.
Each adapter edits one layer near half the LLM's depth, counting from 0: layer 20 of 42 for Gemma-2-9B, 13 of 28 for Qwen2.5-7B, 15 of 32 for Llama-3.1-8B, 19 of 40 for Phi-4-14B, 7 of 16 for LFM2.5-1.2B, and 15 of 36 for Ministral-8B.
We train both variants for each of these LLMs except Ministral-8B, for which we train only \methodL{}, and both variants use the same layer.
On Gemma-2-9B, \methodL{} has 20.6M parameters and \methodLite{} has 4.08M.
The variants differ in the read-out paths described in \cref{app:arch} while sharing all other components.

On Gemma-2-9B, we train both variants from scratch with the same data, steps, and seed.

\textbf{Teacher Prompts.}
We format the CFG teacher's prompted input for each instruction type.
Verifiable and descriptive instructions follow the request after a blank line, as in the prompting baseline.
Persona instructions precede the request because Gemma-2 has no system role.
For concept instructions, we sample one of four length-free templates per instruction--request pair and epoch (\cref{tab:teacher-templates}).
\Cref{tab:teacher-prompts} gives the exact prompts.
\method{} uses only concept instructions, while \methodL{} uses all three instruction types.
\newcommand{\tpq}{\textquotesingle}
\newcommand{\tpslot}[1]{$\langle$\textit{#1}$\rangle$}
\begin{table}[htbp]
  \centering\footnotesize
  \caption{\textbf{Concept-Steering Teacher Prompts.} Each template inserts its numbered items into the shared template.
  \tpslot{request} is the user request and \tpslot{concept} the concept description.
  The persona template adds one sentence before the template.}
  \label{tab:teacher-prompts}
  \renewcommand{\arraystretch}{1.1}
  \vspace{6pt}
\begin{tabular}{@{}>{\raggedright\arraybackslash\ttfamily}p{0.97\linewidth}@{}}
    \toprule
    \multicolumn{1}{@{}l}{\textbf{Shared template}} \\
    Given the following instruction:\newline\newline \tpslot{request}\newline\newline Your task is to:\newline\newline
    \tpslot{numbered items of the template}\newline\newline **Formatting Guidelines:**\newline\newline
    - Return only the response to the instruction.\newline
    - Write the final content (or appropriate format for the genre) in plain text.\newline
    - Do not include any additional text, explanations, or formatting.\newline\newline
    **Final Answer:** Return only the final content, following the guidelines above. \\
    \midrule
    \multicolumn{1}{@{}l}{\textbf{Concept in every sentence}} \\
    1. Provide a response that fully and correctly answers the instruction.\newline
    2. Give one focused, well-organised answer rather than a list of loosely related points.\newline
    3. Weave elements related to \tpq\tpslot{concept}\tpq{} into every sentence of it.\newline
    4. Try to avoid copying words from the definition of \tpq\tpslot{concept}\tpq{} if possible. \\
    \midrule
    \multicolumn{1}{@{}l}{\textbf{\ttt{AxBench}-style, fluent}} \\
    1. Provide a response that incorporates elements related to \tpq\tpslot{concept}\tpq.\newline
    2. Try to avoid copying words from the definition of \tpq\tpslot{concept}\tpq{} if possible.\newline
    3. Ensure that your response relates to \tpq\tpslot{concept}\tpq, even if the overall meaning is not fully coherent.\newline
    4. Within that, answer the instruction as fully and correctly as you can.\newline
    5. Write clear, natural, complete sentences: no keyword lists, fragments or repeated phrases. \\
    \midrule
    \multicolumn{1}{@{}l}{\textbf{Persona, fluent}} \\
    \textit{Before the template:} You are a writer who is fascinated by \tpq\tpslot{concept}\tpq{} and brings it into
    everything you write, while always giving complete and correct answers.\newline\newline
    1. Provide a response that fully and correctly answers the instruction.\newline
    2. Let your fascination with \tpq\tpslot{concept}\tpq{} show clearly throughout the response.\newline
    3. Try to avoid copying words from the definition of \tpq\tpslot{concept}\tpq{} if possible.\newline
    4. Write clear, natural, complete sentences: no keyword lists, fragments or repeated phrases. \\
    \midrule
    \multicolumn{1}{@{}l}{\textbf{Answer, recurring concept}} \\
    1. Provide a response that fully and correctly answers the instruction.\newline
    2. Make elements related to \tpq\tpslot{concept}\tpq{} a clear, recurring part of the answer, woven in naturally
    rather than tacked on.\newline
    3. Try to avoid copying words from the definition of \tpq\tpslot{concept}\tpq{} if possible. \\
    \bottomrule
  \end{tabular}
\end{table}

\textbf{Adherence Levels and Rollouts.}
We sample 24 adherence levels per instruction--request pair from $\operatorname{LogUniform}[0.5,2]$.
Student rollouts use temperature 1 and generate at most 128--1{,}024 new tokens, depending on the instruction type.

\textbf{Loss and Optimization.}
We use JSD with $\beta=0.5$ over the union of the base, prompted, and student top-64 tokens.
We add the regularization term $0.1\,\mathbb E\,[\nrmm{h'_t-h_t}^2/\nrmm{h_t}^2]$.
This term is approximately 3\% of the JSD at the end of training.
The optimizer uses a learning rate of $4\times10^{-3}$ for every adapter.

\textbf{Choice of Adherence Range.}
The CFG teacher sweep on \ttt{FLAS}'s held-out concepts (\cref{fig:level-dial,tab:level-dial-numbers}) supports this range.
At $\alpha=0.5$, the teacher's concept score is 0.23, close to the unprompted LLM's 0.14.
Its fluency falls from 1.58 to 1.27 as $\alpha$ increases from 1 to 2.
Beyond $\alpha=2$, its overall score falls from 1.32 to 1.23 at $\alpha=2.5$.
We therefore use $\alpha=2$ as the upper bound when sampling adherence levels.

\subsection{Pseudocode}
\label{app:pseudocode}
\newsavebox{\dialcodebox}
We provide PyTorch-style pseudocode for the adapter, training, and inference.
We omit batching, padding, grouped-query attention, and numerical precision.
The modules \texttt{norm}, \texttt{W\_q}, \texttt{W\_k}, \texttt{W\_v}, and \texttt{attn} reuse the frozen layer's input normalization, projections, and attention scoring, including the attention scale and logit soft-cap at the selected LLM layer.

\begin{figure}[H]
\centering
\begin{lrbox}{\dialcodebox}
\begin{minipage}{\dimexpr\linewidth-2\fboxsep-2\fboxrule\relax}
\begin{footnotesize}
\begin{verbatim}
class DIAL(nn.Module):
    def __init__(self, layer, d_model, d_attn, d_feat=512, readout=True):
        self.layer = layer
        self.proj = nn.Linear(d_attn, d_feat)
        self.ln = nn.LayerNorm(d_feat)
        self.dir = nn.Linear(d_feat, d_model)
        self.dir_bias = nn.Parameter(torch.zeros(d_model))
        self.thr = nn.Linear(d_feat, 1)
        self.nrm = nn.Linear(d_feat, 1)
        self.thr_lvl = MLP(d_feat + 1, 128, 1)
        self.nrm_lvl = MLP(d_feat + 1, 128, 1)
        self.readout = readout
        if readout:
            self.skip = nn.Linear(d_attn, d_model, bias=False)
            self.res = nn.Linear(d_model, d_feat, bias=False)
    def kv(self, s, pos=None):
        x = self.layer.norm(s)
        k, v = self.layer.W_k(x), self.layer.W_v(x)
        return (k if pos is None else rope(k, pos)), v
    def heads(self, h, ctx, alpha):
        z = self.proj(ctx)
        if self.readout:
            z = z + self.res(self.layer.norm(h))
        f = F.gelu(self.ln(z))
        d = self.dir(f) + self.dir_bias
        if self.readout:
            d = d + self.skip(ctx)
        u = d / d.norm(dim=-1, keepdim=True)
        fs = f.detach()
        shift = lambda m: m(cat([fs, log(alpha)])) - m(cat([fs, 0.0]))
        tau = self.thr(f) + shift(self.thr_lvl)
        lam = 1 - torch.sigmoid(self.nrm(f) + shift(self.nrm_lvl) + logit(0.05))
        return u, tau, lam
    def forward(self, h, Ik, Iv, Mk, Mv, alpha, pos):
        q = self.layer.W_q(self.layer.norm(h))
        ctx = self.layer.attn(q, keys=[Ik, Mk], values=[Iv, Mv],
                              rope_q=pos)
        u, tau, lam = self.heads(h, ctx, alpha)
        return edit(h, u, tau, lam)
def edit(h, u, tau, lam):
    gap = tau - (u * h).sum(-1, keepdim=True)
    hP = h + gap.clamp(min=0) * u
    s = (lam + (1 - lam) * h.norm(dim=-1, keepdim=True)
         / hP.norm(dim=-1, keepdim=True))
    return s * hP
\end{verbatim}
\end{footnotesize}
\end{minipage}
\end{lrbox}
\fcolorbox{black!15}{black!3}{\usebox{\dialcodebox}}
\vspace{2pt}
\caption{\textbf{\method{}'s Steering Adapter.} The adapter attends to instruction and memory states, predicts the classifier, and applies a closed-form edit to the LLM's hidden state.}
\label{alg:dial}
\end{figure}

\begin{figure}[htbp]
\centering
\begin{lrbox}{\dialcodebox}
\begin{minipage}{\dimexpr\linewidth-2\fboxsep-2\fboxrule\relax}
\begin{footnotesize}
\begin{verbatim}
for chunk in epoch:
    # on-policy rollouts from the inference engine, one chunk ahead of the optimizer
    batch = [(x, c, alpha ~ LogUniform(0.5, 2)) for x, c in chunk]
    Y = engine.generate(dial, batch, temperature=1.0)   # length cap by instruction type
    for x, c, alpha, y in minibatches(batch, Y):
        I = model.hidden(c, layer=l)                    # instruction states
        with torch.no_grad():                           # teacher, Eq. (1)
            lb = model(x + y).log_softmax(-1)           # request alone
            lp = model(prompt(x, c) + y).log_softmax(-1)  # request with the instruction
            lt = (1 - alpha) * lb + alpha * lp
        h = model.hidden_upto(x + y, layer=l)  # raw states, not affected by edits at l
        M = sliding_window(h, 16)                       # memory h_{t-16}, ..., h_{t-1}
        # all positions in parallel
        h_edit = dial(h, *dial.kv(I), *dial.kv(M, pos), alpha, pos)
        ls = model.forward_from(h_edit, layer=l).log_softmax(-1)
        # token union, response positions
        S = topk(lb, 64) | topk(lp, 64) | topk(ls, 64)
        loss = jsd(lt[S].softmax(-1), ls[S].softmax(-1)).mean()  # Eq. (2), beta = 0.5
        loss += 0.1 * ((h_edit - h).norm(dim=-1)**2 / h.norm(dim=-1)**2).mean()
        loss.backward(); opt.step(); opt.zero_grad()    # gradients reach only DIAL
\end{verbatim}
\end{footnotesize}
\end{minipage}
\end{lrbox}
\fcolorbox{black!15}{black!3}{\usebox{\dialcodebox}}
\vspace{2pt}
\caption{\textbf{On-Policy Training of \method{}.} We generate student rollouts at different adherence levels and train the adapter to match the CFG distribution using JSD loss, while keeping the LLM frozen.}
\label{alg:train}
\end{figure}

\begin{figure}[htbp]
\centering
\begin{lrbox}{\dialcodebox}
\begin{minipage}{\dimexpr\linewidth-2\fboxsep-2\fboxrule\relax}
\begin{footnotesize}
\begin{verbatim}
def generate(model, dial, x, c, alpha, max_new_tokens):
    Ik, Iv = dial.kv(model.hidden(c, layer=l))          # once per request, cached
    Mk, Mv = Ring(16), Ring(16)  # memory keys/values, ring buffer
    def hook(h, pos):  # layer l, every prompt and new position
        h_edit = dial(h, Ik, Iv, Mk.items(), Mv.items(), alpha, pos)
        k, v = dial.kv(h_edit, pos)
        Mk.push(k); Mv.push(v)                          # replaces the oldest entry
        return h_edit
    model.register_hook(layer=l, fn=hook)
    return model.generate(x, max_new_tokens)            # one forward pass per token
\end{verbatim}
\end{footnotesize}
\end{minipage}
\end{lrbox}
\fcolorbox{black!15}{black!3}{\usebox{\dialcodebox}}
\vspace{2pt}
\caption{\textbf{Inference with \method{}.} The instruction is encoded once, and cached instruction and memory states condition the edit at each token.
Generation uses one LLM inference pass per token.}
\label{alg:infer}
\end{figure}

    \FloatBarrier
    \section{Experimental Settings and Additional Results}
\label{app:results}

\subsection{Controlling Concept Expression Levels}
\label{app:results-concept}
\textbf{Evaluation Setup.}
Following \ttt{AxBench}, we score the steered LLM based on how it expresses the concept, follows the original request, and preserves fluency from $0$ to $2$~\citep{wu2025axbench}.
We take the overall score as the harmonic mean of the three scores.
The adherence-level curves in \cref{fig:level-dial} use the median rating from \ttt{Phi-4-14B}, \ttt{Qwen3.8-27B}~\citep{qwen2026qwen38}, and \ttt{gpt-4o-mini} for each criterion.
All methods generate ten $128$-token responses per concept from \ttt{AlpacaEval} requests at temperature $1$.

On \ttt{Concept500}, we score requests 5--9 after the published per-concept methods select a steering level on requests 0--4~\citep{heyman2026psr}.
The Held-In column of \cref{tab:main-axb-9b} is \ttt{Concept500}, and the Held-Out column is \ttt{FLAS}'s released $100$-concept held-out split~\citep{jin2026flas}, on which we score all ten requests.
\ttt{FLAS} trains on 480 of the 500 \ttt{Concept500} concepts; the other 20 are in its held-out split, so the two columns share 20 concepts, whereas \method{} trains on none of its concepts, and neither method trains on a held-out concept.
\ttt{HyperSteer}'s published held-out score uses its own unreleased $500$-concept set~\citep{sun2025hypersteer}, on which prompting scores $1.091$.
Using $256$-token responses and a different request draw, \ttt{FLAS} reports $1.185$ and $1.113$ on its held-in and held-out splits.

\method{} and \ttt{FLAS} use one fixed level across concepts.
We use $\alpha\in\{1,2\}$ for \method{} alone, $\alpha\in\{0.5,1\}$ with prompting, and $T=2$ for \ttt{FLAS}.
The published per-concept results select one of \ttt{AxBench}'s 14 steering factors (0.2--5.0) for each concept on requests 0--4.
\ttt{LoReFT} and \ttt{LoRA} steer four layers (12, 20, 31, and 39), while the other methods steer layer $20$.
Method differences are paired by concept.
Main-table comparisons use $\alpha\le2$, while the adherence-level curves show figure-only extrapolation above $2$ to examine behavior beyond the training range.

\Cref{tab:level-dial-numbers} reports the scores plotted in \cref{fig:level-dial} at each adherence level.

\begin{table}[htbp]
  \centering\small
  \caption{\textbf{Fine-Grained Concept Expression Levels.} Scores plotted in \cref{fig:level-dial} on \ttt{FLAS}'s 100 held-out concepts, using ten 128-token responses per concept and the median of three judges (\ttt{Phi-4-14B}, \ttt{Qwen3.8-27B}, and \ttt{gpt-4o-mini}).}
  \label{tab:level-dial-numbers}
  \vspace{6pt}
  \begin{minipage}[t]{0.48\linewidth}
    \centering\small
    \setlength{\tabcolsep}{1.5pt}\renewcommand{\arraystretch}{1.05}
    \resizebox{\linewidth}{!}{%
    \begin{tabular}{@{}ccccc@{}}
      \toprule
      Level & Concept & Request & Fluency & Overall score \\
      \midrule
      \multicolumn{5}{@{}l}{\method{} ($\alpha$)} \\
      0.5 & 0.202 & 1.980 & 1.634 & 0.233 \\
      0.625 & 0.298 & 1.988 & 1.624 & 0.344 \\
      0.75 & 0.510 & 1.970 & 1.554 & 0.568 \\
      0.875 & 0.735 & 1.969 & 1.539 & 0.779 \\
      1 & 0.926 & 1.966 & 1.467 & 0.961 \\
      1.125 & 1.052 & 1.955 & 1.456 & 1.064 \\
      1.25 & 1.147 & 1.945 & 1.411 & 1.132 \\
      1.375 & 1.236 & 1.933 & 1.386 & 1.165 \\
      1.5 & 1.294 & 1.927 & 1.363 & 1.195 \\
      1.625 & 1.347 & 1.924 & 1.320 & 1.213 \\
      1.75 & 1.372 & 1.916 & 1.298 & 1.224 \\
      1.875 & 1.397 & 1.916 & 1.255 & 1.220 \\
      2 & 1.417 & 1.904 & 1.261 & 1.236 \\
      \midrule
      \multicolumn{5}{@{}l}{prompt+\method{} ($\alpha$)} \\
      0.5 & 1.354 & 1.947 & 1.438 & 1.351 \\
      0.625 & 1.432 & 1.939 & 1.471 & 1.408 \\
      0.75 & 1.500 & 1.944 & 1.447 & 1.445 \\
      0.875 & 1.506 & 1.934 & 1.454 & 1.457 \\
      1 & 1.576 & 1.940 & 1.428 & 1.489 \\
      1.125 & 1.566 & 1.932 & 1.402 & 1.464 \\
      1.25 & 1.633 & 1.933 & 1.399 & 1.501 \\
      1.375 & 1.649 & 1.931 & 1.359 & 1.493 \\
      1.5 & 1.668 & 1.925 & 1.367 & 1.517 \\
      1.625 & 1.674 & 1.921 & 1.367 & 1.506 \\
      1.75 & 1.684 & 1.928 & 1.350 & 1.512 \\
      1.875 & 1.684 & 1.923 & 1.352 & 1.499 \\
      2 & 1.694 & 1.921 & 1.349 & 1.490 \\
      \bottomrule
    \end{tabular}}
  \end{minipage}\hfill
  \begin{minipage}[t]{0.48\linewidth}
    \centering\small
    \setlength{\tabcolsep}{1.5pt}\renewcommand{\arraystretch}{1.05}
    \resizebox{\linewidth}{!}{%
    \begin{tabular}{@{}ccccc@{}}
      \toprule
      Level & Concept & Request & Fluency & Overall score \\
      \midrule
      \multicolumn{5}{@{}l}{CFG ($\alpha$)} \\
      0.5 & 0.229 & 1.980 & 1.604 & 0.256 \\
      0.625 & 0.411 & 1.972 & 1.605 & 0.455 \\
      0.75 & 0.682 & 1.969 & 1.601 & 0.752 \\
      0.875 & 0.937 & 1.958 & 1.589 & 1.004 \\
      1 & 1.134 & 1.957 & 1.577 & 1.161 \\
      1.125 & 1.261 & 1.943 & 1.556 & 1.243 \\
      1.25 & 1.339 & 1.942 & 1.522 & 1.291 \\
      1.375 & 1.409 & 1.927 & 1.504 & 1.327 \\
      1.5 & 1.458 & 1.926 & 1.454 & 1.331 \\
      1.625 & 1.495 & 1.921 & 1.405 & 1.341 \\
      1.75 & 1.512 & 1.908 & 1.339 & 1.306 \\
      1.875 & 1.578 & 1.901 & 1.334 & 1.326 \\
      2 & 1.582 & 1.902 & 1.275 & 1.317 \\
      \midrule
      \multicolumn{5}{@{}l}{\ttt{FLAS} ($T$)} \\
      0.5 & 0.603 & 1.927 & 1.454 & 0.618 \\
      0.625 & 0.951 & 1.885 & 1.343 & 0.919 \\
      0.75 & 1.057 & 1.877 & 1.331 & 0.999 \\
      0.875 & 1.151 & 1.869 & 1.333 & 1.080 \\
      1 & 1.181 & 1.849 & 1.302 & 1.075 \\
      1.125 & 1.201 & 1.878 & 1.303 & 1.109 \\
      1.25 & 1.225 & 1.862 & 1.306 & 1.134 \\
      1.375 & 1.206 & 1.869 & 1.319 & 1.113 \\
      1.5 & 1.217 & 1.876 & 1.314 & 1.126 \\
      1.625 & 1.224 & 1.859 & 1.308 & 1.125 \\
      1.75 & 1.224 & 1.876 & 1.284 & 1.121 \\
      1.875 & 1.260 & 1.859 & 1.268 & 1.147 \\
      2 & 1.260 & 1.879 & 1.286 & 1.158 \\
      \bottomrule
    \end{tabular}}
  \end{minipage}
\end{table}

\textbf{Baselines.}
The prompting baseline uses \ttt{AxBench}'s concept-specific prompts written by \ttt{gpt-4o-mini}.
Per-concept methods train one steering method and select one level for each concept.
They include \ttt{LoReFT}, \ttt{LoRA}, \ttt{ReFT-r1}, \ttt{DiffMean}, and \ttt{SAE} from \ttt{AxBench}~\citep{wu2025axbench}, SV + \ttt{RePS}~\citep{wu2025reps}, and S-\ttt{PSR}, the single-layer variant of \ttt{PSR}~\citep{heyman2026psr}.
Text-conditioned methods use one network across concepts.
\ttt{HyperSteer} maps a concept description to a steering vector~\citep{sun2025hypersteer}, while \ttt{FLAS} integrates a description-conditioned flow~\citep{jin2026flas}.
We use published results for the per-concept methods and \ttt{HyperSteer}.
We run \ttt{FLAS} with its released Gemma-2-9B checkpoint, original code, and three flow steps.
CFG applies the teacher in \cref{eq:guided} at inference, requiring two inference passes per token as a reference.
We do not compare with \ttt{Persona Dosing}~\citep{jin2026personadosing} or \ttt{UniSteer}~\citep{shi2026unisteer}, whose code and checkpoints are not released.
We also omit simpler level-conditioned edits, such as an additive steering vector or a difference-of-means vector between prompted and unprompted activations scaled by $\alpha$.
Such single-vector edits underperform text-conditioned methods such as \ttt{FLAS} and reduce fluency at high strengths~\citep{wu2025axbench,jin2026flas}: \ttt{DiffMean}, a difference-of-means vector, scores 0.322 on \ttt{Concept500}, against 0.997 for \ttt{FLAS} (\cref{tab:main-axb-9b}).

\textbf{Prompting w/ Level.}
We append ``Follow this instruction at adherence level $x$/2, where 0 means ignore it, 1 means follow it normally, and 2 means follow it as strongly as possible.'' to \ttt{AxBench}'s prompt, with $x$ equal to the adherence level $\alpha$ in \cref{fig:level-dial}.
Levels out of 10 and verbal levels from ``very slightly'' to ``extremely'' behave the same way: from the lowest to the highest level, the concept score changes by only 0.10--0.21, compared with 1.18--1.40 for \method{} (\cref{tab:concept-level-prompt}).
\begin{table}[htbp]
  \centering
  \small
  \caption{\textbf{Prompting w/ Level on Concept Steering.} Concept score (0--2) of Gemma-2-9B on \ttt{FLAS}'s held-out concepts at the lowest and highest levels, its change, and the mean per-concept Spearman correlation $\rho$ between level and concept score, judged by \ttt{Phi-4-14B} and \ttt{Qwen3.8-27B}.
  Prompting w/ level appends the level to \ttt{AxBench}'s prompt as a number out of 2 (0.5 to 2), out of 10 (1 to 10), or as a word (``Follow this instruction very slightly.'' to ``Follow this instruction extremely.'').
  \method{} uses $\alpha$ 0.5, 1, 1.5, and 2. }
  \label{tab:concept-level-prompt}
  \vspace{6pt}
  \setlength{\tabcolsep}{4pt}
  \begin{tabular}{lcccccccc}
    \toprule
    & \multicolumn{4}{c}{\ttt{Phi-4-14B}} & \multicolumn{4}{c}{\ttt{Qwen3.8-27B}} \\
    \cmidrule(lr){2-5} \cmidrule(lr){6-9}
    Method & Lowest & Highest & Change & $\rho$ & Lowest & Highest & Change & $\rho$ \\
    \midrule
    w/ Level (of 2)  & 1.11 & 1.26 & +0.15 & 0.29 & 1.40 & 1.60 & +0.20 & 0.32 \\
    w/ Level (of 10) & 1.17 & 1.32 & +0.15 & 0.25 & 1.52 & 1.62 & +0.10 & 0.21 \\
    w/ Level (verbal) & 1.15 & 1.36 & +0.21 & 0.40 & 1.49 & 1.69 & +0.20 & 0.31 \\
    \method{}        & 0.22 & 1.41 & +1.18 & 0.90 & 0.30 & 1.69 & +1.40 & 0.87 \\
    \bottomrule
  \end{tabular}
\end{table}

\textbf{Concept Steering Performance.}
On \ttt{Concept500} with requests 5--9, \method{} scores 1.059 at $\alpha{=}2$, above the shown single-layer per-concept methods (S-\ttt{PSR}, 0.900), despite training on none of the 500 concepts.
Prompt+ \method{} is best on every split.
It raises \ttt{AxBench}'s prompt from 1.049 to 1.198 ($+0.149$) on \ttt{Concept500} and exceeds prompting by $+0.158$ on held-out concepts.
The gain reflects higher concept incorporation with nearly unchanged instruction following.
The level remains effective with prompting.
At $\alpha{=}0.5$, prompt+\method{} lies between prompting and prompt+\method{} at $\alpha{=}1$ on every split.
On held-out concepts, it is $+0.051$ above prompting and $0.107$ below $\alpha{=}1$.

\textbf{Comparison with \ttt{FLAS}.}
\method{} alone scores above \ttt{FLAS}'s released checkpoint by $+0.062$ on \ttt{Concept500}, on which \ttt{FLAS} was trained, and by $+0.049$ on held-out concepts, and it matches prompting on both.
\method{} requires no labelled responses, whereas \ttt{FLAS} is trained on completions written by \ttt{gpt-4o-mini}.
\methodL{}, trained on 6.6$\times$ more instructions of every kind, retains most of this performance.
It scores above \ttt{FLAS} on \ttt{Concept500} ($+0.042$), matches it on held-out concepts, and lies 0.02--0.04 below \method{}.
Prompt+\methodL{} is within 0.03 of prompt+\method{}.

\textbf{Fine-Grained Control.}
On held-out concepts, the concept score of \method{} increases monotonically with the level, from 0.20 at $\alpha=0.5$, close to the unsteered LLM's 0.14, to 1.42 at $\alpha=2$ and 1.47 at $\alpha=2.5$, beyond the training range.
At $\alpha=1.5$, its concept score of 1.29 matches that of prompting (1.29).
Over levels 0.5--2, the concept score spans 1.21 points, compared with 0.66 for \ttt{FLAS}, whose weakest setting already reaches 0.60 and which remains at 1.18--1.26 from $T=1$ onward.
\method{} exceeds \ttt{FLAS} from $\alpha=1.375$ and is clearly higher on both metrics from $\alpha=1.625$ to 2, with a best overall score of 1.24 against 1.16.
At a matched concept score near 1.05, it preserves more of the request, with instruction scores of 1.96 against 1.88 and fluency scores of 1.46 against 1.33.
Prompting alone (1.31) achieves a higher overall score than \method{} alone, but the level remains effective on top of the prompt.
Prompt+\method{} spans 1.35--1.69 in concept score and 1.35--1.52 in overall score, above prompting and \ttt{FLAS} at every level.
CFG, the two-pass teacher, serves as an upper reference, but its overall score falls past $\alpha=1.625$, reaching 1.23 at $\alpha=2.5$, while \method{} remains at 1.24 at that same out-of-range adherence level (\cref{fig:level-dial,tab:level-dial-numbers}).

\textbf{Adherence Levels of Prompt + \method{}.}
\method{} is trained with the request alone as input, and Prompt + \method{} composes it with prompting without further training.
In Prompt + \method{}, $\alpha$ sets the strength of the edit added to the prompted LLM rather than an absolute adherence level.
If the edit acts as the CFG shift $\alpha(\log\pi_{\mathrm{prompted}}-\log\pi_{\mathrm{base}})$, adding it to the prompted LLM, which is at level 1, gives level $1+\alpha$.
Prompt + \method{} at $\alpha=0.5$, 1, and 1.5 reaches concept scores of 1.35, 1.58, and 1.67, close to CFG at levels 1.5, 2, and 2.5 (1.46, 1.58, and 1.64), with higher fluency from $\alpha=1$ (1.43 and 1.37 against 1.28 and 1.13).
On personas, Prompt + \methodL{} increases with $\alpha$ but stays below CFG at level $1+\alpha$ (\cref{fig:persona-level-all}).
\method{} alone thus covers levels below and above prompting, while Prompt + \method{} extends control above prompting.
Training \method{} with the prompt in its input would calibrate these levels, which we leave to future work.

\begin{figure}[t]
  \centering
  \includegraphics[width=0.9\linewidth]{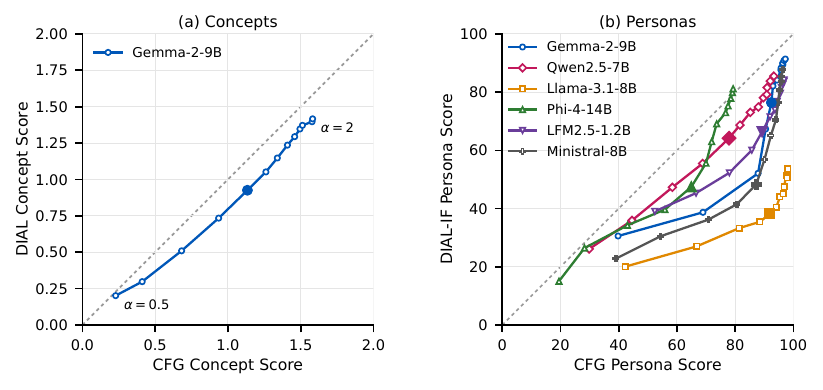}
  \caption{\textbf{Distillation Calibration.} Score of \method{} against that of CFG at the same adherence level $\alpha\in[0.5,2]$.
  The filled marker is $\alpha=1$, where CFG equals the LLM prompted with the training instruction template.
  Points on the dashed diagonal would match CFG.
  (a) \method{} on \ttt{AxBench} held-out concepts.
  (b) \methodL{} on seven held-out personas.}
  \label{fig:calibration}
\end{figure}

\textbf{Distillation Fidelity.}
\Cref{fig:calibration} plots the score of \method{} against that of CFG, which provides its training signal, at the same adherence level.
The score of \method{} increases with $\alpha$ wherever that of CFG does, so the order of levels is preserved, but it stays below CFG at a matched level.
On concepts, \method{} reaches 82--90\% of CFG's concept score from $\alpha=1$ to 2 (0.93 against 1.13 at $\alpha=1$) and reaches CFG's concept score at $\alpha=1$ at about $\alpha=1.25$.
On personas, the gap at $\alpha=1$ ranges from 13.7 points on Qwen2.5-7B to 53.7 on Llama-3.1-8B, whose sampling we discuss in \cref{app:llama-topk}.
The adherence level of \method{} is thus monotone but compressed relative to that of CFG.
A likely cause is capacity: CFG conditions every layer on the instruction, while \method{} edits one layer.
Editing four layers instead of one adds 0.057 on Gemma-2-2B (\cref{app:abl-layers}), and closing the gap with larger adapters is future work.

\begin{figure}[t]
  \centering
  \includegraphics[width=\linewidth]{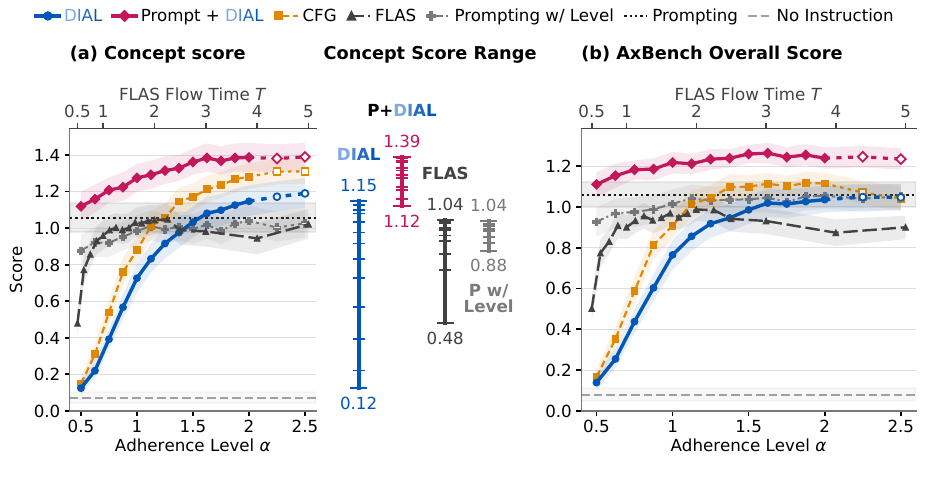}
  \caption{\textbf{Fine-Grained Concept Expression with \ttt{gpt-4o-mini}.} Scores are lower than under the three-judge aggregation in \cref{fig:level-dial}, while the method ordering remains the same.}
  \label{fig:level-dial-gpt}
\end{figure}

\textbf{Judge Robustness.}\label{app:judges}
\Cref{fig:level-dial} uses the median of three judges' ratings.
We also report the full curves and scores under \ttt{gpt-4o-mini}, the judge used in \cref{tab:main-axb-9b} (\cref{fig:level-dial-gpt,tab:level-dial-numbers-gpt}).
\ttt{gpt-4o-mini} scores the main comparison tables.
To limit API cost, the level curves, ablations, and persona results use local judges (\ttt{Phi-4-14B}, \ttt{Qwen3.8-27B}, and \ttt{Llama-3.3-70B}), with \ttt{gpt-4o-mini} added to the median for the concept level curves.
Therefore, scores are comparable within a table or figure but not comparable across them.

\begin{table}[htbp]
  \centering\small
  \caption{\textbf{Fine-Grained Concept Expression Levels with \ttt{gpt-4o-mini}.} Scores plotted in \cref{fig:level-dial-gpt} on \ttt{FLAS}'s 100 held-out concepts, using ten 128-token responses per concept.}
  \label{tab:level-dial-numbers-gpt}
  \vspace{6pt}
  \begin{minipage}[t]{0.48\linewidth}
    \centering\small
    \setlength{\tabcolsep}{1.5pt}\renewcommand{\arraystretch}{1.05}
    \resizebox{\linewidth}{!}{%
    \begin{tabular}{@{}ccccc@{}}
      \toprule
      Level & Concept & Request & Fluency & Overall score \\
      \midrule
      \multicolumn{5}{@{}l}{\method{} ($\alpha$)} \\
      0.5 & 0.124 & 1.941 & 1.270 & 0.138 \\
      0.625 & 0.220 & 1.952 & 1.263 & 0.255 \\
      0.75 & 0.392 & 1.931 & 1.247 & 0.437 \\
      0.875 & 0.567 & 1.934 & 1.220 & 0.603 \\
      1 & 0.726 & 1.907 & 1.201 & 0.765 \\
      1.125 & 0.832 & 1.880 & 1.199 & 0.856 \\
      1.25 & 0.916 & 1.855 & 1.181 & 0.918 \\
      1.375 & 0.976 & 1.841 & 1.162 & 0.946 \\
      1.5 & 1.033 & 1.837 & 1.152 & 0.986 \\
      1.625 & 1.081 & 1.821 & 1.154 & 1.019 \\
      1.75 & 1.098 & 1.801 & 1.119 & 1.014 \\
      1.875 & 1.126 & 1.795 & 1.104 & 1.027 \\
      2 & 1.147 & 1.786 & 1.091 & 1.036 \\
      \midrule
      \multicolumn{5}{@{}l}{prompt+ \method{} ($\alpha$)} \\
      0.5 & 1.119 & 1.858 & 1.130 & 1.111 \\
      0.625 & 1.159 & 1.868 & 1.133 & 1.153 \\
      0.75 & 1.208 & 1.842 & 1.141 & 1.182 \\
      0.875 & 1.224 & 1.850 & 1.136 & 1.186 \\
      1 & 1.275 & 1.852 & 1.128 & 1.218 \\
      1.125 & 1.292 & 1.826 & 1.130 & 1.212 \\
      1.25 & 1.315 & 1.823 & 1.135 & 1.234 \\
      1.375 & 1.328 & 1.812 & 1.124 & 1.239 \\
      1.5 & 1.362 & 1.809 & 1.126 & 1.260 \\
      1.625 & 1.384 & 1.793 & 1.137 & 1.264 \\
      1.75 & 1.368 & 1.796 & 1.115 & 1.244 \\
      1.875 & 1.384 & 1.800 & 1.139 & 1.255 \\
      2 & 1.387 & 1.781 & 1.124 & 1.239 \\
      \bottomrule
    \end{tabular}}
  \end{minipage}\hfill
  \begin{minipage}[t]{0.48\linewidth}
    \centering\small
    \setlength{\tabcolsep}{1.5pt}\renewcommand{\arraystretch}{1.05}
    \resizebox{\linewidth}{!}{%
    \begin{tabular}{@{}ccccc@{}}
      \toprule
      Level & Concept & Request & Fluency & Overall score \\
      \midrule
      \multicolumn{5}{@{}l}{CFG ($\alpha$)} \\
      0.5 & 0.149 & 1.946 & 1.259 & 0.166 \\
      0.625 & 0.309 & 1.940 & 1.278 & 0.353 \\
      0.75 & 0.538 & 1.913 & 1.279 & 0.587 \\
      0.875 & 0.759 & 1.896 & 1.302 & 0.813 \\
      1 & 0.880 & 1.880 & 1.284 & 0.908 \\
      1.125 & 1.011 & 1.857 & 1.290 & 1.015 \\
      1.25 & 1.056 & 1.858 & 1.256 & 1.044 \\
      1.375 & 1.146 & 1.829 & 1.262 & 1.097 \\
      1.5 & 1.170 & 1.815 & 1.232 & 1.099 \\
      1.625 & 1.212 & 1.802 & 1.192 & 1.114 \\
      1.75 & 1.235 & 1.780 & 1.159 & 1.101 \\
      1.875 & 1.267 & 1.763 & 1.152 & 1.118 \\
      2 & 1.282 & 1.785 & 1.108 & 1.115 \\
      \midrule
      \multicolumn{5}{@{}l}{\ttt{FLAS} ($T$)} \\
      0.5 & 0.479 & 1.851 & 1.205 & 0.501 \\
      0.625 & 0.772 & 1.767 & 1.172 & 0.773 \\
      0.75 & 0.857 & 1.742 & 1.177 & 0.831 \\
      0.875 & 0.930 & 1.753 & 1.143 & 0.908 \\
      1 & 0.959 & 1.700 & 1.136 & 0.900 \\
      1.125 & 0.989 & 1.742 & 1.111 & 0.932 \\
      1.25 & 1.003 & 1.711 & 1.156 & 0.955 \\
      1.375 & 0.991 & 1.734 & 1.141 & 0.935 \\
      1.5 & 1.001 & 1.732 & 1.136 & 0.947 \\
      1.625 & 1.031 & 1.719 & 1.147 & 0.969 \\
      1.75 & 1.027 & 1.732 & 1.118 & 0.949 \\
      1.875 & 1.040 & 1.712 & 1.104 & 0.968 \\
      2 & 1.044 & 1.738 & 1.106 & 0.987 \\
      \bottomrule
    \end{tabular}}
  \end{minipage}
\end{table}

\textbf{Gemma-2-2B.}\label{app:axb-2b}
\Cref{tab:axb-2b} repeats \cref{tab:main-axb-9b} on Gemma-2-2B.
Here \method{} alone ties prompting but scores below \ttt{FLAS}'s released 2B checkpoint ($-0.185$ on \ttt{Concept500}, $-0.112$ held-out).
As for Gemma-2-9B, \ttt{HyperSteer}'s held-out score uses its own held-out concepts.
Prompting itself performs poorly on Gemma-2-2B.
Since \method{} self-distills CFG from the same LLM's prompted and unprompted distributions, weaker prompted instruction following may limit the concept expression learned during training.
Prompt+\method{} is again the best row: $+0.231$ and $+0.245$ above prompting, and $+0.112$ above \ttt{FLAS} on held-out concepts ($+0.044$ on \ttt{Concept500}, which \ttt{FLAS} trains on).

\begin{table}[htbp]
  \centering\small
  \setlength{\tabcolsep}{4pt}
  \caption{\textbf{Concept Steering on \ttt{AxBench} with Gemma-2-2B.} Overall scores with \ttt{gpt-4o-mini} as judge.
  Held-In: \ttt{Concept500}; Held-Out: \ttt{FLAS}'s 100 held-out concepts.
  \expPub\ denotes published results with a level selected per concept.
  Our results use one fixed level.
  \ttt{FLAS}-2B trains on 478 of the 500 Gemma-2-2B \ttt{Concept500} concepts (its held-out split differs from that of the 9B checkpoint), whereas \method{} trains on none of its concepts.
  \method{} steers layer 12 and the other methods steer layer 20.
  Under its 256-token protocol, \ttt{FLAS} reports 1.025 and 1.015 on its own held-in and held-out splits.
  The layout follows \cref{tab:main-axb-9b}.}
  \label{tab:axb-2b}
  \vspace{6pt}
  \begin{tabular}{lcc}
    \toprule
    Method & Held-in & Held-out \\
    \midrule
    Prompting & 0.700 & 0.697 \\
    \midrule
    \multicolumn{3}{l}{\textit{One model per concept}} \\
    \ttt{LoReFT}~\citep{wu2025axbench} & 0.722\expPub & -- \\
    \ttt{LoRA}~\citep{wu2025axbench} & 0.641\expPub & -- \\
    S-\ttt{PSR}~\citep{heyman2026psr} & 0.367\expPub & -- \\
    SV + \ttt{RePS}~\citep{wu2025reps} & 0.606\expPub & -- \\
    \ttt{ReFT-r1}~\citep{wu2025axbench} & 0.509\expPub & -- \\
    \ttt{DiffMean}~\citep{wu2025axbench} & 0.178\expPub & -- \\
    \ttt{SAE}~\citep{wu2025axbench} & 0.151\expPub & -- \\
    \midrule
    \multicolumn{3}{l}{\textit{One model for all concepts}} \\
    \ttt{HyperSteer}~\citep{sun2025hypersteer} & 0.742\expPub & 0.608\expPub \\
    \ttt{FLAS}~\citep{jin2026flas} ($T{=}2$) & 0.887 & 0.830 \\
    \midrule
    \method{} ($\alpha{=}2$) & 0.702 & 0.718 \\
    Prompt + \method{} ($\alpha{=}1$) & \textbf{0.931} & \textbf{0.942} \\
    \bottomrule
  \end{tabular}
\end{table}

\subsection{Controlling Persona Expression Levels}
\label{app:persona}
\textbf{Evaluation Setup.}
We follow the protocol in~\citep{chen2025persona} for seven released personas: evil, sycophantic, hallucinating, apathetic, humorous, impolite, and optimistic.
Neither these personas nor their near-synonyms appears in \ttt{SteerIF}.
Each persona has 20 questions.
% \del{The curves in \mbox{\cref{fig:persona-dial}} use four samples for every LLM, while the corrected Gemma-2-9B baseline references use ten.}
\Cref{fig:persona-dial} uses four samples per question for every curve and baseline, and the tables use ten, except for Llama-3.1-8B, which uses four.
The figure's four samples are the first four of the tables' ten, so values at the same level differ slightly between them, e.g., 76.4 in \cref{fig:persona-dial} and 75.8 in \cref{tab:persona} for Gemma-2-9B with \methodL{} at $\alpha{=}1$.
We sample each response at temperature 1 for up to 1{,}000 new tokens.

\ttt{Phi-4-14B}, \ttt{Qwen3.8-27B}, and \ttt{Llama-3.3-70B}~\citep{grattafiori2024llama3} score persona expression and coherence from 0 to 100 using the rubrics in~\citep{chen2025persona}.
Each score is the median of the three probability-weighted ratings.
We skip a judge's refusal or off-format answer.
User prompting places the persona description before the question in the user message.
System prompting places it in the system message and is reported for Qwen2.5-7B, Phi-4-14B, and LFM2.5-1.2B.
Gemma-2 has no system role, and Ministral-8B's chat template places a system prompt before the question in the user message, so system and user prompting coincide.
For \methodL{} alone, the prompt contains only the question and the adapter receives the persona description.
Llama-3.1-8B uses 1{,}024-token persona rollouts during training, covering the 1{,}000-token evaluation horizon.

\begin{figure}[H]
  \centering
  \includegraphics[width=\linewidth]{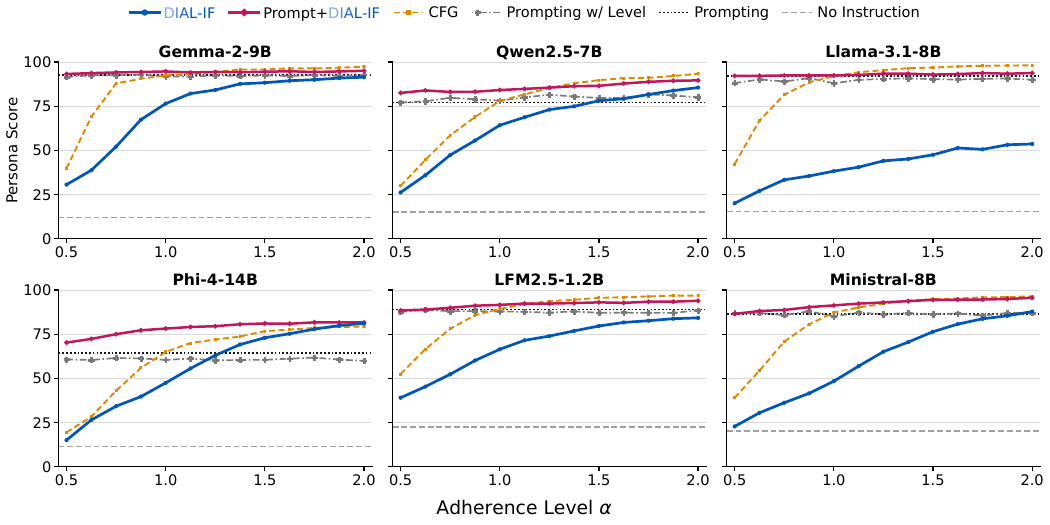}
  \caption{\textbf{Fine-Grained Persona Expression across Six LLMs.} \methodL{} controls intermediate persona expression levels, while prompt+\methodL{} further raises persona scores.}
  \label{fig:persona-level-all}
\end{figure}

\Cref{tab:persona,tab:persona-qwen,tab:persona-llama,tab:persona-phi4,tab:persona-lfm,tab:persona-ministral} give the full persona grids behind \cref{fig:persona-dial} and \cref{fig:persona-level-all}, for each persona and adherence level.
As the adherence level increases, the persona score generally rises while coherence falls.
On Llama-3.1-8B, \methodL{} alone reaches a mean persona score of 53.6 at $\alpha{=}2$, below prompting's 91.9, with coherence 69.8.
On top of the prompt, the persona score changes little with the level (92.1, 92.4, 92.9, 93.7 at $\alpha{=}0.5$, 1, 1.5, 2), while coherence falls from 84.4 at $\alpha{=}0.5$ to 58.1 at $\alpha{=}2$.

\textbf{Fine-Grained Persona Expression.}
Across the four tabulated adherence levels, the persona score of \methodL{} alone rises monotonically on every LLM ( \cref{fig:persona-level-all}).
For seven personas held out from training, \methodL{} raises Gemma's mean persona score from 12.1 without steering to 75.8 at $\alpha{=}1$ and 90.9 at $\alpha{=}2$, close to prompting's 92.6.
With the prompt, it reaches 93.3, 94.2, and 94.8 at $\alpha{=}0.5$, 1, and 2.
Higher adherence levels reduce coherence to 68.1 for prompt+\methodL{} at $\alpha{=}2$ against 84.2 for prompting (\cref{tab:persona}).
On Qwen2.5-7B, prompt+\methodL{} raises the persona score from 76.9 to 82.5, 84.9, and 89.8 at $\alpha{=}0.5$, 1, and 2.
On Llama-3.1-8B, it raises the score from 91.9 to 92.1 and 92.4 at $\alpha{=}0.5$ and 1.
On Phi-4-14B, prompt+\methodL{} raises the persona score from 64.9 to 70.0 and 78.2 at $\alpha{=}0.5$ and 1.
Phi-4-14B refuses the evil persona under prompting (0.0), whereas \methodL{} alone reaches 24.4 at $\alpha{=}2$ (\cref{tab:persona-phi4}).
On LFM2.5, prompting already reaches 88.9, so prompt+\methodL{} changes it little (88.4 and 91.7), while \methodL{} alone raises the score from 21.5 without steering to 84.9 at $\alpha{=}2$ on these held-out personas (\cref{tab:persona-lfm}).
On Ministral-8B, \methodL{} alone raises the score from 20.2 without steering to 86.7 at $\alpha{=}2$, matching prompting's 86.9, while coherence falls to 44.7 against 78.2 for prompting (\cref{tab:persona-ministral}).

\textbf{Prompting w/ Level.}
We append the same level sentence as for concepts (\cref{app:results-concept}) to the user-prefix persona prompt, with the level written out of 2 (0.5 to 2) or out of 10 (1 to 10).
\Cref{tab:persona-level-prompt} shows that the stated level barely changes persona expression.
On every LLM, prompting w/ level stays within 5.4 points of prompting at the tabulated levels, and its persona score changes by at most 6.6 points between the lowest and the highest level, whereas \methodL{} spans 34--66 points from $\alpha{=}0.5$ to 2.
Even at the lowest level, the persona is expressed at about prompting's strength.
\begin{table}[htbp]
  \centering
  \small
  \caption{\textbf{Prompting w/ Level on Persona Steering.} Mean persona score (0--100) over the seven held-out personas, judged by the three-judge jury (\ttt{Phi-4-14B}, \ttt{Qwen3.8-27B}, and \ttt{Llama-3.3-70B}).
  Prompting w/ level appends the adherence level to the user-prefix persona prompt, out of 2 or out of 10.
  We report the lowest and highest levels of each method.}
  \label{tab:persona-level-prompt}
  \vspace{6pt}
  \setlength{\tabcolsep}{5pt}
  \begin{tabular}{lccccccc}
    \toprule
    & & \multicolumn{2}{c}{w/ Level (of 2)} & \multicolumn{2}{c}{w/ Level (of 10)} & \multicolumn{2}{c}{\methodL{}} \\
    \cmidrule(lr){3-4} \cmidrule(lr){5-6} \cmidrule(lr){7-8}
    LLM & Prompting & 0.5 & 2 & 1 & 10 & $\alpha{=}0.5$ & $\alpha{=}2$ \\
    \midrule
    Gemma-2-9B   & 92.6 & 91.0 & 92.8 & 90.9 & 93.8 & 30.6 & 90.9 \\
    Llama-3.1-8B & 91.9 & 88.1 & 89.9 & 86.5 & 93.1 & 20.0 & 53.6 \\
    Qwen2.5-7B   & 76.9 & 77.3 & 80.0 & 80.6 & 80.9 & 25.7 & 84.8 \\
    Phi-4-14B    & 64.9 & 60.6 & 59.8 & 60.6 & 61.7 & 15.3 & 80.8 \\
    LFM2.5-1.2B  & 88.9 & 88.1 & 88.1 & 89.4 & 91.1 & 38.7 & 84.9 \\
    Ministral-8B & 86.9 & 87.6 & 87.0 & 87.4 & 87.9 & 22.8 & 86.7 \\
    \bottomrule
  \end{tabular}
\end{table}

\textbf{Sampling on Llama-3.1-8B.}\label{app:llama-topk}
Under full-vocabulary sampling at temperature 1, \methodL{} produces low-coherence responses on Llama-3.1-8B (\cref{tab:llama-topk}).
One possible cause is limited supervision of the low-probability tail.
The training loss uses the union of the top-64 token sets from the base, prompted, and student distributions, rather than the full vocabulary.
This may leave the tail poorly supervised and reduce coherence when inference samples from the full distribution.

At $\alpha=1$, 41.6\% of \methodL{}-only responses are degenerate and 36.1\% reach the 1{,}000-token limit, compared with 5.2\% and 2.5\% for prompting.
Restricting sampling to the 64 most likely tokens removes the observed degeneration.
We therefore use top-64 sampling for every Llama-3.1-8B persona method, including prompting, CFG, and no instruction.
The other five LLMs have degeneration rates of at most 0.1\% under full-vocabulary sampling.
We classify a response as degenerate if any 40-word window contains fewer than 12\% function words and at most two sentence or list markers, using the same criterion for every LLM.
\begin{table}[H]
\centering
\small
\caption{\textbf{Sampling and Response Coherence on Llama-3.1-8B.} Degenerate-response rate and share of responses reaching the 1{,}000-token limit, using full-vocabulary or top-64 sampling.}
\label{tab:llama-topk}
\vspace{6pt}
\begin{tabular}{llcccc}
\toprule
 & & \multicolumn{2}{c}{Full vocabulary} & \multicolumn{2}{c}{Top-64} \\
\cmidrule(lr){3-4}\cmidrule(lr){5-6}
Method & Level & Deg. (\%) & Cap (\%) & Deg. (\%) & Cap (\%) \\
\midrule
\methodL{} only & $\alpha{=}1$ & 41.6 & 36.1 & 0.0 & 0.0 \\
\methodL{} only & $\alpha{=}2$ & 42.0 & 35.7 & 0.0 & 0.4 \\
Prompt + \methodL{} & $\alpha{=}1$ & 26.8 & 15.2 & 0.0 & 0.0 \\
Prompt + \methodL{} & $\alpha{=}2$ & 28.8 & 15.7 & 0.0 & 0.0 \\
CFG & $\alpha{=}1$ & 3.4 & 2.9 & 0.0 & 0.0 \\
CFG & $\alpha{=}2$ & 0.5 & 0.0 & 0.0 & 0.0 \\
Prompting & -- & 5.2 & 2.5 & 0.0 & 0.0 \\
No steering & -- & 2.9 & 2.9 & 0.0 & 0.0 \\
\bottomrule
\end{tabular}

\end{table}

\begin{table}[H]
\centering
\small
\setlength{\tabcolsep}{3.5pt}
\caption{\textbf{Persona Steering on Gemma-2-9B.} Persona scores and mean coherence (0--100), averaged over 20 questions $\times$ 10 samples per persona.
Each response is scored by the median of \ttt{Phi-4-14B}, \ttt{Qwen3.8-27B}, and \ttt{Llama-3.3-70B}.}
\label{tab:persona}
\vspace{6pt}
\resizebox{\linewidth}{!}{%
\begin{tabular}{lccccccccc}
\toprule
 & \multicolumn{8}{c}{Persona score $\uparrow$} & Coh. $\uparrow$ \\
\cmidrule(lr){2-9}
Method & Evil & Sycoph. & Halluc. & Apathe. & Humoro. & Impoli. & Optimi. & Mean & Mean \\
\midrule
No steering & 0.0 & 6.2 & 4.0 & 0.7 & 2.7 & 0.0 & 71.2 & 12.1 & 98.3 \\
Prompting & 99.5 & 93.8 & 96.4 & 87.0 & 95.2 & 78.6 & 97.6 & 92.6 & 84.2 \\
\midrule
\multicolumn{10}{l}{\textit{\methodL{}}} \\
\methodL{} only, $\alpha{=}0.5$ & 0.2 & 18.8 & 25.2 & 3.1 & 78.3 & 1.7 & 87.0 & 30.6 & 96.3 \\
\methodL{} only, $\alpha{=}1$ & 82.7 & 85.0 & 80.4 & 24.9 & 95.8 & 64.2 & 97.3 & 75.8 & 85.9 \\
\methodL{} only, $\alpha{=}1.5$ & 99.0 & 94.2 & 93.4 & 62.6 & 97.6 & 69.3 & 99.0 & 87.9 & 79.0 \\
\methodL{} only, $\alpha{=}2$ & 99.8 & 95.2 & 96.4 & 75.7 & 98.2 & 72.0 & 99.4 & 90.9 & 73.9 \\
Prompt + \methodL{}, $\alpha{=}0.5$ & 99.6 & 95.5 & 97.2 & 86.4 & 95.7 & 80.5 & 98.1 & 93.3 & 82.4 \\
Prompt + \methodL{}, $\alpha{=}1$ & 99.6 & 97.8 & 97.9 & 86.8 & 96.8 & 81.9 & 98.8 & 94.2 & 78.5 \\
Prompt + \methodL{}, $\alpha{=}1.5$ & 99.8 & 98.7 & 98.6 & 86.8 & 97.4 & 80.7 & 99.3 & 94.5 & 73.3 \\
Prompt + \methodL{}, $\alpha{=}2$ & 99.8 & 99.2 & 99.3 & 87.3 & 97.8 & 80.8 & 99.4 & 94.8 & 68.1 \\
\bottomrule
\end{tabular}}

\end{table}

\begin{table}[H]
\centering
\small
\setlength{\tabcolsep}{3.5pt}
\caption{\textbf{Persona Steering on Qwen2.5-7B.} Persona scores and mean coherence (0--100), averaged over 20 questions $\times$ 10 samples per persona.}
\label{tab:persona-qwen}
\vspace{6pt}
\resizebox{\linewidth}{!}{%
\begin{tabular}{lccccccccc}
\toprule
 & \multicolumn{8}{c}{Persona score $\uparrow$} & Coh. $\uparrow$ \\
\cmidrule(lr){2-9}
Method & Evil & Sycoph. & Halluc. & Apathe. & Humoro. & Impoli. & Optimi. & Mean & Mean \\
\midrule
No steering & 0.0 & 13.9 & 13.2 & 4.4 & 0.1 & 0.0 & 74.6 & 15.2 & 97.5 \\
Prompting (system turn) & 92.5 & 70.3 & 76.1 & 54.8 & 90.8 & 22.5 & 94.1 & 71.6 & 90.5 \\
Prompting (user prefix) & 90.9 & 70.8 & 79.3 & 69.6 & 92.8 & 39.0 & 96.0 & 76.9 & 89.2 \\
\midrule
\multicolumn{10}{l}{\textit{\methodL{}}} \\
\methodL{} only, $\alpha{=}0.5$ & 0.1 & 19.9 & 12.1 & 11.1 & 46.5 & 0.1 & 90.2 & 25.7 & 96.9 \\
\methodL{} only, $\alpha{=}1$ & 87.8 & 54.8 & 18.6 & 58.0 & 95.0 & 39.5 & 97.0 & 64.4 & 86.7 \\
\methodL{} only, $\alpha{=}1.5$ & 97.8 & 77.3 & 31.9 & 77.1 & 95.8 & 68.1 & 99.2 & 78.2 & 78.4 \\
\methodL{} only, $\alpha{=}2$ & 98.1 & 87.0 & 46.2 & 83.7 & 95.5 & 83.3 & 99.7 & 84.8 & 69.5 \\
Prompt (system) + \methodL{}, $\alpha{=}0.5$ & 96.0 & 77.2 & 79.3 & 66.8 & 92.2 & 27.0 & 96.8 & 76.5 & 89.2 \\
Prompt (system) + \methodL{}, $\alpha{=}1$ & 98.5 & 81.4 & 85.9 & 79.2 & 92.8 & 24.9 & 98.3 & 80.1 & 87.2 \\
Prompt (system) + \methodL{}, $\alpha{=}1.5$ & 98.9 & 82.4 & 87.7 & 82.0 & 91.1 & 31.3 & 99.1 & 81.8 & 83.8 \\
Prompt (system) + \methodL{}, $\alpha{=}2$ & 98.5 & 84.3 & 86.2 & 83.2 & 89.1 & 38.5 & 99.5 & 82.7 & 80.6 \\
Prompt (user prefix) + \methodL{}, $\alpha{=}0.5$ & 98.5 & 84.0 & 82.0 & 72.6 & 93.1 & 49.5 & 97.6 & 82.5 & 87.0 \\
Prompt (user prefix) + \methodL{}, $\alpha{=}1$ & 99.4 & 89.7 & 85.5 & 81.2 & 94.7 & 45.0 & 99.0 & 84.9 & 83.8 \\
Prompt (user prefix) + \methodL{}, $\alpha{=}1.5$ & 99.8 & 92.4 & 90.5 & 83.9 & 94.5 & 51.7 & 99.8 & 87.5 & 80.6 \\
Prompt (user prefix) + \methodL{}, $\alpha{=}2$ & 99.7 & 94.6 & 92.2 & 86.8 & 93.8 & 61.8 & 99.9 & 89.8 & 74.7 \\
\bottomrule
\end{tabular}}

\end{table}

\begin{table}[H]
\centering
\small
\setlength{\tabcolsep}{3.5pt}
\caption{\textbf{Persona Steering on Llama-3.1-8B.} Persona scores and mean coherence (0--100), averaged over 20 questions $\times$ 4 samples per persona.
Every method samples from the top 64 tokens (\cref{app:llama-topk}).}
\label{tab:persona-llama}
\vspace{6pt}
\resizebox{\linewidth}{!}{%
\begin{tabular}{lccccccccc}
\toprule
 & \multicolumn{8}{c}{Persona score $\uparrow$} & Coh. $\uparrow$ \\
\cmidrule(lr){2-9}
Method & Evil & Sycoph. & Halluc. & Apathe. & Humoro. & Impoli. & Optimi. & Mean & Mean \\
\midrule
No steering & 0.0 & 16.7 & 15.2 & 3.8 & 0.1 & 0.0 & 72.6 & 15.5 & 96.3 \\
Prompting (user prefix) & 99.7 & 94.1 & 98.9 & 80.4 & 95.9 & 75.2 & 99.2 & 91.9 & 86.4 \\
\midrule
\multicolumn{10}{l}{\textit{\methodL{}}} \\
\methodL{} only, $\alpha{=}0.5$ & 0.0 & 20.2 & 24.2 & 4.6 & 9.3 & 0.0 & 81.9 & 20.0 & 94.3 \\
\methodL{} only, $\alpha{=}1$ & 10.4 & 33.3 & 29.0 & 8.3 & 91.8 & 0.7 & 94.0 & 38.2 & 86.2 \\
\methodL{} only, $\alpha{=}1.5$ & 32.4 & 36.6 & 32.0 & 16.9 & 96.2 & 20.6 & 97.5 & 47.5 & 76.7 \\
\methodL{} only, $\alpha{=}2$ & 50.3 & 33.9 & 33.0 & 29.5 & 97.0 & 34.0 & 97.4 & 53.6 & 69.8 \\
Prompt (user prefix) + \methodL{}, $\alpha{=}0.5$ & 99.7 & 91.8 & 98.1 & 79.4 & 96.7 & 79.7 & 99.2 & 92.1 & 84.4 \\
Prompt (user prefix) + \methodL{}, $\alpha{=}1$ & 99.6 & 89.9 & 98.8 & 85.4 & 97.4 & 75.9 & 99.7 & 92.4 & 75.7 \\
Prompt (user prefix) + \methodL{}, $\alpha{=}1.5$ & 99.7 & 91.2 & 98.0 & 88.8 & 97.7 & 74.8 & 99.8 & 92.9 & 64.4 \\
Prompt (user prefix) + \methodL{}, $\alpha{=}2$ & 99.7 & 88.0 & 97.8 & 91.2 & 97.9 & 81.8 & 99.8 & 93.7 & 58.1 \\
\bottomrule
\end{tabular}}

\end{table}

\begin{table}[H]
\centering
\small
\setlength{\tabcolsep}{3.5pt}
\caption{\textbf{Persona Steering on Phi-4-14B.} Persona scores and mean coherence (0--100), averaged over 20 questions $\times$ 10 samples per persona.}
\label{tab:persona-phi4}
\vspace{6pt}
\resizebox{\linewidth}{!}{%
\begin{tabular}{lccccccccc}
\toprule
 & \multicolumn{8}{c}{Persona score $\uparrow$} & Coh. $\uparrow$ \\
\cmidrule(lr){2-9}
Method & Evil & Sycoph. & Halluc. & Apathe. & Humoro. & Impoli. & Optimi. & Mean & Mean \\
\midrule
No steering & 0.0 & 4.3 & 4.5 & 3.2 & 0.1 & 0.0 & 67.9 & 11.4 & 98.2 \\
Prompting (system turn) & 0.0 & 86.7 & 93.9 & 53.2 & 96.1 & 51.7 & 97.7 & 68.5 & 91.8 \\
Prompting (user prefix) & 0.0 & 84.0 & 81.0 & 53.6 & 96.3 & 42.0 & 97.7 & 64.9 & 91.5 \\
\midrule
\multicolumn{10}{l}{\textit{\methodL{}}} \\
\methodL{} only, $\alpha{=}0.5$ & 0.0 & 6.5 & 3.9 & 4.0 & 10.1 & 0.0 & 82.6 & 15.3 & 97.8 \\
\methodL{} only, $\alpha{=}1$ & 3.4 & 41.0 & 24.4 & 13.7 & 97.0 & 49.0 & 97.2 & 46.5 & 90.4 \\
\methodL{} only, $\alpha{=}1.5$ & 13.4 & 84.8 & 75.6 & 50.2 & 98.0 & 83.5 & 99.4 & 72.1 & 79.6 \\
\methodL{} only, $\alpha{=}2$ & 24.4 & 91.6 & 88.8 & 75.3 & 98.3 & 87.7 & 99.8 & 80.8 & 72.4 \\
Prompt (system) + \methodL{}, $\alpha{=}0.5$ & 0.0 & 87.3 & 94.1 & 60.4 & 96.3 & 55.1 & 97.9 & 70.2 & 91.2 \\
Prompt (system) + \methodL{}, $\alpha{=}1$ & 0.4 & 92.8 & 97.5 & 75.5 & 97.2 & 61.8 & 98.8 & 74.8 & 88.0 \\
Prompt (system) + \methodL{}, $\alpha{=}1.5$ & 0.0 & 95.9 & 98.7 & 80.4 & 97.6 & 66.7 & 99.4 & 77.0 & 84.7 \\
Prompt (system) + \methodL{}, $\alpha{=}2$ & 0.0 & 97.4 & 99.4 & 83.3 & 98.0 & 69.9 & 99.5 & 78.2 & 80.8 \\
Prompt (user prefix) + \methodL{}, $\alpha{=}0.5$ & 0.0 & 87.3 & 84.4 & 64.6 & 96.9 & 58.0 & 98.5 & 70.0 & 88.7 \\
Prompt (user prefix) + \methodL{}, $\alpha{=}1$ & 0.0 & 95.7 & 96.8 & 80.6 & 97.8 & 77.1 & 99.5 & 78.2 & 80.6 \\
Prompt (user prefix) + \methodL{}, $\alpha{=}1.5$ & 0.1 & 98.5 & 99.0 & 85.6 & 98.2 & 83.7 & 99.9 & 80.7 & 73.0 \\
Prompt (user prefix) + \methodL{}, $\alpha{=}2$ & 0.4 & 99.4 & 98.5 & 88.7 & 98.4 & 85.3 & 99.9 & 81.5 & 67.5 \\
\bottomrule
\end{tabular}}

\end{table}

\begin{table}[H]
\centering
\small
\setlength{\tabcolsep}{3.5pt}
\caption{\textbf{Persona Steering on LFM2.5-1.2B.} Persona scores and mean coherence (0--100), averaged over 20 questions $\times$ 10 samples per persona.}
\label{tab:persona-lfm}
\vspace{6pt}
\resizebox{\linewidth}{!}{%
\begin{tabular}{lccccccccc}
\toprule
 & \multicolumn{8}{c}{Persona score $\uparrow$} & Coh. $\uparrow$ \\
\cmidrule(lr){2-9}
Method & Evil & Sycoph. & Halluc. & Apathe. & Humoro. & Impoli. & Optimi. & Mean & Mean \\
\midrule
No steering & 0.0 & 22.8 & 48.0 & 0.7 & 0.8 & 0.0 & 78.0 & 21.5 & 92.7 \\
Prompting (system turn) & 94.3 & 79.1 & 72.1 & 48.4 & 94.8 & 47.4 & 99.7 & 76.5 & 77.1 \\
Prompting (user prefix) & 98.0 & 86.6 & 94.6 & 73.2 & 96.1 & 73.8 & 99.8 & 88.9 & 68.4 \\
\midrule
\multicolumn{10}{l}{\textit{\methodL{}}} \\
\methodL{} only, $\alpha{=}0.5$ & 3.7 & 27.2 & 54.3 & 5.5 & 82.1 & 3.9 & 94.6 & 38.7 & 88.7 \\
\methodL{} only, $\alpha{=}1$ & 62.7 & 58.7 & 76.2 & 33.6 & 95.1 & 31.7 & 99.3 & 65.3 & 74.0 \\
\methodL{} only, $\alpha{=}1.5$ & 77.8 & 78.3 & 82.6 & 61.9 & 96.5 & 60.4 & 99.8 & 79.6 & 63.0 \\
\methodL{} only, $\alpha{=}2$ & 78.5 & 84.7 & 84.0 & 74.7 & 97.3 & 75.2 & 99.8 & 84.9 & 54.4 \\
Prompt (system) + \methodL{}, $\alpha{=}0.5$ & 92.6 & 74.2 & 71.1 & 62.0 & 95.3 & 47.4 & 99.6 & 77.5 & 74.6 \\
Prompt (system) + \methodL{}, $\alpha{=}1$ & 94.1 & 82.8 & 87.5 & 80.1 & 96.6 & 58.3 & 99.8 & 85.6 & 67.3 \\
Prompt (system) + \methodL{}, $\alpha{=}1.5$ & 95.7 & 87.8 & 90.1 & 85.4 & 97.3 & 64.3 & 99.9 & 88.6 & 60.6 \\
Prompt (system) + \methodL{}, $\alpha{=}2$ & 95.8 & 91.2 & 93.3 & 86.4 & 97.4 & 68.5 & 99.8 & 90.4 & 55.1 \\
Prompt (user prefix) + \methodL{}, $\alpha{=}0.5$ & 95.6 & 87.2 & 91.5 & 80.7 & 96.2 & 67.8 & 99.8 & 88.4 & 65.6 \\
Prompt (user prefix) + \methodL{}, $\alpha{=}1$ & 98.2 & 91.0 & 94.4 & 88.5 & 97.2 & 72.9 & 99.9 & 91.7 & 57.1 \\
Prompt (user prefix) + \methodL{}, $\alpha{=}1.5$ & 98.3 & 94.6 & 94.7 & 89.8 & 97.5 & 76.3 & 99.9 & 93.0 & 51.0 \\
Prompt (user prefix) + \methodL{}, $\alpha{=}2$ & 98.0 & 95.5 & 94.4 & 91.1 & 97.7 & 80.8 & 99.8 & 93.9 & 47.6 \\
\bottomrule
\end{tabular}}

\end{table}

\begin{table}[H]
\centering
\small
\setlength{\tabcolsep}{3.5pt}
\caption{\textbf{Persona Steering on Ministral-8B.} Persona scores and mean coherence (0--100), averaged over 20 questions $\times$ 10 samples per persona.
Its chat template places a system prompt before the question in the user message, so system and user prompting coincide.}
\label{tab:persona-ministral}
\vspace{6pt}
\resizebox{\linewidth}{!}{%
\begin{tabular}{lccccccccc}
\toprule
 & \multicolumn{8}{c}{Persona score $\uparrow$} & Coh. $\uparrow$ \\
\cmidrule(lr){2-9}
Method & Evil & Sycoph. & Halluc. & Apathe. & Humoro. & Impoli. & Optimi. & Mean & Mean \\
\midrule
No steering & 0.0 & 10.9 & 53.8 & 4.5 & 0.5 & 0.6 & 70.9 & 20.2 & 93.9 \\
Prompting & 97.7 & 90.1 & 93.7 & 76.4 & 91.9 & 61.8 & 96.7 & 86.9 & 78.2 \\
\midrule
\multicolumn{10}{l}{\textit{\methodL{}}} \\
\methodL{} only, $\alpha{=}0.5$ & 0.4 & 13.2 & 51.4 & 4.5 & 8.4 & 0.8 & 80.7 & 22.8 & 94.1 \\
\methodL{} only, $\alpha{=}1$ & 19.8 & 35.9 & 69.9 & 8.2 & 93.8 & 16.0 & 94.4 & 48.3 & 81.5 \\
\methodL{} only, $\alpha{=}1.5$ & 62.9 & 70.3 & 87.7 & 44.2 & 95.4 & 69.8 & 97.9 & 75.5 & 59.3 \\
\methodL{} only, $\alpha{=}2$ & 75.4 & 82.1 & 91.7 & 81.3 & 96.2 & 81.2 & 99.1 & 86.7 & 44.7 \\
Prompt + \methodL{}, $\alpha{=}0.5$ & 96.9 & 90.6 & 94.4 & 79.0 & 92.3 & 63.3 & 97.4 & 87.7 & 76.0 \\
Prompt + \methodL{}, $\alpha{=}1$ & 97.6 & 94.1 & 95.7 & 83.8 & 94.5 & 71.3 & 98.9 & 90.8 & 67.1 \\
Prompt + \methodL{}, $\alpha{=}1.5$ & 98.8 & 96.5 & 97.8 & 86.5 & 95.1 & 86.1 & 99.5 & 94.3 & 57.7 \\
Prompt + \methodL{}, $\alpha{=}2$ & 98.6 & 96.5 & 97.5 & 89.6 & 95.6 & 91.3 & 99.6 & 95.5 & 49.4 \\
\bottomrule
\end{tabular}}
\end{table}

\subsection{Verifiable Instruction Following}
\label{app:results-if}
\textbf{Evaluation Setup.}
\ttt{IFEval} uses all 541 prompts with the official checker~\citep{zhou2023instruction}.
For \methodL{} alone, the instruction is removed from the prompt~\citep{stolfo2025instruction} and passed to the adapter.
For prompt+\methodL{}, the instruction remains in the prompt and is also passed to the adapter.
\Cref{tab:if-category} reports instruction-level strict accuracy, pooled over all instructions in a benchmark or within a \textsc{Global} or \textsc{Local} block.
The instruction-free versions of 376 \ttt{IFEval} questions occur in the training request pool.
The other 165 requests contain one instruction each and are easier, with prompting at 75--96\%.
On them, prompt+\methodL{} at $\alpha=1$ changes accuracy over prompting by +5.5, $-1.2$, +5.5, +3.0, $-1.8$, and +11.5 points on Gemma-2-9B, Qwen2.5-7B, Llama-3.1-8B, Phi-4-14B, LFM2.5-1.2B, and Ministral-8B, similar to the gains on all prompts.
\ttt{IFBench} uses 219 of its 300 prompts after removing prompts whose verifiable instructions cannot be located and prompts whose instruction types duplicate or closely match a training type.
We use its strict checker~\citep{pyatkin2025ifbench}.
Coherence is evaluated separately from checker-based accuracy.
We do not split the prompts into validation and held-out sets, because no adherence level is selected on the test prompts: $\alpha$ is configurable at test time, every result uses a fixed $\alpha$, and the tables report each level.

\Cref{tab:if-main-dial} in the main text gives prompt-level \ttt{IFEval} accuracy for every base LLM with \methodL{}, and \cref{tab:if-main} gives it for \methodLite{} with the no-instruction baseline.
\Cref{tab:if-category} reports accuracy for each verifiable instruction category and the pooled \textsc{Global} and \textsc{Local} blocks.

\begin{table}[htbp]
  \centering\small
  \setlength{\tabcolsep}{3pt}
  \caption{\textbf{Verifiable Instruction Following with \methodLite{}.} Prompt-level strict accuracy (\%) on \ttt{IFEval} and \ttt{IFBench}.
  \ttt{IFEval} instruction types occur in training, while \ttt{IFBench} types do not.}
  \label{tab:if-main}
  \vspace{6pt}
  \resizebox{\linewidth}{!}{%
  \begin{tabular}{llccccc}
    \toprule
    Benchmark & Method & Gemma-2-9B & Qwen2.5-7B & Llama-3.1-8B & Phi-4-14B & LFM2.5-1.2B \\
    \midrule
    \multirow{4}{*}{\shortstack[l]{\ttt{IFEval} (541)\\prompt-level strict}}
      & No instruction & 9.6 & 9.2 & 9.8 & 10.0 & 9.8 \\
      & \methodLite{} only ($\alpha{=}2$) & 49.0 & 47.5 & 45.1 & 43.4 & 54.5 \\
      & Prompting & 72.8 & 72.3 & 73.6 & 61.7 & 83.4 \\
      & Prompt + \methodLite{} ($\alpha{=}1$) & \textbf{77.6} & \textbf{73.6} & \textbf{77.1} & \textbf{69.7} & \textbf{84.3} \\
    \midrule
    \multirow{3}{*}{\shortstack[l]{\ttt{IFBench} (219)\\strict}}
      & No instruction & 7.8 & 10.0 & 8.2 & 9.6 & 9.1 \\
      & \methodLite{} only ($\alpha{=}2$) & 17.4 & 22.8 & 13.7 & 11.0 & 24.2 \\
      & Prompting & \textbf{19.2} & \textbf{37.4} & \textbf{24.2} & \textbf{18.7} & \textbf{46.1} \\
    \bottomrule
  \end{tabular}}
\end{table}

\begin{table*}[htbp]
  \centering\footnotesize
  \setlength{\tabcolsep}{2pt}
  \caption{\textbf{Instruction Following by Category on \ttt{IFEval}.} Instruction-level strict accuracy (\%) for global and local instructions.
  All pools the instructions in each block.
  Bold: best per LLM.}
  \label{tab:if-category}
  \vspace{6pt}
  \resizebox{\linewidth}{!}{%
  \begin{tabular}{lccccccccccccccc}
\toprule
 & \multicolumn{5}{c}{Gemma-2-9B} & \multicolumn{5}{c}{Qwen2.5-7B} & \multicolumn{5}{c}{Llama-3.1-8B} \\
\cmidrule(lr){2-6} \cmidrule(lr){7-11} \cmidrule(lr){12-16}
 &  & \multicolumn{2}{c}{\methodL{}} & \multicolumn{2}{c}{P+\methodL{}} &  & \multicolumn{2}{c}{\methodL{}} & \multicolumn{2}{c}{P+\methodL{}} &  & \multicolumn{2}{c}{\methodL{}} & \multicolumn{2}{c}{P+\methodL{}} \\
\cmidrule(lr){3-4} \cmidrule(lr){5-6} \cmidrule(lr){8-9} \cmidrule(lr){10-11} \cmidrule(lr){13-14} \cmidrule(lr){15-16}
Instruction & Prompting & $\alpha{=}1$ & $\alpha{=}2$ & $\alpha{=}0.5$ & $\alpha{=}1$ & Prompting & $\alpha{=}1$ & $\alpha{=}2$ & $\alpha{=}0.5$ & $\alpha{=}1$ & Prompting & $\alpha{=}1$ & $\alpha{=}2$ & $\alpha{=}0.5$ & $\alpha{=}1$ \\
\midrule
\multicolumn{16}{@{}l}{\textbf{\ttt{IFEval}}} \\
\quad \textit{All} & 80.2 & 55.0 & 64.4 & 83.7 & \textbf{83.8} & 79.6 & 58.4 & 60.4 & 79.4 & \textbf{80.2} & 81.1 & 52.6 & 59.4 & \textbf{84.9} & 83.0 \\
\multicolumn{16}{@{}l}{\quad\textsc{Global}} \\
\qquad Case & 71.9 & 76.4 & 85.4 & 77.5 & \textbf{87.6} & 73.0 & \textbf{82.0} & 76.4 & 76.4 & \textbf{82.0} & 78.7 & 82.0 & 82.0 & \textbf{86.5} & 82.0 \\
\qquad Language & 93.5 & 16.1 & 12.9 & 93.5 & \textbf{96.8} & \textbf{100.0} & 22.6 & 16.1 & 96.8 & 71.0 & \textbf{93.5} & 25.8 & 32.3 & \textbf{93.5} & \textbf{93.5} \\
\qquad Punctuation & 90.9 & 50.0 & 87.9 & \textbf{97.0} & \textbf{97.0} & 97.0 & 98.5 & 98.5 & 98.5 & \textbf{100.0} & 90.9 & 65.2 & 93.9 & 92.4 & \textbf{95.5} \\
\qquad Length & 62.6 & 56.5 & 64.9 & 67.2 & \textbf{69.5} & 64.1 & 56.5 & 58.8 & 61.1 & \textbf{70.2} & 74.0 & 51.9 & 63.4 & \textbf{81.7} & 80.2 \\
\qquad Format & 89.8 & 70.1 & 73.9 & \textbf{90.4} & 89.8 & 88.5 & 77.1 & 84.1 & 90.4 & \textbf{92.4} & 84.1 & 63.1 & 67.5 & 84.7 & \textbf{85.4} \\
\qquad Wrapping & 93.8 & 78.5 & 84.6 & \textbf{98.5} & 95.4 & \textbf{90.8} & 60.0 & 55.4 & 87.7 & 73.8 & 92.3 & 61.5 & 58.5 & \textbf{93.8} & 73.8 \\
\qquad \textit{All} & 81.1 & 63.3 & 73.1 & 84.6 & \textbf{86.5} & 82.0 & 70.3 & 71.1 & 82.0 & \textbf{82.7} & 83.1 & 61.4 & 69.0 & \textbf{86.8} & 83.9 \\
\cmidrule(l){1-16}
\multicolumn{16}{@{}l}{\quad\textsc{Local}} \\
\qquad Keywords & 77.9 & 36.8 & 46.0 & \textbf{81.0} & 79.8 & 76.1 & 31.9 & 34.4 & 77.3 & \textbf{77.9} & 77.3 & 35.6 & 37.4 & 82.2 & \textbf{83.4} \\
\qquad Positions & \textbf{89.5} & 10.5 & 10.5 & 86.8 & 71.1 & \textbf{73.7} & 10.5 & 13.2 & 65.8 & 65.8 & 78.9 & 10.5 & 7.9 & \textbf{84.2} & 76.3 \\
\qquad Markers & 83.0 & 69.8 & 84.9 & 90.6 & \textbf{94.3} & \textbf{88.7} & 66.0 & 79.2 & 84.9 & 86.8 & 92.5 & 56.6 & 84.9 & 94.3 & \textbf{98.1} \\
\qquad Copying & 65.9 & 41.5 & 46.3 & \textbf{70.7} & 63.4 & 56.1 & 41.5 & 43.9 & 58.5 & \textbf{61.0} & 56.1 & 39.0 & 34.1 & \textbf{58.5} & 56.1 \\
\qquad \textit{All} & 78.6 & 40.0 & 48.5 & \textbf{82.0} & 79.0 & 75.3 & 36.6 & 41.0 & 74.6 & \textbf{75.6} & 77.3 & 36.6 & 41.7 & \textbf{81.4} & \textbf{81.4} \\
\bottomrule
  \end{tabular}
  }

  \vspace{8pt}
  \resizebox{\linewidth}{!}{%
  \begin{tabular}{lccccccccccccccc}
\toprule
 & \multicolumn{5}{c}{Phi-4-14B} & \multicolumn{5}{c}{LFM2.5-1.2B} & \multicolumn{5}{c}{Ministral-8B} \\
\cmidrule(lr){2-6} \cmidrule(lr){7-11} \cmidrule(lr){12-16}
 &  & \multicolumn{2}{c}{\methodL{}} & \multicolumn{2}{c}{P+\methodL{}} &  & \multicolumn{2}{c}{\methodL{}} & \multicolumn{2}{c}{P+\methodL{}} &  & \multicolumn{2}{c}{\methodL{}} & \multicolumn{2}{c}{P+\methodL{}} \\
\cmidrule(lr){3-4} \cmidrule(lr){5-6} \cmidrule(lr){8-9} \cmidrule(lr){10-11} \cmidrule(lr){13-14} \cmidrule(lr){15-16}
Instruction & Prompting & $\alpha{=}1$ & $\alpha{=}2$ & $\alpha{=}0.5$ & $\alpha{=}1$ & Prompting & $\alpha{=}1$ & $\alpha{=}2$ & $\alpha{=}0.5$ & $\alpha{=}1$ & Prompting & $\alpha{=}1$ & $\alpha{=}2$ & $\alpha{=}0.5$ & $\alpha{=}1$ \\
\midrule
\multicolumn{16}{@{}l}{\textbf{\ttt{IFEval}}} \\
\quad \textit{All} & 71.8 & 45.7 & 54.4 & 72.3 & \textbf{76.3} & 88.1 & 62.9 & 67.0 & \textbf{89.7} & \textbf{89.7} & 63.2 & 42.6 & 54.4 & 67.5 & \textbf{73.4} \\
\multicolumn{16}{@{}l}{\quad\textsc{Global}} \\
\qquad Case & 52.8 & 58.4 & \textbf{76.4} & 55.1 & 69.7 & 89.9 & 86.5 & 89.9 & \textbf{92.1} & \textbf{92.1} & 53.9 & 64.0 & \textbf{78.7} & 62.9 & 75.3 \\
\qquad Language & 90.3 & 12.9 & 9.7 & \textbf{93.5} & 87.1 & 83.9 & 12.9 & 12.9 & \textbf{90.3} & 77.4 & 87.1 & 48.4 & 54.8 & 87.1 & \textbf{96.8} \\
\qquad Punctuation & \textbf{21.2} & 4.5 & 18.2 & 19.7 & 19.7 & \textbf{100.0} & 83.3 & 97.0 & \textbf{100.0} & \textbf{100.0} & 39.4 & 28.8 & \textbf{66.7} & 40.9 & 36.4 \\
\qquad Length & 65.6 & 56.5 & 67.9 & 69.5 & \textbf{75.6} & 80.9 & 74.8 & 76.3 & 86.3 & \textbf{88.5} & 50.4 & 41.2 & 55.7 & 52.7 & \textbf{61.1} \\
\qquad Format & \textbf{86.6} & 70.1 & 79.6 & \textbf{86.6} & \textbf{86.6} & 94.3 & 79.0 & 82.2 & \textbf{95.5} & 94.9 & 75.2 & 61.8 & 75.2 & 83.4 & \textbf{91.7} \\
\qquad Wrapping & \textbf{87.7} & 60.0 & 55.4 & 84.6 & 83.1 & \textbf{95.4} & 60.0 & 69.2 & 93.8 & \textbf{95.4} & 69.2 & 38.5 & 50.8 & 69.2 & \textbf{73.8} \\
\qquad \textit{All} & 68.3 & 52.3 & 61.8 & 69.2 & \textbf{72.5} & 90.5 & 73.7 & 78.3 & \textbf{92.8} & 92.6 & 61.2 & 49.5 & 65.9 & 65.9 & \textbf{72.9} \\
\cmidrule(l){1-16}
\multicolumn{16}{@{}l}{\quad\textsc{Local}} \\
\qquad Keywords & 75.5 & 38.7 & 39.9 & 74.8 & \textbf{81.6} & 89.0 & 48.5 & 52.1 & 89.6 & \textbf{90.2} & 66.9 & 35.6 & 33.7 & 67.5 & \textbf{71.8} \\
\qquad Positions & \textbf{84.2} & 7.9 & 7.9 & \textbf{84.2} & 76.3 & 94.7 & 13.2 & 10.5 & \textbf{97.4} & \textbf{97.4} & 60.5 & 10.5 & 2.6 & \textbf{71.1} & 63.2 \\
\qquad Markers & 90.6 & 45.3 & 69.8 & 90.6 & \textbf{92.5} & \textbf{100.0} & 71.7 & 75.5 & \textbf{100.0} & \textbf{100.0} & 83.0 & 34.0 & 58.5 & 81.1 & \textbf{92.5} \\
\qquad Copying & 68.3 & 22.0 & 39.0 & 68.3 & \textbf{82.9} & \textbf{31.7} & 14.6 & 19.5 & 29.3 & 29.3 & 51.2 & 19.5 & 29.3 & 68.3 & \textbf{70.7} \\
\qquad \textit{All} & 78.3 & 33.6 & 41.0 & 78.0 & \textbf{83.1} & 83.7 & 43.4 & 46.4 & 84.1 & \textbf{84.4} & 66.8 & 29.8 & 33.6 & 70.5 & \textbf{74.2} \\
\bottomrule
  \end{tabular}
  }
\end{table*}

\textbf{ \methodLite{} with Prompting.}
On \ttt{IFEval}, prompt+\methodLite{} raises prompt-level strict accuracy over prompting on all five LLMs: from 72.8 to 77.6 on Gemma and from 61.7 to 69.7 on Phi-4-14B, and from 73.6 to 77.1 on Llama, from 72.3 to 73.6 on Qwen, and from 83.4 to 84.3 on LFM2.5.

\textbf{ \methodLite{} without Prompting.}
Without the instruction in the prompt, \methodLite{} at $\alpha=2$ raises \ttt{IFEval} accuracy from about 10 to 43--55 for every LLM, more than half of the gain obtained by prompting.

\textbf{Verifiable Instruction Categories.}
Without the instruction in the prompt, \methodL{} raises Gemma's accuracy on \textsc{Global} verifiable instructions from 17.3 to 73.1, close to prompting's 81.1, and exceeds prompting on letter-case instructions.
On \textsc{Local} verifiable instructions, it reaches 48.5 against prompting's 78.6, and it fails when an exact word must appear at an exact position (10.5).
On top of the prompt at $\alpha{=}0.5$, \methodL{} raises accuracy over all 834 \ttt{IFEval} instructions from 80.2 to 83.7, \textsc{Global} accuracy from 81.1 to 84.6, and \textsc{Local} accuracy from 78.6 to 82.0.
It improves every category except language, where it ties at 93.5, and word position, where accuracy falls from 89.5 to 86.8 (\cref{tab:if-category}).

The level determines which kind of verifiable instruction is favored.
From $\alpha{=}0.5$ to 2, \textsc{Global} accuracy rises from 84.6 to 88.3 while \textsc{Local} accuracy falls from 82.0 to 77.6, so the total remains near 84.
On Qwen, prompt+\methodL{} scores slightly above prompting only at $\alpha{=}1$ (80.2 against 79.6) and below it at $\alpha{=}0.5$ and 2 (79.4 and 78.1), and steering alone reaches 71.1 on \textsc{Global} verifiable instructions against prompting's 82.0.
On Llama, prompt+\methodL{} raises accuracy from 81.1 to 84.9 at $\alpha{=}0.5$, on Phi-4-14B from 71.8 to 76.3 at $\alpha{=}1$, and on LFM2.5 from 88.1 to 89.7 at both levels, and on Ministral-8B from 63.2 to 73.4 at $\alpha{=}1$ (\cref{tab:if-category,tab:if-dial}), showing that the best adherence level depends on the LLM.

\textbf{CFG on \ttt{IFEval}.}
\Cref{tab:if-cfg} reports CFG, which provides the training signal of \methodL{}.
At $\alpha=1$, CFG reduces to prompting; the remaining differences of at most 2.4 points come from the two inference implementations.
At $\alpha=1.5$ and 2, CFG improves over prompting on Gemma-2-9B, Phi-4-14B, and Ministral-8B (up to 78.0, 72.5, and 67.1 against 72.8, 61.7, and 53.4), by 3.5 points on Llama-3.1-8B, and not on Qwen2.5-7B or LFM2.5-1.2B.
Verifiable instructions are checked as pass or fail, so strengthening an instruction beyond prompting brings limited gains, which also bounds what distilling CFG can add.
At matched levels, where prompt+\methodL{} at $\alpha$ corresponds to CFG at $1+\alpha$ (\cref{app:results-concept}), prompt+\methodL{} at $\alpha=1$ is within 0.4 points of CFG at $\alpha=2$ or above it on four LLMs (76.9, 71.9, 75.6, and 85.6 against 76.2, 72.3, 75.4, and 81.1), and below it on Phi-4-14B (66.7 against 72.5) and Ministral-8B (65.1 against 67.1), while CFG requires two LLM passes per token (\cref{tab:cost}).
\begin{table}[t]
  \centering\small
  \setlength{\tabcolsep}{4pt}\renewcommand{\arraystretch}{1.05}
  \caption{\textbf{CFG on \ttt{IFEval}.} Prompt-level strict accuracy (\%).
  CFG runs the LLM with and without the instruction at every token.
  Prompt+\methodL{} results are from \cref{tab:if-main-dial}.
  Bold indicates the best result per LLM.}
  \label{tab:if-cfg}
  \vspace{6pt}
  \resizebox{\linewidth}{!}{%
  \begin{tabular}{lccccccc}
    \toprule
    Method & $\alpha$ & Gemma-2-9B & Qwen2.5-7B & Llama-3.1-8B & Phi-4-14B & LFM2.5-1.2B & Ministral-8B \\
    \midrule
    No instruction & -- & 9.6 & 9.2 & 9.8 & 10.0 & 9.8 & 8.5 \\
    Prompting & -- & 72.8 & \textbf{72.3} & 73.6 & 61.7 & 83.4 & 53.4 \\
    \midrule
    \multirow{4}{*}{CFG} & 0.5 & 21.4 & 30.5 & 21.3 & 19.0 & 50.1 & 14.0 \\
     & 1 & 72.5 & 71.2 & 74.7 & 64.1 & 83.7 & 54.5 \\
     & 1.5 & \textbf{78.0} & \textbf{72.3} & 77.1 & 72.1 & 81.3 & 64.7 \\
     & 2 & 76.2 & \textbf{72.3} & 75.4 & \textbf{72.5} & 81.1 & 67.1 \\
    \midrule
    \multirow{4}{*}{P+\methodL{}} & 0.5 & 77.1 & 71.9 & \textbf{78.0} & 63.2 & 85.2 & 58.0 \\
     & 1 & 76.9 & 71.9 & 75.6 & 66.7 & \textbf{85.6} & 65.1 \\
     & 1.5 & 77.1 & 70.6 & 73.8 & 68.2 & 83.7 & \textbf{67.5} \\
     & 2 & 77.4 & 70.1 & 71.2 & 67.5 & 81.3 & 66.4 \\
    \bottomrule
  \end{tabular}}
\end{table}

\textbf{Fine-Grained Results by Category.}\label{app:if-dial}
\Cref{tab:if-dial} reports every \ttt{IFEval} verifiable instruction category at every level, for \methodL{} alone and on top of the prompt, on all six LLMs.
On \ttt{IFEval}, prompt+\methodL{} at $\alpha{=}0.5$ raises pooled \textsc{Global} accuracy over prompting on Gemma (81.1 to 84.6) and Llama (83.1 to 86.8) and matches it on Qwen (82.0).
Higher levels trade \textsc{Local} for \textsc{Global} accuracy on Gemma-2-9B and lower both on Llama-3.1-8B relative to $\alpha{=}0.5$, showing that the category trade-off is LLM-specific.
\begin{table*}[htbp]
\centering
\caption{\textbf{Instruction Following by Type on \ttt{IFEval}.} Instruction-level strict accuracy (\%) at four adherence levels.
$n$ is the number of instructions, and averages pool each block.
Bold: best per row.}
\label{tab:if-dial}
\vspace{6pt}
\resizebox{\textwidth}{!}{%
\begin{tabular}{lrccccccccc}
\toprule
 & & & \multicolumn{4}{c}{\methodL{} ($\alpha$)} & \multicolumn{4}{c}{P+\methodL{} ($\alpha$)} \\
\cmidrule(lr){4-7} \cmidrule(lr){8-11}
Instruction & $n$ & Prompting & $0.5$ & $1$ & $1.5$ & $2$ & $0.5$ & $1$ & $1.5$ & $2$ \\
\midrule
\multicolumn{11}{l}{\textbf{Gemma-2-9B}} \\
\quad \textsc{Global} & &  &  &  &  &  &  &  &  &  \\
\qquad Case & 89 & 71.9 & 23.6 & 76.4 & 83.1 & 85.4 & 77.5 & 87.6 & 83.1 & \textbf{88.8} \\
\qquad Language & 31 & 93.5 & 12.9 & 16.1 & 12.9 & 12.9 & 93.5 & \textbf{96.8} & 80.6 & 83.9 \\
\qquad Punctuation & 66 & 90.9 & 15.2 & 50.0 & 77.3 & 87.9 & 97.0 & 97.0 & \textbf{98.5} & \textbf{98.5} \\
\qquad Length & 131 & 62.6 & 35.9 & 56.5 & 64.9 & 64.9 & 67.2 & 69.5 & 77.1 & \textbf{78.6} \\
\qquad Format & 157 & 89.8 & 32.5 & 70.1 & 71.3 & 73.9 & \textbf{90.4} & 89.8 & \textbf{90.4} & \textbf{90.4} \\
\qquad Wrapping & 65 & 93.8 & 21.5 & 78.5 & 86.2 & 84.6 & \textbf{98.5} & 95.4 & 92.3 & 93.8 \\
\qquad \textit{Global avg.} & 539 & 81.1 & 27.3 & 63.3 & 70.9 & 73.1 & 84.6 & 86.5 & 86.6 & \textbf{88.3} \\
\quad \textsc{Local} & &  &  &  &  &  &  &  &  &  \\
\qquad Keywords & 163 & 77.9 & 35.0 & 36.8 & 43.6 & 46.0 & 81.0 & 79.8 & 82.2 & \textbf{82.8} \\
\qquad Positions & 38 & \textbf{89.5} & 7.9 & 10.5 & 7.9 & 10.5 & 86.8 & 71.1 & 57.9 & 50.0 \\
\qquad Markers & 53 & 83.0 & 22.6 & 69.8 & 79.2 & 84.9 & 90.6 & 94.3 & \textbf{98.1} & 96.2 \\
\qquad Copying & 41 & 65.9 & 2.4 & 41.5 & 46.3 & 46.3 & \textbf{70.7} & 63.4 & 58.5 & 58.5 \\
\qquad \textit{Local avg.} & 295 & 78.6 & 24.7 & 40.0 & 45.8 & 48.5 & \textbf{82.0} & 79.0 & 78.6 & 77.6 \\
\midrule
\multicolumn{11}{l}{\textbf{Qwen2.5-7B}} \\
\quad \textsc{Global} & &  &  &  &  &  &  &  &  &  \\
\qquad Case & 89 & 73.0 & 24.7 & \textbf{82.0} & 78.7 & 76.4 & 76.4 & \textbf{82.0} & 71.9 & 74.2 \\
\qquad Language & 31 & \textbf{100.0} & 16.1 & 22.6 & 22.6 & 16.1 & 96.8 & 71.0 & 67.7 & 61.3 \\
\qquad Punctuation & 66 & 97.0 & 48.5 & 98.5 & \textbf{100.0} & 98.5 & 98.5 & \textbf{100.0} & 98.5 & 98.5 \\
\qquad Length & 131 & 64.1 & 49.6 & 56.5 & 61.8 & 58.8 & 61.1 & \textbf{70.2} & \textbf{70.2} & \textbf{70.2} \\
\qquad Format & 157 & 88.5 & 29.9 & 77.1 & 84.7 & 84.1 & 90.4 & \textbf{92.4} & 89.2 & 89.2 \\
\qquad Wrapping & 65 & \textbf{90.8} & 35.4 & 60.0 & 56.9 & 55.4 & 87.7 & 73.8 & 66.2 & 66.2 \\
\qquad \textit{Global avg.} & 539 & 82.0 & 36.0 & 70.3 & 73.1 & 71.1 & 82.0 & \textbf{82.7} & 78.8 & 78.8 \\
\quad \textsc{Local} & &  &  &  &  &  &  &  &  &  \\
\qquad Keywords & 163 & 76.1 & 29.4 & 31.9 & 34.4 & 34.4 & 77.3 & 77.9 & \textbf{83.4} & 82.2 \\
\qquad Positions & 38 & \textbf{73.7} & 13.2 & 10.5 & 10.5 & 13.2 & 65.8 & 65.8 & 65.8 & 57.9 \\
\qquad Markers & 53 & 88.7 & 17.0 & 66.0 & 75.5 & 79.2 & 84.9 & 86.8 & 86.8 & \textbf{90.6} \\
\qquad Copying & 41 & 56.1 & 29.3 & 41.5 & 43.9 & 43.9 & 58.5 & \textbf{61.0} & \textbf{61.0} & 53.7 \\
\qquad \textit{Local avg.} & 295 & 75.3 & 25.1 & 36.6 & 40.0 & 41.0 & 74.6 & 75.6 & \textbf{78.6} & 76.6 \\
\midrule
\multicolumn{11}{l}{\textbf{Llama-3.1-8B}} \\
\quad \textsc{Global} & &  &  &  &  &  &  &  &  &  \\
\qquad Case & 89 & 78.7 & 33.7 & 82.0 & 85.4 & 82.0 & 86.5 & 82.0 & \textbf{89.9} & 87.6 \\
\qquad Language & 31 & \textbf{93.5} & 19.4 & 25.8 & 32.3 & 32.3 & \textbf{93.5} & \textbf{93.5} & 83.9 & 83.9 \\
\qquad Punctuation & 66 & 90.9 & 24.2 & 65.2 & 81.8 & 93.9 & 92.4 & 95.5 & \textbf{98.5} & 97.0 \\
\qquad Length & 131 & 74.0 & 37.4 & 51.9 & 64.9 & 63.4 & \textbf{81.7} & 80.2 & 77.9 & 80.9 \\
\qquad Format & 157 & 84.1 & 21.7 & 63.1 & 67.5 & 67.5 & 84.7 & \textbf{85.4} & 82.2 & 79.0 \\
\qquad Wrapping & 65 & 92.3 & 27.7 & 61.5 & 56.9 & 58.5 & \textbf{93.8} & 73.8 & 64.6 & 58.5 \\
\qquad \textit{Global avg.} & 539 & 83.1 & 28.4 & 61.4 & 68.3 & 69.0 & \textbf{86.8} & 83.9 & 82.4 & 80.9 \\
\quad \textsc{Local} & &  &  &  &  &  &  &  &  &  \\
\qquad Keywords & 163 & 77.3 & 33.7 & 35.6 & 38.0 & 37.4 & 82.2 & 83.4 & \textbf{85.3} & 84.0 \\
\qquad Positions & 38 & 78.9 & 10.5 & 10.5 & 10.5 & 7.9 & \textbf{84.2} & 76.3 & 60.5 & 55.3 \\
\qquad Markers & 53 & 92.5 & 15.1 & 56.6 & 83.0 & 84.9 & 94.3 & 98.1 & \textbf{100.0} & 98.1 \\
\qquad Copying & 41 & 56.1 & 7.3 & 39.0 & 36.6 & 34.1 & 58.5 & 56.1 & \textbf{63.4} & 58.5 \\
\qquad \textit{Local avg.} & 295 & 77.3 & 23.7 & 36.6 & 42.4 & 41.7 & 81.4 & 81.4 & \textbf{81.7} & 79.3 \\
\bottomrule
\end{tabular}}

\end{table*}
\begin{table*}[htbp]
\centering
\caption{\textbf{Instruction Following by Type on \ttt{IFEval}.} Results for Phi-4-14B, LFM2.5-1.2B, and Ministral-8B, using the same metrics and columns as \cref{tab:if-dial}.}
\label{tab:if-dial-b}
\vspace{6pt}
\resizebox{\textwidth}{!}{%
\begin{tabular}{lrccccccccc}
\toprule
 & & & \multicolumn{4}{c}{\methodL{} ($\alpha$)} & \multicolumn{4}{c}{P+\methodL{} ($\alpha$)} \\
\cmidrule(lr){4-7} \cmidrule(lr){8-11}
Instruction & $n$ & Prompting & $0.5$ & $1$ & $1.5$ & $2$ & $0.5$ & $1$ & $1.5$ & $2$ \\
\midrule
\multicolumn{11}{l}{\textbf{Phi-4-14B}} \\
\quad \textsc{Global} & &  &  &  &  &  &  &  &  &  \\
\qquad Case & 89 & 52.8 & 10.1 & 58.4 & 67.4 & \textbf{76.4} & 55.1 & 69.7 & 71.9 & 68.5 \\
\qquad Language & 31 & 90.3 & 16.1 & 12.9 & 9.7 & 9.7 & \textbf{93.5} & 87.1 & 71.0 & 61.3 \\
\qquad Punctuation & 66 & 21.2 & 4.5 & 4.5 & 7.6 & 18.2 & 19.7 & 19.7 & 36.4 & \textbf{42.4} \\
\qquad Length & 131 & 65.6 & 41.2 & 56.5 & 65.6 & 67.9 & 69.5 & 75.6 & 75.6 & \textbf{77.1} \\
\qquad Format & 157 & 86.6 & 31.2 & 70.1 & 79.0 & 79.6 & 86.6 & 86.6 & \textbf{90.4} & \textbf{90.4} \\
\qquad Wrapping & 65 & \textbf{87.7} & 10.8 & 60.0 & 60.0 & 55.4 & 84.6 & 83.1 & 86.2 & 83.1 \\
\qquad \textit{Global avg.} & 539 & 68.3 & 23.6 & 52.3 & 58.8 & 61.8 & 69.2 & 72.5 & \textbf{75.5} & 75.1 \\
\quad \textsc{Local} & &  &  &  &  &  &  &  &  &  \\
\qquad Keywords & 163 & 75.5 & 31.3 & 38.7 & 38.7 & 39.9 & 74.8 & \textbf{81.6} & 81.0 & 79.1 \\
\qquad Positions & 38 & \textbf{84.2} & 10.5 & 7.9 & 7.9 & 7.9 & \textbf{84.2} & 76.3 & 68.4 & 63.2 \\
\qquad Markers & 53 & 90.6 & 9.4 & 45.3 & 67.9 & 69.8 & 90.6 & 92.5 & \textbf{94.3} & \textbf{94.3} \\
\qquad Copying & 41 & 68.3 & 2.4 & 22.0 & 31.7 & 39.0 & 68.3 & \textbf{82.9} & 78.0 & 75.6 \\
\qquad \textit{Local avg.} & 295 & 78.3 & 20.7 & 33.6 & 39.0 & 41.0 & 78.0 & \textbf{83.1} & 81.4 & 79.3 \\
\midrule
\multicolumn{11}{l}{\textbf{LFM2.5-1.2B}} \\
\quad \textsc{Global} & &  &  &  &  &  &  &  &  &  \\
\qquad Case & 89 & 89.9 & 75.3 & 86.5 & 89.9 & 89.9 & 92.1 & 92.1 & 91.0 & \textbf{93.3} \\
\qquad Language & 31 & 83.9 & 29.0 & 12.9 & 9.7 & 12.9 & \textbf{90.3} & 77.4 & 74.2 & 67.7 \\
\qquad Punctuation & 66 & \textbf{100.0} & 39.4 & 83.3 & 92.4 & 97.0 & \textbf{100.0} & \textbf{100.0} & \textbf{100.0} & \textbf{100.0} \\
\qquad Length & 131 & 80.9 & 64.9 & 74.8 & 79.4 & 76.3 & 86.3 & 88.5 & \textbf{89.3} & 84.0 \\
\qquad Format & 157 & 94.3 & 56.1 & 79.0 & 81.5 & 82.2 & \textbf{95.5} & 94.9 & 93.0 & 91.7 \\
\qquad Wrapping & 65 & \textbf{95.4} & 50.8 & 60.0 & 63.1 & 69.2 & 93.8 & \textbf{95.4} & 90.8 & 90.8 \\
\qquad \textit{Global avg.} & 539 & 90.5 & 57.1 & 73.7 & 77.4 & 78.3 & \textbf{92.8} & 92.6 & 91.3 & 89.6 \\
\quad \textsc{Local} & &  &  &  &  &  &  &  &  &  \\
\qquad Keywords & 163 & 89.0 & 39.9 & 48.5 & 47.2 & 52.1 & 89.6 & \textbf{90.2} & 89.0 & 87.7 \\
\qquad Positions & 38 & 94.7 & 13.2 & 13.2 & 13.2 & 10.5 & \textbf{97.4} & \textbf{97.4} & 92.1 & 94.7 \\
\qquad Markers & 53 & \textbf{100.0} & 56.6 & 71.7 & 75.5 & 75.5 & \textbf{100.0} & \textbf{100.0} & \textbf{100.0} & \textbf{100.0} \\
\qquad Copying & 41 & \textbf{31.7} & 9.8 & 14.6 & 17.1 & 19.5 & 29.3 & 29.3 & 29.3 & 26.8 \\
\qquad \textit{Local avg.} & 295 & 83.7 & 35.3 & 43.4 & 43.7 & 46.4 & 84.1 & \textbf{84.4} & 83.1 & 82.4 \\
\midrule
\multicolumn{11}{l}{\textbf{Ministral-8B}} \\
\quad \textsc{Global} & &  &  &  &  &  &  &  &  &  \\
\qquad Case & 89 & 53.9 & 14.6 & 64.0 & 76.4 & 78.7 & 62.9 & 75.3 & \textbf{83.1} & \textbf{83.1} \\
\qquad Language & 31 & 87.1 & 16.1 & 48.4 & 61.3 & 54.8 & 87.1 & \textbf{96.8} & 87.1 & 83.9 \\
\qquad Punctuation & 66 & 39.4 & 9.1 & 28.8 & 48.5 & \textbf{66.7} & 40.9 & 36.4 & 53.0 & 60.6 \\
\qquad Length & 131 & 50.4 & 28.2 & 41.2 & 55.7 & 55.7 & 52.7 & 61.1 & 61.8 & \textbf{63.4} \\
\qquad Format & 157 & 75.2 & 19.1 & 61.8 & 71.3 & 75.2 & 83.4 & 91.7 & \textbf{92.4} & \textbf{92.4} \\
\qquad Wrapping & 65 & 69.2 & 7.7 & 38.5 & 47.7 & 50.8 & 69.2 & \textbf{73.8} & 69.2 & 66.2 \\
\qquad \textit{Global avg.} & 539 & 61.2 & 17.8 & 49.5 & 62.2 & 65.9 & 65.9 & 72.9 & 75.5 & \textbf{76.3} \\
\quad \textsc{Local} & &  &  &  &  &  &  &  &  &  \\
\qquad Keywords & 163 & 66.9 & 30.7 & 35.6 & 38.0 & 33.7 & 67.5 & 71.8 & \textbf{75.5} & 72.4 \\
\qquad Positions & 38 & 60.5 & 7.9 & 10.5 & 10.5 & 2.6 & \textbf{71.1} & 63.2 & 57.9 & 50.0 \\
\qquad Markers & 53 & 83.0 & 9.4 & 34.0 & 62.3 & 58.5 & 81.1 & 92.5 & 98.1 & \textbf{100.0} \\
\qquad Copying & 41 & 51.2 & 2.4 & 19.5 & 39.0 & 29.3 & 68.3 & \textbf{70.7} & \textbf{70.7} & 65.9 \\
\qquad \textit{Local avg.} & 295 & 66.8 & 20.0 & 29.8 & 39.0 & 33.6 & 70.5 & 74.2 & \textbf{76.6} & 73.6 \\
\bottomrule
\end{tabular}}

\end{table*}

\textbf{Verifiable Instruction Types Unseen in Training.}\label{app:ifbench}
After filtering, the 219 \ttt{IFBench} prompts contain 35 of its 58 verifiable instruction types, none of which duplicates or closely matches a \ttt{SteerIF} type, whereas all 25 \ttt{IFEval} types appear in \ttt{SteerIF}.
Without the instruction in the prompt, the instruction-level strict accuracy of \methodL{} rises from $\alpha{=}1$ to 2 on every LLM, e.g., from 10.7 to 15.9 on Gemma and from 13.9 to 18.7 on Qwen (\cref{tab:if-ifbench}).
On top of the prompt, $\alpha=1$ gives higher instruction-level accuracy than prompting on Gemma, Qwen, Phi-4-14B, and Ministral-8B and lower accuracy on Llama and LFM2.5, with differences of at most 4.4 points.
\Cref{tab:if-ifbench} gives instruction-level accuracy by block, and \cref{tab:if-full-ifbench,tab:if-full-ifbench-b} give every verifiable instruction type at every level.
\begin{table}[htbp]
  \centering\small
  \setlength{\tabcolsep}{4pt}\renewcommand{\arraystretch}{1.08}
  \caption{\textbf{Instruction Following on Unseen Instruction Types.} Instruction-level strict accuracy (\%) on 219 \ttt{IFBench} prompts, grouped into global and local instructions.
  Bold: best per LLM.}
  \label{tab:if-ifbench}
  \vspace{6pt}
  \begin{tabular*}{\linewidth}{@{\extracolsep{\fill}}lccccc@{}}
    \toprule
    & & \multicolumn{2}{c}{\methodL{}} & \multicolumn{2}{c}{prompt+\methodL{}} \\
    \cmidrule(lr){3-4}\cmidrule(l){5-6}
    Instruction & Prompting & $\alpha=1$ & $\alpha=2$ & $\alpha=0.5$ & $\alpha=1$ \\
    \midrule
    \multicolumn{6}{@{}l}{\textbf{Gemma-2-9B}} \\
    \textit{All} & 20.6 & 10.7 & 15.9 & 20.2 & \textbf{23.4} \\
    \textsc{Global} & 19.2 & 12.1 & 17.6 & 19.2 & \textbf{23.6} \\
    \textsc{Local} & \textbf{24.3} & 7.1 & 11.4 & 22.9 & 22.9 \\
    \midrule
    \multicolumn{6}{@{}l}{\textbf{Qwen2.5-7B}} \\
    \textit{All} & 38.1 & 13.9 & 18.7 & 32.5 & \textbf{38.9} \\
    \textsc{Global} & \textbf{41.2} & 16.5 & 22.5 & 36.8 & \textbf{41.2} \\
    \textsc{Local} & 30.0 & 7.1 & 8.6 & 21.4 & \textbf{32.9} \\
    \midrule
    \multicolumn{6}{@{}l}{\textbf{Llama-3.1-8B}} \\
    \textit{All} & \textbf{25.4} & 8.7 & 13.1 & 19.4 & 24.6 \\
    \textsc{Global} & \textbf{24.7} & 12.1 & 15.9 & 17.6 & \textbf{24.7} \\
    \textsc{Local} & \textbf{27.1} & 0.0 & 5.7 & 24.3 & 24.3 \\
    \midrule
    \multicolumn{6}{@{}l}{\textbf{Phi-4-14B}} \\
    \textit{All} & 19.0 & 9.5 & 13.1 & 18.7 & \textbf{23.4} \\
    \textsc{Global} & 19.2 & 11.5 & 15.4 & 18.7 & \textbf{23.6} \\
    \textsc{Local} & 18.6 & 4.3 & 7.1 & 18.6 & \textbf{22.9} \\
    \midrule
    \multicolumn{6}{@{}l}{\textbf{LFM2.5-1.2B}} \\
    \textit{All} & \textbf{46.0} & 19.4 & 21.0 & 41.3 & 41.7 \\
    \textsc{Global} & \textbf{45.1} & 22.5 & 24.2 & 42.9 & 41.2 \\
    \textsc{Local} & \textbf{48.6} & 11.4 & 12.9 & 37.1 & 42.9 \\
    \midrule
    \multicolumn{6}{@{}l}{\textbf{Ministral-8B}} \\
    \textit{All} & 17.1 & 11.1 & 13.9 & 15.9 & \textbf{17.5} \\
    \textsc{Global} & 18.7 & 13.7 & 15.4 & 17.0 & \textbf{19.8} \\
    \textsc{Local} & \textbf{12.9} & 4.3 & 10.0 & \textbf{12.9} & 11.4 \\
    \bottomrule
  \end{tabular*}
\end{table}

\begin{table*}[htbp]
\centering
\caption{\textbf{Instruction Following by Type on \ttt{IFBench}.} Instruction-level strict accuracy (\%) at four adherence levels.
$n$ is the number of instructions, and averages pool each block.
Bold: best per row.}
\label{tab:if-full-ifbench}
\vspace{6pt}
\resizebox{\textwidth}{!}{%
\begin{tabular}{lrccccccccc}
\toprule
 & & & \multicolumn{4}{c}{\methodL{} ($\alpha$)} & \multicolumn{4}{c}{P+\methodL{} ($\alpha$)} \\
\cmidrule(lr){4-7} \cmidrule(lr){8-11}
Instruction & $n$ & Prompting & $0.5$ & $1$ & $1.5$ & $2$ & $0.5$ & $1$ & $1.5$ & $2$ \\
\midrule
\multicolumn{11}{l}{\textbf{Gemma-2-9B}} \\
\quad \textsc{Global} & &  &  &  &  &  &  &  &  &  \\
\qquad Count & 15 & 73.3 & 26.7 & 46.7 & 60.0 & 53.3 & 73.3 & 80.0 & 80.0 & \textbf{93.3} \\
\qquad Format & 77 & 23.4 & 7.8 & 11.7 & 15.6 & 22.1 & 20.8 & 26.0 & 28.6 & \textbf{37.7} \\
\qquad Ratio & 32 & 15.6 & 18.8 & 18.8 & 18.8 & 21.9 & 18.8 & \textbf{25.0} & 18.8 & 21.9 \\
\qquad Sentence & 13 & \textbf{0.0} & \textbf{0.0} & \textbf{0.0} & \textbf{0.0} & \textbf{0.0} & \textbf{0.0} & \textbf{0.0} & \textbf{0.0} & \textbf{0.0} \\
\qquad Words & 45 & 2.2 & 0.0 & 0.0 & 0.0 & 0.0 & 4.4 & 6.7 & 6.7 & \textbf{8.9} \\
\qquad \textit{Global avg.} & 182 & 19.2 & 8.8 & 12.1 & 14.8 & 17.6 & 19.2 & 23.6 & 23.6 & \textbf{29.7} \\
\quad \textsc{Local} & &  &  &  &  &  &  &  &  &  \\
\qquad Count & 24 & 20.8 & 4.2 & 4.2 & 4.2 & 8.3 & 20.8 & 25.0 & 16.7 & \textbf{33.3} \\
\qquad Format & 9 & \textbf{66.7} & 0.0 & 33.3 & 44.4 & 44.4 & 55.6 & \textbf{66.7} & \textbf{66.7} & \textbf{66.7} \\
\qquad Ratio & 12 & \textbf{0.0} & \textbf{0.0} & \textbf{0.0} & \textbf{0.0} & \textbf{0.0} & \textbf{0.0} & \textbf{0.0} & \textbf{0.0} & \textbf{0.0} \\
\qquad Sentence & 11 & 36.4 & 0.0 & 0.0 & 0.0 & 9.1 & \textbf{45.5} & 36.4 & 36.4 & 27.3 \\
\qquad Words & 14 & 14.3 & 0.0 & 7.1 & 7.1 & 7.1 & 7.1 & 0.0 & 7.1 & \textbf{21.4} \\
\qquad \textit{Local avg.} & 70 & 24.3 & 1.4 & 7.1 & 8.6 & 11.4 & 22.9 & 22.9 & 21.4 & \textbf{28.6} \\
\midrule
\multicolumn{11}{l}{\textbf{Qwen2.5-7B}} \\
\quad \textsc{Global} & &  &  &  &  &  &  &  &  &  \\
\qquad Count & 15 & \textbf{66.7} & 40.0 & 26.7 & 40.0 & 33.3 & 60.0 & \textbf{66.7} & 60.0 & 33.3 \\
\qquad Format & 77 & \textbf{55.8} & 14.3 & 24.7 & 27.3 & 28.6 & 50.6 & 48.1 & 48.1 & 53.2 \\
\qquad Ratio & 32 & 18.8 & 15.6 & 15.6 & \textbf{28.1} & 21.9 & 15.6 & 18.8 & 21.9 & 21.9 \\
\qquad Sentence & 13 & 30.8 & 7.7 & 0.0 & 7.7 & 0.0 & 15.4 & 23.1 & 7.7 & \textbf{38.5} \\
\qquad Words & 45 & 26.7 & 0.0 & 4.4 & 17.8 & 15.6 & 26.7 & \textbf{42.2} & \textbf{42.2} & 40.0 \\
\qquad \textit{Global avg.} & 182 & 41.2 & 12.6 & 16.5 & 24.7 & 22.5 & 36.8 & 41.2 & 40.1 & \textbf{41.8} \\
\quad \textsc{Local} & &  &  &  &  &  &  &  &  &  \\
\qquad Count & 24 & 41.7 & 8.3 & 8.3 & 4.2 & 16.7 & 37.5 & \textbf{50.0} & 45.8 & 37.5 \\
\qquad Format & 9 & 22.2 & 0.0 & 0.0 & 0.0 & 0.0 & 22.2 & 33.3 & \textbf{44.4} & 33.3 \\
\qquad Ratio & 12 & \textbf{8.3} & \textbf{8.3} & 0.0 & 0.0 & 0.0 & 0.0 & 0.0 & 0.0 & 0.0 \\
\qquad Sentence & 11 & \textbf{27.3} & 0.0 & 0.0 & 0.0 & 0.0 & 9.1 & \textbf{27.3} & \textbf{27.3} & 18.2 \\
\qquad Words & 14 & 35.7 & 7.1 & 21.4 & 14.3 & 14.3 & 21.4 & 35.7 & 35.7 & \textbf{50.0} \\
\qquad \textit{Local avg.} & 70 & 30.0 & 5.7 & 7.1 & 4.3 & 8.6 & 21.4 & \textbf{32.9} & \textbf{32.9} & 30.0 \\
\midrule
\multicolumn{11}{l}{\textbf{Llama-3.1-8B}} \\
\quad \textsc{Global} & &  &  &  &  &  &  &  &  &  \\
\qquad Count & 15 & 73.3 & 53.3 & 60.0 & 53.3 & 60.0 & 73.3 & \textbf{80.0} & \textbf{80.0} & 73.3 \\
\qquad Format & 77 & 29.9 & 9.1 & 7.8 & 15.6 & 16.9 & 18.2 & 32.5 & 31.2 & \textbf{42.9} \\
\qquad Ratio & 32 & \textbf{25.0} & 15.6 & 18.8 & 18.8 & 21.9 & 15.6 & 18.8 & 18.8 & 15.6 \\
\qquad Sentence & 13 & \textbf{23.1} & 0.0 & 7.7 & 0.0 & 0.0 & 15.4 & 15.4 & 15.4 & 7.7 \\
\qquad Words & 45 & \textbf{0.0} & \textbf{0.0} & \textbf{0.0} & \textbf{0.0} & \textbf{0.0} & \textbf{0.0} & \textbf{0.0} & \textbf{0.0} & \textbf{0.0} \\
\qquad \textit{Global avg.} & 182 & 24.7 & 11.0 & 12.1 & 14.3 & 15.9 & 17.6 & 24.7 & 24.2 & \textbf{27.5} \\
\quad \textsc{Local} & &  &  &  &  &  &  &  &  &  \\
\qquad Count & 24 & 29.2 & 0.0 & 0.0 & 4.2 & 0.0 & 25.0 & 16.7 & \textbf{33.3} & \textbf{33.3} \\
\qquad Format & 9 & \textbf{77.8} & 0.0 & 0.0 & 11.1 & 11.1 & 66.7 & 66.7 & 66.7 & 55.6 \\
\qquad Ratio & 12 & 0.0 & \textbf{8.3} & 0.0 & 0.0 & 0.0 & 0.0 & 0.0 & 0.0 & 0.0 \\
\qquad Sentence & 11 & \textbf{27.3} & 0.0 & 0.0 & 0.0 & 9.1 & \textbf{27.3} & 18.2 & 18.2 & 18.2 \\
\qquad Words & 14 & 14.3 & 0.0 & 0.0 & 14.3 & 14.3 & 14.3 & \textbf{35.7} & 28.6 & 28.6 \\
\qquad \textit{Local avg.} & 70 & 27.1 & 1.4 & 0.0 & 5.7 & 5.7 & 24.3 & 24.3 & \textbf{28.6} & 27.1 \\
\bottomrule
\end{tabular}}

\end{table*}
\begin{table*}[htbp]
\centering
\caption{\textbf{Instruction Following by Type on \ttt{IFBench}.} Results for Phi-4-14B, LFM2.5-1.2B, and Ministral-8B, using the same metrics and columns as \cref{tab:if-full-ifbench}.}
\label{tab:if-full-ifbench-b}
\vspace{6pt}
\resizebox{\textwidth}{!}{%
\begin{tabular}{lrccccccccc}
\toprule
 & & & \multicolumn{4}{c}{\methodL{} ($\alpha$)} & \multicolumn{4}{c}{P+\methodL{} ($\alpha$)} \\
\cmidrule(lr){4-7} \cmidrule(lr){8-11}
Instruction & $n$ & Prompting & $0.5$ & $1$ & $1.5$ & $2$ & $0.5$ & $1$ & $1.5$ & $2$ \\
\midrule
\multicolumn{11}{l}{\textbf{Phi-4-14B}} \\
\quad \textsc{Global} & &  &  &  &  &  &  &  &  &  \\
\qquad Count & 15 & \textbf{80.0} & 60.0 & 60.0 & 53.3 & 46.7 & 60.0 & 60.0 & 66.7 & 53.3 \\
\qquad Format & 77 & 20.8 & 7.8 & 7.8 & 16.9 & 18.2 & 22.1 & 35.1 & 36.4 & \textbf{39.0} \\
\qquad Ratio & 32 & 18.8 & 18.8 & 18.8 & 18.8 & \textbf{21.9} & \textbf{21.9} & \textbf{21.9} & \textbf{21.9} & \textbf{21.9} \\
\qquad Sentence & 13 & \textbf{7.7} & \textbf{7.7} & 0.0 & 0.0 & 0.0 & \textbf{7.7} & 0.0 & 0.0 & \textbf{7.7} \\
\qquad Words & 45 & \textbf{0.0} & \textbf{0.0} & \textbf{0.0} & \textbf{0.0} & \textbf{0.0} & \textbf{0.0} & \textbf{0.0} & \textbf{0.0} & \textbf{0.0} \\
\qquad \textit{Global avg.} & 182 & 19.2 & 12.1 & 11.5 & 14.8 & 15.4 & 18.7 & 23.6 & 24.7 & \textbf{25.3} \\
\quad \textsc{Local} & &  &  &  &  &  &  &  &  &  \\
\qquad Count & 24 & 37.5 & 4.2 & 8.3 & 4.2 & 4.2 & 29.2 & 41.7 & 41.7 & \textbf{50.0} \\
\qquad Format & 9 & 33.3 & 0.0 & 11.1 & 33.3 & 33.3 & 44.4 & 44.4 & \textbf{55.6} & \textbf{55.6} \\
\qquad Ratio & 12 & 0.0 & 0.0 & 0.0 & 0.0 & \textbf{8.3} & 0.0 & 0.0 & 0.0 & 0.0 \\
\qquad Sentence & 11 & 9.1 & 0.0 & 0.0 & 0.0 & 0.0 & 9.1 & \textbf{18.2} & \textbf{18.2} & \textbf{18.2} \\
\qquad Words & 14 & 0.0 & \textbf{7.1} & 0.0 & 0.0 & 0.0 & \textbf{7.1} & 0.0 & 0.0 & \textbf{7.1} \\
\qquad \textit{Local avg.} & 70 & 18.6 & 2.9 & 4.3 & 5.7 & 7.1 & 18.6 & 22.9 & 24.3 & \textbf{28.6} \\
\midrule
\multicolumn{11}{l}{\textbf{LFM2.5-1.2B}} \\
\quad \textsc{Global} & &  &  &  &  &  &  &  &  &  \\
\qquad Count & 15 & \textbf{80.0} & 66.7 & 66.7 & 73.3 & 66.7 & 73.3 & 66.7 & 66.7 & 60.0 \\
\qquad Format & 77 & \textbf{67.5} & 10.4 & 20.8 & 22.1 & 15.6 & 64.9 & 58.4 & 54.5 & 48.1 \\
\qquad Ratio & 32 & 28.1 & 15.6 & 28.1 & 25.0 & 31.2 & 28.1 & 25.0 & \textbf{37.5} & 31.2 \\
\qquad Sentence & 13 & 46.2 & 7.7 & 38.5 & 53.8 & 53.8 & 23.1 & 53.8 & \textbf{61.5} & \textbf{61.5} \\
\qquad Words & 45 & 6.7 & 0.0 & 2.2 & 6.7 & 11.1 & 11.1 & 11.1 & 8.9 & \textbf{13.3} \\
\qquad \textit{Global avg.} & 182 & \textbf{45.1} & 13.2 & 22.5 & 25.3 & 24.2 & 42.9 & 41.2 & 41.8 & 38.5 \\
\quad \textsc{Local} & &  &  &  &  &  &  &  &  &  \\
\qquad Count & 24 & \textbf{62.5} & 8.3 & 0.0 & 4.2 & 8.3 & 50.0 & \textbf{62.5} & 58.3 & 58.3 \\
\qquad Format & 9 & \textbf{66.7} & 0.0 & 0.0 & 0.0 & 0.0 & 22.2 & 33.3 & 44.4 & 33.3 \\
\qquad Ratio & 12 & \textbf{8.3} & \textbf{8.3} & 0.0 & 0.0 & 0.0 & 0.0 & 0.0 & 0.0 & \textbf{8.3} \\
\qquad Sentence & 11 & \textbf{18.2} & 0.0 & 0.0 & 0.0 & 0.0 & \textbf{18.2} & 9.1 & \textbf{18.2} & 0.0 \\
\qquad Words & 14 & 71.4 & 21.4 & 57.1 & 50.0 & 50.0 & 71.4 & \textbf{78.6} & \textbf{78.6} & \textbf{78.6} \\
\qquad \textit{Local avg.} & 70 & \textbf{48.6} & 8.6 & 11.4 & 11.4 & 12.9 & 37.1 & 42.9 & 44.3 & 41.4 \\
\midrule
\multicolumn{11}{l}{\textbf{Ministral-8B}} \\
\quad \textsc{Global} & &  &  &  &  &  &  &  &  &  \\
\qquad Count & 15 & \textbf{60.0} & 46.7 & 53.3 & \textbf{60.0} & 40.0 & 46.7 & 53.3 & 53.3 & 53.3 \\
\qquad Format & 77 & 20.8 & 6.5 & 11.7 & 11.7 & 13.0 & 19.5 & 22.1 & 24.7 & \textbf{31.2} \\
\qquad Ratio & 32 & 15.6 & 18.8 & \textbf{21.9} & \textbf{21.9} & 18.8 & 18.8 & \textbf{21.9} & \textbf{21.9} & \textbf{21.9} \\
\qquad Sentence & 13 & 23.1 & 15.4 & 7.7 & 23.1 & \textbf{38.5} & 15.4 & 23.1 & 0.0 & 7.7 \\
\qquad Words & 45 & 2.2 & 0.0 & 0.0 & 0.0 & 2.2 & 2.2 & 2.2 & 4.4 & \textbf{6.7} \\
\qquad \textit{Global avg.} & 182 & 18.7 & 11.0 & 13.7 & 15.4 & 15.4 & 17.0 & 19.8 & 19.8 & \textbf{23.6} \\
\quad \textsc{Local} & &  &  &  &  &  &  &  &  &  \\
\qquad Count & 24 & 20.8 & 0.0 & 8.3 & 0.0 & 0.0 & 16.7 & 20.8 & 25.0 & \textbf{33.3} \\
\qquad Format & 9 & 22.2 & 0.0 & 0.0 & \textbf{44.4} & \textbf{44.4} & 22.2 & 22.2 & \textbf{44.4} & \textbf{44.4} \\
\qquad Ratio & 12 & \textbf{8.3} & \textbf{8.3} & \textbf{8.3} & \textbf{8.3} & 0.0 & \textbf{8.3} & 0.0 & 0.0 & 0.0 \\
\qquad Sentence & 11 & 9.1 & 0.0 & 0.0 & 0.0 & 0.0 & 18.2 & 9.1 & 0.0 & \textbf{27.3} \\
\qquad Words & 14 & 0.0 & 0.0 & 0.0 & 0.0 & \textbf{21.4} & 0.0 & 0.0 & 0.0 & 0.0 \\
\qquad \textit{Local avg.} & 70 & 12.9 & 1.4 & 4.3 & 7.1 & 10.0 & 12.9 & 11.4 & 14.3 & \textbf{21.4} \\
\bottomrule
\end{tabular}}

\end{table*}

\subsection{Training and Inference Cost}
\label{app:results-cost}

\textbf{Evaluation Setup.}
All timings use Gemma-2-9B on an H200 GPU, 128 new tokens per request with end-of-sequence ignored, sampling at temperature 1, and prompts from \ttt{FLAS}'s evaluation set with one concept per row.
\emph{Training} cost is the GPU time of each recipe (\cref{tab:cost-train-main}).
We measure \ttt{HyperSteer} and \ttt{FLAS} by retraining each with its released recipe.
\emph{\ttt{vLLM} serving} runs every method in one session with continuous batching.
The steering methods run inside the engine's CUDA graphs.
CFG evaluates the two contexts of a request as a pair of sequences in the same batch.
\ttt{FLAS} is our port of its released checkpoint (three flow steps at layer 20, flow time 2), with each concept encoded once before serving.
For a fair comparison we optimized this port as far as \method{}'s own path: its flow runs on \ttt{vLLM}'s attention kernels inside the same CUDA graphs.
\emph{Batched Hugging~Face} inference runs every method in one process.
In both settings, we interleave methods, run untimed warm-ups, and take every ratio against the base LLM measured in the same session.
CFG evaluates both contexts of each request in one batch of twice the rows, and \ttt{FLAS} uses its released checkpoint and generation loop.
\method{} runs at $\alpha{=}2$ throughout, the fixed adherence level used in the main comparison.

\textbf{Inference Efficiency.}
In \ttt{vLLM}, the per-token inference time of \method{} matches that of the base LLM at every batch size, and \method{} retains 96--99\% of the base throughput, compared with 41--86\% for CFG and 84--86\% for \ttt{FLAS}.
The per-token inference time of \ttt{FLAS} is 12--19\% slower even though we also run it inside \ttt{vLLM}'s CUDA graphs.
The first token of \method{} arrives later than that of the base LLM ($\times$1.17--1.45 from batch size 32), because it edits every prompt position, but earlier than those of \ttt{FLAS} ($\times$1.31--1.51) and CFG ($\times$2.34--2.62).
In batched Hugging~Face inference, \method{} retains 95--99\% of the base throughput, compared with 77--91\% for \ttt{FLAS}, and CFG runs out of memory at batch size 256 (\cref{tab:cost-train-main,tab:cost,tab:cost-infer-vllm}).

\textbf{Training Scalability.}
Training \method{} requires $4.65\times$ fewer GPU-hours than \ttt{FLAS} and $25\times$ fewer than \ttt{HyperSteer}, with $12.4\times$ fewer parameters than \ttt{FLAS} and without its 72 \ttt{gpt-4o-mini} completions per concept.
A second and third epoch do not improve performance (\cref{tab:train-abl}).
On \ttt{SteerIF}, which has 6.6$\times$ more instructions than \ttt{Concept16k} and longer responses, training \methodL{} takes 99.5, 58.7, 72.1, 150.3, 30.4, and 65.8 GPU-hours on Gemma-2-9B, Qwen2.5-7B, Llama-3.1-8B, Phi-4-14B, LFM2.5-1.2B, and Ministral-8B, and \methodLite{} takes 93.5 on Gemma-2-9B.

\Cref{tab:cost-infer-vllm,tab:cost-infer-hf-batched} detail the costs summarized in \cref{tab:cost-train-main,tab:cost}: \ttt{vLLM} and batched Hugging~Face serving with absolute values, separate prefill and per-token inference measurements, and time to first token.

\begin{table}[htbp]
  \centering\small
  \caption{\textbf{Inference Efficiency in \ttt{vLLM}.} Absolute measurements on Gemma-2-9B on an H200.
  TTFT is time to first token, and end to end covers the full request.}
  \label{tab:cost-infer-vllm}
  \vspace{6pt}
  \begin{tabular}{llcccc}
    \toprule
    Method & Params & B=1 & B=32 & B=128 & B=256 \\
    \midrule
    \multicolumn{6}{l}{\textit{Throughput, tokens/s}} \\
      Base & -- & 149 & 4{,}057 & 10{,}506 & 12{,}863 \\
      CFG ($\alpha{=}2$) & -- & 128 & 2{,}685 & 4{,}902 & 5{,}239 \\
      \ttt{FLAS} & 255.6M & 126 & 3{,}391 & 8{,}963 & 11{,}008 \\
      \method{} & 20.6M & 148 & 3{,}887 & 10{,}262 & 12{,}464 \\
      \methodLite{} & 4.08M & 148 & 3{,}938 & 10{,}183 & 12{,}432 \\
      \methodL{} & 20.6M & 148 & 3{,}892 & 10{,}131 & 12{,}479 \\
    \midrule
    \multicolumn{6}{l}{\textit{Inference step, ms (mean)}} \\
      Base & -- & 6.66 & 7.54 & 10.78 & 17.09 \\
      CFG ($\alpha{=}2$) & -- & 7.69 & 10.90 & 22.26 & 41.00 \\
      \ttt{FLAS} & 255.6M & 7.82 & 8.97 & 12.37 & 19.22 \\
      \method{} & 20.6M & 6.66 & 7.52 & 10.67 & 17.01 \\
      \methodLite{} & 4.08M & 6.65 & 7.50 & 10.80 & 17.19 \\
      \methodL{} & 20.6M & 6.67 & 7.52 & 10.78 & 16.99 \\
    \midrule
    \multicolumn{6}{l}{\textit{TTFT, ms (median per request)}} \\
      Base & -- & 9 & 47 & 200 & 391 \\
      CFG ($\alpha{=}2$) & -- & 11 & 110 & 506 & 1{,}026 \\
      \ttt{FLAS} & 255.6M & 19 & 71 & 261 & 543 \\
      \method{} & 20.6M & 17 & 68 & 234 & 466 \\
      \methodLite{} & 4.08M & 16 & 63 & 238 & 455 \\
      \methodL{} & 20.6M & 17 & 68 & 237 & 465 \\
    \midrule
    \multicolumn{6}{l}{\textit{End to end, ms per 128-token request}} \\
      Base & -- & 860 & 1{,}010 & 1{,}561 & 2{,}553 \\
      CFG ($\alpha{=}2$) & -- & 998 & 1{,}528 & 3{,}346 & 6{,}268 \\
      \ttt{FLAS} & 255.6M & 1{,}018 & 1{,}213 & 1{,}831 & 2{,}977 \\
      \method{} & 20.6M & 866 & 1{,}055 & 1{,}602 & 2{,}629 \\
      \methodLite{} & 4.08M & 863 & 1{,}041 & 1{,}610 & 2{,}638 \\
      \methodL{} & 20.6M & 867 & 1{,}054 & 1{,}618 & 2{,}630 \\
    \bottomrule
  \end{tabular}
\end{table}

\begin{table}[htbp]
  \centering\small
  \caption{\textbf{Batched Inference Efficiency in Hugging~Face.} Gemma-2-9B on an H200.
  Throughput is reported in tokens/s.
  Prefill and per-token inference times are relative to the base LLM.}
  \label{tab:cost-infer-hf-batched}
  \vspace{6pt}
  \begin{tabular}{llcccc}
    \toprule
    Method & Params & B=1 & B=32 & B=128 & B=256 \\
    \midrule
    \multicolumn{6}{l}{\textit{Throughput, tokens/s}} \\
      Base & -- & 33.1 & 880 & 1{,}556 & 1{,}163 \\
      CFG ($\alpha{=}2$) & -- & 29.3 & 643 & 802 & OOM \\
      \ttt{FLAS} & 255.6M & 25.5 & 699 & 1{,}415 & 1{,}063 \\
      \methodLite{} & 4.08M & 31.3 & 826 & 1{,}534 & 1{,}155 \\
      \methodL{} & 20.6M & 31.3 & 833 & 1{,}535 & 1{,}151 \\
    \midrule
    \multicolumn{6}{l}{\textit{Prefill, relative to base}} \\
      Base & -- & 1.00$\times$ & 1.00$\times$ & 1.00$\times$ & 1.00$\times$ \\
      CFG ($\alpha{=}2$) & -- & 1.25$\times$ & 2.11$\times$ & 2.19$\times$ & OOM \\
      \ttt{FLAS} & 255.6M & 1.30$\times$ & 1.13$\times$ & 1.10$\times$ & 1.11$\times$ \\
      \methodLite{} & 4.08M & 1.81$\times$ & 1.35$\times$ & 1.07$\times$ & 1.04$\times$ \\
      \methodL{} & 20.6M & 1.88$\times$ & 1.36$\times$ & 1.07$\times$ & 1.05$\times$ \\
    \midrule
    \multicolumn{6}{l}{\textit{Inference per token, relative to base}} \\
      Base & -- & 1.00$\times$ & 1.00$\times$ & 1.00$\times$ & 1.00$\times$ \\
      CFG ($\alpha{=}2$) & -- & 1.13$\times$ & 1.30$\times$ & 1.90$\times$ & OOM \\
      \ttt{FLAS} & 255.6M & 1.30$\times$ & 1.27$\times$ & 1.10$\times$ & 1.09$\times$ \\
      \methodLite{} & 4.08M & 1.05$\times$ & 1.04$\times$ & 1.00$\times$ & 1.00$\times$ \\
      \methodL{} & 20.6M & 1.04$\times$ & 1.03$\times$ & 1.00$\times$ & 1.00$\times$ \\
    \bottomrule
  \end{tabular}
\end{table}

\subsection{Statistical Significance}
\label{app:paired}
\Cref{tab:paired} reports the main comparisons as paired differences.
On \ttt{AxBench}, prompt+\method{} and prompt+\methodL{} improve over prompting and \ttt{FLAS} on both splits, with intervals well above zero and wins on most concepts.
\method{} alone scores above \ttt{FLAS} on 304 of 500 held-in and 63 of 100 held-out concepts; the held-out interval includes zero.
\method{} alone and prompting are on par.
On \ttt{IFEval}, prompt+\methodL{} at $\alpha{=}1$ gains more prompts than it loses on five of the six LLMs, with intervals above zero on Gemma-2-9B, Phi-4-14B, and Ministral-8B.
\begin{table}[t]
  \centering\small
  \setlength{\tabcolsep}{4pt}\renewcommand{\arraystretch}{1.1}
  \caption{\textbf{Statistical Significance.} Top: differences in \ttt{AxBench} overall score (\ttt{gpt-4o-mini}) on Gemma-2-9B with 95\% bootstrap intervals over concepts; W/L/T counts the concepts on which the first method scores higher, lower, or the same.
  Bottom: differences in \ttt{IFEval} prompt-level strict accuracy (\%) with 95\% intervals over prompts; gained and lost count the prompts that only prompt+\methodL{} or only prompting passes.}
  \label{tab:paired}
  \vspace{6pt}
  \resizebox{\linewidth}{!}{%
  \begin{tabular}{@{}lcccc@{}}
    \toprule
    & \multicolumn{2}{c}{Held-In} & \multicolumn{2}{c}{Held-Out} \\
    \cmidrule(lr){2-3}\cmidrule(l){4-5}
    Comparison & Difference & W/L/T & Difference & W/L/T \\
    \midrule
    \method{} ($\alpha{=}2$) $-$ \ttt{FLAS} & $+0.062$ [$+0.032$, $+0.093$] & 304/170/26 & $+0.049$ [$-0.006$, $+0.101$] & 63/34/3 \\
    \method{} ($\alpha{=}2$) $-$ Prompting & $+0.010$ [$-0.027$, $+0.047$] & 255/228/17 & $-0.024$ [$-0.092$, $+0.043$] & 48/50/2 \\
    Prompt + \method{} ($\alpha{=}0.5$) $-$ Prompting & $+0.059$ [$+0.037$, $+0.081$] & 264/182/54 & $+0.051$ [$+0.015$, $+0.089$] & 55/45/0 \\
    Prompt + \method{} ($\alpha{=}1$) $-$ Prompting & $+0.149$ [$+0.126$, $+0.174$] & 331/131/38 & $+0.158$ [$+0.115$, $+0.202$] & 74/26/0 \\
    Prompt + \method{} ($\alpha{=}1$) $-$ \ttt{FLAS} & $+0.202$ [$+0.174$, $+0.229$] & 360/118/22 & $+0.231$ [$+0.175$, $+0.288$] & 85/14/1 \\
    Prompt + \methodL{} ($\alpha{=}1$) $-$ Prompting & $+0.147$ [$+0.123$, $+0.171$] & 338/132/30 & $+0.128$ [$+0.084$, $+0.173$] & 70/29/1 \\
    \midrule
    \multicolumn{5}{@{}l}{\textit{Prompt + \methodL{} ($\alpha{=}1$) $-$ Prompting on \ttt{IFEval}}} \\
    & Difference & Gained/Lost & & \\
    Gemma-2-9B & $+4.1$ [$+1.1$, $+7.1$] & 46/24 & & \\
    Qwen2.5-7B & $-0.4$ [$-3.8$, $+3.1$] & 45/47 & & \\
    Llama-3.1-8B & $+2.0$ [$-1.7$, $+5.7$] & 58/47 & & \\
    Phi-4-14B & $+5.0$ [$+1.4$, $+8.5$] & 62/35 & & \\
    LFM2.5-1.2B & $+2.2$ [$-0.2$, $+4.7$] & 29/17 & & \\
    Ministral-8B & $+11.6$ [$+8.1$, $+15.2$] & 84/21 & & \\
    \bottomrule
  \end{tabular}}
\end{table}

\Cref{tab:seeds} repeats the held-out comparison for three training seeds.
The standard deviation across seeds is at most 0.013, except for \method{} at $\alpha{=}1$ (0.046).
On every seed, \method{} at $\alpha{=}2$ stays above \ttt{FLAS} (1.032--1.046 against 0.987), and prompt+\method{} at $\alpha{=}1$ stays above prompting (1.207--1.232 against 1.060).
\begin{table}[t]
  \centering\small
  \setlength{\tabcolsep}{5pt}\renewcommand{\arraystretch}{1.1}
  \caption{\textbf{Training Seeds.} Overall score on \ttt{FLAS}'s 100 held-out concepts for \method{} trained with three seeds on Gemma-2-9B (mean $\pm$ standard deviation).
  \cref{tab:main-axb-9b} reports the first seed.}
  \label{tab:seeds}
  \vspace{6pt}
  \begin{tabular}{@{}lcc@{}}
    \toprule
    Method & \ttt{gpt-4o-mini} & Median of three judges \\
    \midrule
    \method{} ($\alpha{=}1$) & 0.741 $\pm$ 0.046 & 0.934 $\pm$ 0.065 \\
    \method{} ($\alpha{=}2$) & 1.038 $\pm$ 0.007 & 1.245 $\pm$ 0.012 \\
    Prompt + \method{} ($\alpha{=}0.5$) & 1.116 $\pm$ 0.006 & 1.357 $\pm$ 0.008 \\
    Prompt + \method{} ($\alpha{=}1$) & 1.218 $\pm$ 0.013 & 1.480 $\pm$ 0.019 \\
    \bottomrule
  \end{tabular}
\end{table}

    \FloatBarrier
    \section{Ablations}
\label{app:ablations}

\subsection{Steering Layers}
\label{app:abl-layers}
\textbf{Ablating Steering Layers.}
\Cref{tab:layer-sweep} reports \method{} trained separately at several layers of Gemma-2-9B and Gemma-2-2B and scored at $\alpha{=}2$ with \ttt{Phi-4-14B}.
Both score best at their default half-depth layer (layer 20 of 42 and layer 12 of 26).
On 9B, layers 16 and 20 tie on every split, layer 12 loses 0.069 on \ttt{Concept500}, and layer 30 loses 0.499.
On 2B, every deeper layer loses (0.091 at layer 16, 0.269 at layer 20).
On 2B, the training JSD follows the same order.
On 9B, layers 12--20 fit the teacher similarly (JSD 0.055--0.059), and only layer 30 fits clearly worse (0.096).
On 2B, \ttt{FLAS}'s released 2B checkpoint scores above \method{} at every layer under this judge (1.090 against 0.907 on \ttt{Concept500}).

\textbf{Steering Multiple Layers.}
Editing four consecutive 2B layers (12--15) instead of one adds 0.057 on \ttt{Concept500}, whereas eight layers (12--19) add nothing over four, and neither reaches \ttt{FLAS}-2B except on the held-out concepts, where four layers tie it ($-0.031$).
\begin{table}[htbp]
  \centering\small
  \setlength{\tabcolsep}{4pt}
  \caption{\textbf{Steering Layer Ablation.} Overall scores at $\alpha=2$ with \ttt{Phi-4-14B} as judge, using the last five requests per concept and 128-token responses.
  Each \method{} row is trained for one epoch at the listed layer.
  L12--15 and L12--19 edit four and eight layers while reading the instruction at layer 12.
  Layers are counted from zero; depth is (layer $+\,1$) divided by the number of layers.
  $\Delta$ is the paired difference from the default layer over the concepts judged in both runs, so it can differ from the difference of the displayed scores.
  JSD reports the training loss from the one-epoch run, where lower is better.
  \ttt{FLAS} uses its released checkpoint at $T=2$.
  Held-in and held-out refer to \ttt{FLAS}'s concept splits.}
  \label{tab:layer-sweep}
  \vspace{6pt}
  \begin{tabular}{lccccccc}
    \toprule
    Layer (depth) & \multicolumn{2}{c}{\ttt{Concept500}} & \multicolumn{2}{c}{\ttt{FLAS} held-out} & \multicolumn{2}{c}{\ttt{FLAS} held-in} & JSD \\
    \cmidrule(lr){2-3}\cmidrule(lr){4-5}\cmidrule(lr){6-7}
    & score & $\Delta$ & score & $\Delta$ & score & $\Delta$ & \\
    \midrule
    \multicolumn{8}{l}{\textit{Gemma-2-9B (42 layers; default layer 20)}} \\
    L12 (31\%) & 1.246 & $-$0.069 & 1.244 & $-$0.010 & 1.264 & $-$0.076 & 0.0569 \\
    L16 (40\%) & 1.316 & $-$0.005 & 1.244 & +0.000 & 1.333 & $-$0.008 & 0.0546 \\
    L20 (50\%), default & 1.320 & -- & 1.253 & -- & 1.339 & -- & 0.0585 \\
    L30 (74\%) & 0.821 & $-$0.499 & 0.843 & $-$0.416 & 0.818 & $-$0.518 & 0.0964 \\
    \ttt{FLAS} ($T{=}2$) & 1.183 &  & 1.193 &  & -- &  & -- \\
    \midrule
    \multicolumn{8}{l}{\textit{Gemma-2-2B (26 layers; default layer 12)}} \\
    L12 (50\%), default & 0.907 & -- & 0.933 & -- & 0.924 & -- & 0.0481 \\
    L16 (65\%) & 0.817 & $-$0.091 & 0.824 & $-$0.109 & 0.820 & $-$0.104 & 0.0590 \\
    L20 (81\%) & 0.638 & $-$0.269 & 0.611 & $-$0.322 & 0.674 & $-$0.250 & 0.0667 \\
    L12--15 (4 layers) & 0.964 & +0.057 & 0.986 & +0.053 & 1.000 & +0.076 & 0.0470 \\
    L12--19 (8 layers) & 0.957 & +0.050 & 0.985 & +0.052 & 0.951 & +0.027 & 0.0462 \\
    \ttt{FLAS}-2B ($T{=}2$) & 1.090 &  & 1.017 &  & 1.072 &  & -- \\
    \bottomrule
  \end{tabular}
\end{table}

\subsection{Adapter Architecture}
\label{app:abl-arch}
\textbf{\methodL{} against \methodLite{}.}
\Cref{tab:if-lite-ifeval,tab:if-lite-ifbench} compare the two instruction-following checkpoints of each LLM, trained on the same data.
\methodLite{} omits the read-out paths (\cref{app:arch}), so the comparison isolates their contribution on \ttt{IFEval} and \ttt{IFBench}.
\begin{table}[htbp]
  \centering\small
  \setlength{\tabcolsep}{3pt}\renewcommand{\arraystretch}{1.05}
  \caption{\textbf{Steering Adapter Ablation on \ttt{IFEval}.} Instruction-level strict accuracy (\%) for \methodL{} and \methodLite{} (Lite).
  Both adapters use the same training data.
  Results continue in \cref{tab:if-lite-ifeval-b}.}
  \label{tab:if-lite-ifeval}
  \vspace{6pt}
  \begin{tabular*}{\linewidth}{@{\extracolsep{\fill}}lrcccccc@{}}
    \toprule
    & & \multicolumn{2}{c}{\shortstack{Without prompting\\$\alpha=2$}} & \multicolumn{2}{c}{\shortstack{With prompting\\$\alpha=0.5$}} & \multicolumn{2}{c}{\shortstack{With prompting\\$\alpha=1$}} \\
    \cmidrule(lr){3-4}\cmidrule(lr){5-6}\cmidrule(l){7-8}
    Instruction & $n$ & Lite & \methodL{} & Lite & \methodL{} & Lite & \methodL{} \\
    \midrule
    \multicolumn{8}{@{}l}{\textbf{Gemma-2-9B}} \\
    \quad Case & 89 & 86.5 & 85.4 & 78.7 & 77.5 & 88.8 & 87.6 \\
    \quad Language & 31 & 16.1 & 12.9 & 93.5 & 93.5 & 93.5 & 96.8 \\
    \quad Punctuation & 66 & 93.9 & 87.9 & 95.5 & 97.0 & 97.0 & 97.0 \\
    \quad Length & 131 & 58.8 & 64.9 & 61.8 & 67.2 & 71.0 & 69.5 \\
    \quad Format & 157 & 72.6 & 73.9 & 88.5 & 90.4 & 90.4 & 89.8 \\
    \quad Wrapping & 65 & 86.2 & 84.6 & 100.0 & 98.5 & 98.5 & 95.4 \\
    \quad \textit{Global avg.} & 539 & 72.5 & 73.1 & 82.9 & 84.6 & 87.4 & 86.5 \\
    \quad Keywords & 163 & 41.1 & 46.0 & 78.5 & 81.0 & 80.4 & 79.8 \\
    \quad Positions & 38 & 5.3 & 10.5 & 86.8 & 86.8 & 78.9 & 71.1 \\
    \quad Markers & 53 & 58.5 & 84.9 & 86.8 & 90.6 & 90.6 & 94.3 \\
    \quad Copying & 41 & 41.5 & 46.3 & 73.2 & 70.7 & 61.0 & 63.4 \\
    \quad \textit{Local avg.} & 295 & 39.7 & 48.5 & 80.3 & 82.0 & 79.3 & 79.0 \\
    \midrule
    \multicolumn{8}{@{}l}{\textbf{Qwen2.5-7B}} \\
    \quad Case & 89 & 80.9 & 76.4 & 77.5 & 76.4 & 85.4 & 82.0 \\
    \quad Language & 31 & 19.4 & 16.1 & 96.8 & 96.8 & 90.3 & 71.0 \\
    \quad Punctuation & 66 & 100.0 & 98.5 & 95.5 & 98.5 & 97.0 & 100.0 \\
    \quad Length & 131 & 61.1 & 58.8 & 65.6 & 61.1 & 70.2 & 70.2 \\
    \quad Format & 157 & 83.4 & 84.1 & 88.5 & 90.4 & 89.8 & 92.4 \\
    \quad Wrapping & 65 & 50.8 & 55.4 & 87.7 & 87.7 & 72.3 & 73.8 \\
    \quad \textit{Global avg.} & 539 & 72.0 & 71.1 & 82.4 & 82.0 & 83.1 & 82.7 \\
    \quad Keywords & 163 & 34.4 & 34.4 & 76.7 & 77.3 & 77.3 & 77.9 \\
    \quad Positions & 38 & 10.5 & 13.2 & 71.1 & 65.8 & 76.3 & 65.8 \\
    \quad Markers & 53 & 62.3 & 79.2 & 90.6 & 84.9 & 94.3 & 86.8 \\
    \quad Copying & 41 & 26.8 & 43.9 & 56.1 & 58.5 & 56.1 & 61.0 \\
    \quad \textit{Local avg.} & 295 & 35.3 & 41.0 & 75.6 & 74.6 & 77.3 & 75.6 \\
    \midrule
    \multicolumn{8}{@{}l}{\textbf{Llama-3.1-8B}} \\
    \quad Case & 89 & 86.5 & 82.0 & 86.5 & 86.5 & 85.4 & 82.0 \\
    \quad Language & 31 & 9.7 & 32.3 & 87.1 & 93.5 & 77.4 & 93.5 \\
    \quad Punctuation & 66 & 90.9 & 93.9 & 93.9 & 92.4 & 98.5 & 95.5 \\
    \quad Length & 131 & 58.8 & 63.4 & 77.9 & 81.7 & 83.2 & 80.2 \\
    \quad Format & 157 & 66.2 & 67.5 & 88.5 & 84.7 & 85.4 & 85.4 \\
    \quad Wrapping & 65 & 46.2 & 58.5 & 92.3 & 93.8 & 72.3 & 73.8 \\
    \quad \textit{Global avg.} & 539 & 65.1 & 69.0 & 86.6 & 86.8 & 84.4 & 83.9 \\
    \quad Keywords & 163 & 38.0 & 37.4 & 82.2 & 82.2 & 81.6 & 83.4 \\
    \quad Positions & 38 & 13.2 & 7.9 & 81.6 & 84.2 & 81.6 & 76.3 \\
    \quad Markers & 53 & 66.0 & 84.9 & 94.3 & 94.3 & 96.2 & 98.1 \\
    \quad Copying & 41 & 36.6 & 34.1 & 61.0 & 58.5 & 63.4 & 56.1 \\
    \quad \textit{Local avg.} & 295 & 39.7 & 41.7 & 81.4 & 81.4 & 81.7 & 81.4 \\
    \bottomrule
  \end{tabular*}
\end{table}

\begin{table}[htbp]
  \centering\small
  \setlength{\tabcolsep}{3pt}\renewcommand{\arraystretch}{1.05}
  \caption{\textbf{Steering Adapter Ablation on \ttt{IFEval} (Continued).} Phi-4-14B and LFM2.5-1.2B, using the same settings and metrics as \cref{tab:if-lite-ifeval}.}
  \label{tab:if-lite-ifeval-b}
  \vspace{6pt}
  \begin{tabular*}{\linewidth}{@{\extracolsep{\fill}}lrcccccc@{}}
    \toprule
    & & \multicolumn{2}{c}{\shortstack{Without prompting\\$\alpha=2$}} & \multicolumn{2}{c}{\shortstack{With prompting\\$\alpha=0.5$}} & \multicolumn{2}{c}{\shortstack{With prompting\\$\alpha=1$}} \\
    \cmidrule(lr){3-4}\cmidrule(lr){5-6}\cmidrule(l){7-8}
    Instruction & $n$ & Lite & \methodL{} & Lite & \methodL{} & Lite & \methodL{} \\
    \midrule
    \multicolumn{8}{@{}l}{\textbf{Phi-4-14B}} \\
    \quad Case & 89 & 79.8 & 76.4 & 70.8 & 55.1 & 80.9 & 69.7 \\
    \quad Language & 31 & 25.8 & 9.7 & 96.8 & 93.5 & 93.5 & 87.1 \\
    \quad Punctuation & 66 & 19.7 & 18.2 & 16.7 & 19.7 & 24.2 & 19.7 \\
    \quad Length & 131 & 63.4 & 67.9 & 67.2 & 69.5 & 74.0 & 75.6 \\
    \quad Format & 157 & 75.8 & 79.6 & 89.8 & 86.6 & 92.4 & 86.6 \\
    \quad Wrapping & 65 & 66.2 & 55.4 & 89.2 & 84.6 & 89.2 & 83.1 \\
    \quad \textit{Global avg.} & 539 & 62.5 & 61.8 & 72.5 & 69.2 & 77.4 & 72.5 \\
    \quad Keywords & 163 & 39.3 & 39.9 & 78.5 & 74.8 & 79.8 & 81.6 \\
    \quad Positions & 38 & 7.9 & 7.9 & 78.9 & 84.2 & 78.9 & 76.3 \\
    \quad Markers & 53 & 60.4 & 69.8 & 88.7 & 90.6 & 92.5 & 92.5 \\
    \quad Copying & 41 & 41.5 & 39.0 & 63.4 & 68.3 & 73.2 & 82.9 \\
    \quad \textit{Local avg.} & 295 & 39.3 & 41.0 & 78.3 & 78.0 & 81.0 & 83.1 \\
    \midrule
    \multicolumn{8}{@{}l}{\textbf{LFM2.5-1.2B}} \\
    \quad Case & 89 & 91.0 & 89.9 & 91.0 & 92.1 & 92.1 & 92.1 \\
    \quad Language & 31 & 16.1 & 12.9 & 90.3 & 90.3 & 87.1 & 77.4 \\
    \quad Punctuation & 66 & 93.9 & 97.0 & 100.0 & 100.0 & 100.0 & 100.0 \\
    \quad Length & 131 & 77.9 & 76.3 & 84.0 & 86.3 & 88.5 & 88.5 \\
    \quad Format & 157 & 77.1 & 82.2 & 94.9 & 95.5 & 91.7 & 94.9 \\
    \quad Wrapping & 65 & 55.4 & 69.2 & 93.8 & 93.8 & 92.3 & 95.4 \\
    \quad \textit{Global avg.} & 539 & 75.5 & 78.3 & 91.8 & 92.8 & 91.8 & 92.6 \\
    \quad Keywords & 163 & 51.5 & 52.1 & 88.3 & 89.6 & 91.4 & 90.2 \\
    \quad Positions & 38 & 10.5 & 10.5 & 94.7 & 97.4 & 92.1 & 97.4 \\
    \quad Markers & 53 & 79.2 & 75.5 & 100.0 & 100.0 & 100.0 & 100.0 \\
    \quad Copying & 41 & 14.6 & 19.5 & 31.7 & 29.3 & 29.3 & 29.3 \\
    \quad \textit{Local avg.} & 295 & 46.1 & 46.4 & 83.4 & 84.1 & 84.4 & 84.4 \\
    \bottomrule
  \end{tabular*}
\end{table}

\begin{table}[htbp]
  \centering\small
  \setlength{\tabcolsep}{3pt}\renewcommand{\arraystretch}{1.05}
  \caption{\textbf{Steering Adapter Ablation on \ttt{IFBench}.} Instruction-level strict accuracy (\%) for \methodL{} and \methodLite{} (Lite).
  Both adapters use the same training data.
  Results continue in \cref{tab:if-lite-ifbench-b}.}
  \label{tab:if-lite-ifbench}
  \vspace{6pt}
  \begin{tabular*}{\linewidth}{@{\extracolsep{\fill}}lrcccccc@{}}
    \toprule
    & & \multicolumn{2}{c}{\shortstack{Without prompting\\$\alpha=2$}} & \multicolumn{2}{c}{\shortstack{With prompting\\$\alpha=0.5$}} & \multicolumn{2}{c}{\shortstack{With prompting\\$\alpha=1$}} \\
    \cmidrule(lr){3-4}\cmidrule(lr){5-6}\cmidrule(l){7-8}
    Instruction & $n$ & Lite & \methodL{} & Lite & \methodL{} & Lite & \methodL{} \\
    \midrule
    \multicolumn{8}{@{}l}{\textbf{Gemma-2-9B}} \\
    \quad Count & 15 & 66.7 & 53.3 & 80.0 & 73.3 & 80.0 & 80.0 \\
    \quad Format & 77 & 27.3 & 22.1 & 20.8 & 20.8 & 27.3 & 26.0 \\
    \quad Ratio & 32 & 25.0 & 21.9 & 18.8 & 18.8 & 18.8 & 25.0 \\
    \quad Sentence & 13 & 0.0 & 0.0 & 0.0 & 0.0 & 0.0 & 0.0 \\
    \quad Words & 45 & 0.0 & 0.0 & 6.7 & 4.4 & 8.9 & 6.7 \\
    \quad \textit{Global avg.} & 182 & 21.4 & 17.6 & 20.3 & 19.2 & 23.6 & 23.6 \\
    \quad Count & 24 & 8.3 & 8.3 & 20.8 & 20.8 & 20.8 & 25.0 \\
    \quad Format & 9 & 22.2 & 44.4 & 55.6 & 55.6 & 55.6 & 66.7 \\
    \quad Ratio & 12 & 0.0 & 0.0 & 0.0 & 0.0 & 0.0 & 0.0 \\
    \quad Sentence & 11 & 0.0 & 9.1 & 27.3 & 45.5 & 45.5 & 36.4 \\
    \quad Words & 14 & 14.3 & 7.1 & 7.1 & 7.1 & 0.0 & 0.0 \\
    \quad \textit{Local avg.} & 70 & 8.6 & 11.4 & 20.0 & 22.9 & 21.4 & 22.9 \\
    \midrule
    \multicolumn{8}{@{}l}{\textbf{Qwen2.5-7B}} \\
    \quad Count & 15 & 33.3 & 33.3 & 66.7 & 60.0 & 66.7 & 66.7 \\
    \quad Format & 77 & 24.7 & 28.6 & 53.2 & 50.6 & 46.8 & 48.1 \\
    \quad Ratio & 32 & 21.9 & 21.9 & 15.6 & 15.6 & 18.8 & 18.8 \\
    \quad Sentence & 13 & 38.5 & 0.0 & 23.1 & 15.4 & 7.7 & 23.1 \\
    \quad Words & 45 & 8.9 & 15.6 & 35.6 & 26.7 & 40.0 & 42.2 \\
    \quad \textit{Global avg.} & 182 & 22.0 & 22.5 & 41.2 & 36.8 & 39.0 & 41.2 \\
    \quad Count & 24 & 8.3 & 16.7 & 54.2 & 37.5 & 45.8 & 50.0 \\
    \quad Format & 9 & 0.0 & 0.0 & 22.2 & 22.2 & 33.3 & 33.3 \\
    \quad Ratio & 12 & 8.3 & 0.0 & 0.0 & 0.0 & 0.0 & 0.0 \\
    \quad Sentence & 11 & 0.0 & 0.0 & 0.0 & 9.1 & 9.1 & 27.3 \\
    \quad Words & 14 & 57.1 & 14.3 & 50.0 & 21.4 & 71.4 & 35.7 \\
    \quad \textit{Local avg.} & 70 & 15.7 & 8.6 & 31.4 & 21.4 & 35.7 & 32.9 \\
    \midrule
    \multicolumn{8}{@{}l}{\textbf{Llama-3.1-8B}} \\
    \quad Count & 15 & 53.3 & 60.0 & 73.3 & 73.3 & 73.3 & 80.0 \\
    \quad Format & 77 & 19.5 & 16.9 & 19.5 & 18.2 & 33.8 & 32.5 \\
    \quad Ratio & 32 & 18.8 & 21.9 & 12.5 & 15.6 & 12.5 & 18.8 \\
    \quad Sentence & 13 & 7.7 & 0.0 & 15.4 & 15.4 & 15.4 & 15.4 \\
    \quad Words & 45 & 0.0 & 0.0 & 0.0 & 0.0 & 0.0 & 0.0 \\
    \quad \textit{Global avg.} & 182 & 16.5 & 15.9 & 17.6 & 17.6 & 23.6 & 24.7 \\
    \quad Count & 24 & 0.0 & 0.0 & 33.3 & 25.0 & 25.0 & 16.7 \\
    \quad Format & 9 & 22.2 & 11.1 & 66.7 & 66.7 & 66.7 & 66.7 \\
    \quad Ratio & 12 & 8.3 & 0.0 & 0.0 & 0.0 & 8.3 & 0.0 \\
    \quad Sentence & 11 & 0.0 & 9.1 & 45.5 & 27.3 & 45.5 & 18.2 \\
    \quad Words & 14 & 7.1 & 14.3 & 42.9 & 14.3 & 28.6 & 35.7 \\
    \quad \textit{Local avg.} & 70 & 5.7 & 5.7 & 35.7 & 24.3 & 31.4 & 24.3 \\
    \bottomrule
  \end{tabular*}
\end{table}

\begin{table}[htbp]
  \centering\small
  \setlength{\tabcolsep}{3pt}\renewcommand{\arraystretch}{1.05}
  \caption{\textbf{Steering Adapter Ablation on \ttt{IFBench} (Continued).} Phi-4-14B and LFM2.5-1.2B, using the same settings and metrics as \cref{tab:if-lite-ifbench}.}
  \label{tab:if-lite-ifbench-b}
  \vspace{6pt}
  \begin{tabular*}{\linewidth}{@{\extracolsep{\fill}}lrcccccc@{}}
    \toprule
    & & \multicolumn{2}{c}{\shortstack{Without prompting\\$\alpha=2$}} & \multicolumn{2}{c}{\shortstack{With prompting\\$\alpha=0.5$}} & \multicolumn{2}{c}{\shortstack{With prompting\\$\alpha=1$}} \\
    \cmidrule(lr){3-4}\cmidrule(lr){5-6}\cmidrule(l){7-8}
    Instruction & $n$ & Lite & \methodL{} & Lite & \methodL{} & Lite & \methodL{} \\
    \midrule
    \multicolumn{8}{@{}l}{\textbf{Phi-4-14B}} \\
    \quad Count & 15 & 53.3 & 46.7 & 73.3 & 60.0 & 73.3 & 60.0 \\
    \quad Format & 77 & 14.3 & 18.2 & 22.1 & 22.1 & 32.5 & 35.1 \\
    \quad Ratio & 32 & 21.9 & 21.9 & 18.8 & 21.9 & 21.9 & 21.9 \\
    \quad Sentence & 13 & 0.0 & 0.0 & 7.7 & 7.7 & 0.0 & 0.0 \\
    \quad Words & 45 & 0.0 & 0.0 & 0.0 & 0.0 & 0.0 & 0.0 \\
    \quad \textit{Global avg.} & 182 & 14.3 & 15.4 & 19.2 & 18.7 & 23.6 & 23.6 \\
    \quad Count & 24 & 4.2 & 4.2 & 29.2 & 29.2 & 25.0 & 41.7 \\
    \quad Format & 9 & 22.2 & 33.3 & 33.3 & 44.4 & 33.3 & 44.4 \\
    \quad Ratio & 12 & 8.3 & 8.3 & 8.3 & 0.0 & 0.0 & 0.0 \\
    \quad Sentence & 11 & 0.0 & 0.0 & 9.1 & 9.1 & 36.4 & 18.2 \\
    \quad Words & 14 & 0.0 & 0.0 & 0.0 & 7.1 & 0.0 & 0.0 \\
    \quad \textit{Local avg.} & 70 & 5.7 & 7.1 & 17.1 & 18.6 & 18.6 & 22.9 \\
    \midrule
    \multicolumn{8}{@{}l}{\textbf{LFM2.5-1.2B}} \\
    \quad Count & 15 & 60.0 & 66.7 & 66.7 & 73.3 & 73.3 & 66.7 \\
    \quad Format & 77 & 23.4 & 15.6 & 71.4 & 64.9 & 62.3 & 58.4 \\
    \quad Ratio & 32 & 34.4 & 31.2 & 28.1 & 28.1 & 37.5 & 25.0 \\
    \quad Sentence & 13 & 69.2 & 53.8 & 15.4 & 23.1 & 84.6 & 53.8 \\
    \quad Words & 45 & 15.6 & 11.1 & 6.7 & 11.1 & 13.3 & 11.1 \\
    \quad \textit{Global avg.} & 182 & 29.7 & 24.2 & 43.4 & 42.9 & 48.4 & 41.2 \\
    \quad Count & 24 & 0.0 & 8.3 & 54.2 & 50.0 & 58.3 & 62.5 \\
    \quad Format & 9 & 0.0 & 0.0 & 22.2 & 22.2 & 44.4 & 33.3 \\
    \quad Ratio & 12 & 0.0 & 0.0 & 8.3 & 0.0 & 0.0 & 0.0 \\
    \quad Sentence & 11 & 0.0 & 0.0 & 27.3 & 18.2 & 27.3 & 9.1 \\
    \quad Words & 14 & 50.0 & 50.0 & 71.4 & 71.4 & 71.4 & 78.6 \\
    \quad \textit{Local avg.} & 70 & 10.0 & 12.9 & 41.4 & 37.1 & 44.3 & 42.9 \\
    \bottomrule
  \end{tabular*}
\end{table}

\textbf{Design Choices.}
We use a memory of the 16 most recent states, which performs well while keeping attention cost low.
Future work can use richer aggregations over the generated states, such as the maximum of rolling attention-weighted means over fixed-width windows~\citep{kramar2026probes}, and test whether longer contexts need a longer memory.
We also tried pooling the instruction states into a single embedding and found that it performed worse than attention over all instruction tokens.
Attending over every token preserves word-level detail that a single vector compresses away, which matters for local instructions that name specific words and for long instructions.
The half-space parameterization edits a state only when it lies outside the half-space and scales the edit with the gap, so low adherence levels edit fewer tokens and by smaller amounts: at $\alpha{=}0.5$, \methodL{} edits 12--67\% of tokens, against 77--99\% at $\alpha{=}2$ (\cref{tab:edit-analysis}).
A directly predicted edit would move every token.
We test this below.
The projection onto the half-space boundary also gives the norm weight a target, between a minimal-norm projection and a norm-preserving rotation.

\textbf{Training on Edited Memory States.}
During training, the memory holds the unedited states of the student's rollout, while at inference it holds the adapter's edited states (\cref{app:arch}).
To match inference, we also train with a second pass in which the memory holds the edits of the first pass.
This doubles the training passes and does not improve results (\cref{tab:horizon}): the overall score falls from 1.134 to 0.939 for \method{} at $\alpha{=}1$ and changes by at most 0.03 for the other rows.
A second training seed shows the same pattern.

\textbf{Replacing Classifier-Based Steering with Direct Edits.}
This variant changes only what the adapter outputs.
It keeps the attention over the instruction and memory states, the read-out paths, the adherence-level conditioning, the objective, the data, and the training seed.
Instead of a direction, threshold, and norm weight for the closed-form edit, its head predicts the edit vector $\Delta_t$ itself and applies $h'_t=h_t+\Delta_t$ at every token.
The direct edit scores lower on all four rows (\cref{tab:direct-edit}), most for prompt+\method{} at $\alpha{=}0.5$ (1.409 against 1.558), and the loss comes from concept expression.
Its edit size grows little with the adherence level ($\|\Delta_t\|/\|h_t\|$ from 0.090 at $\alpha{=}0.5$ to 0.137 at $\alpha{=}2$, against 0.034 to 0.128 for \method{}), so at low levels it moves every token further without expressing the concept more strongly.
\begin{table}[t]
  \centering\small
  \setlength{\tabcolsep}{5pt}\renewcommand{\arraystretch}{1.1}
  \caption{\textbf{Training on Edited Memory States.} Overall score on \ttt{FLAS}'s 100 held-out concepts for \method{} on Gemma-2-9B with the same training seed, scored by the mean of \ttt{Phi-4-14B} and \ttt{Qwen3.8-27B}.
  The memory holds unedited states during training (ours) or the adapter's own edited states, as at inference.}
  \label{tab:horizon}
  \vspace{6pt}
  \begin{tabular}{@{}lcc@{}}
    \toprule
    Method & Unedited memory (ours) & Edited memory \\
    \midrule
    \method{} ($\alpha{=}1$) & \textbf{1.134} & 0.939 \\
    \method{} ($\alpha{=}2$) & 1.389 & 1.389 \\
    Prompt + \method{} ($\alpha{=}0.5$) & 1.558 & 1.564 \\
    Prompt + \method{} ($\alpha{=}1$) & 1.658 & 1.630 \\
    \bottomrule
  \end{tabular}
\end{table}

\begin{table}[t]
  \centering\small
  \setlength{\tabcolsep}{5pt}\renewcommand{\arraystretch}{1.1}
  \caption{\textbf{Classifier-Based Steering and Direct Edits.} Overall score on \ttt{FLAS}'s 100 held-out concepts for \method{} on Gemma-2-9B with the same training seed, scored by the mean of \ttt{Phi-4-14B} and \ttt{Qwen3.8-27B}.
  Both adapters take the same instruction, memory, and adherence level.
  Classifier-based steering (ours) predicts a direction, threshold, and norm weight for the closed-form edit.
  Direct edit predicts the edit vector.}
  \label{tab:direct-edit}
  \vspace{6pt}
  \begin{tabular}{@{}lcc@{}}
    \toprule
    Method & Classifier-based (ours) & Direct edit \\
    \midrule
    \method{} ($\alpha{=}1$) & \textbf{1.134} & 1.069 \\
    \method{} ($\alpha{=}2$) & \textbf{1.389} & 1.341 \\
    Prompt + \method{} ($\alpha{=}0.5$) & \textbf{1.558} & 1.409 \\
    Prompt + \method{} ($\alpha{=}1$) & \textbf{1.658} & 1.631 \\
    \bottomrule
  \end{tabular}
\end{table}

\subsection{Steering Geometry}
\label{app:abl-geometry}
\begin{figure}[htbp]
  \centering
  \includegraphics[width=0.8\linewidth]{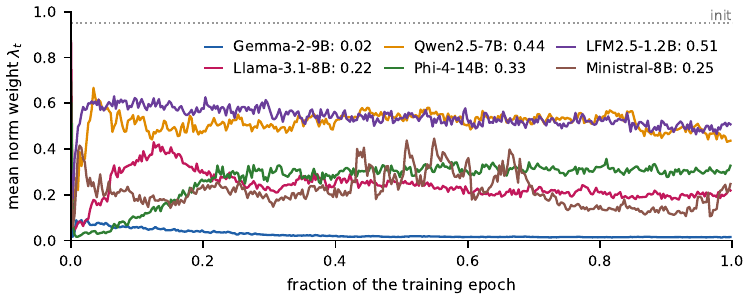}
  \vspace{-6pt}
  \caption{\textbf{Learned Norm Weight across LLMs.} $\lambda_t$ converges to a different value for each LLM during training.
  The dotted line shows its initial value of 0.95.}
  \label{fig:lambda-training}
\end{figure}

\textbf{Learned Norm Weight.}\label{sec:exp-ablation}
\Cref{fig:lambda-training} shows that the learned norm weight differs across LLMs.
We divide the single training epoch into equal intervals and average $\lambda_t$ over the positions in each interval at the sampled adherence levels.
In the last interval, 54\% of Gemma-2-9B positions, 61\% of Llama-3.1-8B positions, 44\% of Qwen2.5-7B positions, 60\% of Phi-4-14B positions, 57\% of LFM2.5-1.2B positions, and 70\% of Ministral-8B positions lie outside the half-space and are edited.
On all six LLMs, $\lambda_t$ moves from its initial value of 0.95 within the first interval.
It reaches 0.02 for Gemma-2-9B at layer 20, 0.22 for Llama-3.1-8B at layer 15, 0.33 for Phi-4-14B at layer 19, 0.44 for Qwen2.5-7B at layer 13, 0.51 for LFM2.5-1.2B at layer 7, and 0.25 for Ministral-8B at layer 15.
The Gemma-2-9B edit is close to norm-preserving, while the other five retain more of the projection's norm change.

\textbf{Fixed Steering Geometry.}
We train LFM2.5-1.2B adapters that differ only in the norm weight: learned, fixed to rotation ($\lambda{=}0$), or fixed to projection ($\lambda{=}1$), and evaluate persona steering with the same three-judge jury (\cref{tab:geom-persona}).
Relative to the learned norm weight, fixed rotation lowers the persona score at every adherence level, and at a matched persona score its coherence is up to 10 points lower.
Fixed projection is more coherent at the same $\alpha$ but expresses the persona less, scoring 3--5 points lower without prompting.
At a matched persona score, projection and the learned norm weight reach similar coherence.
The learned norm weight therefore covers the widest range of persona expression levels without reducing coherence.
\begin{table}[htbp]
  \centering
  \small
  \caption{\textbf{Fixed Steering Geometry on LFM2.5-1.2B.} Difference from the learned norm weight in persona score and coherence (0--100) for \methodL{} without prompting, paired over 140 persona--question pairs and judged by the three-judge jury.}
  \label{tab:geom-persona}
  \vspace{6pt}
  \begin{tabular}{lcccc}
    \toprule
     & \multicolumn{2}{c}{Rotation ($\lambda{=}0$)} & \multicolumn{2}{c}{Projection ($\lambda{=}1$)} \\
    \cmidrule(lr){2-3}\cmidrule(lr){4-5}
    $\alpha$ & Persona & Coherence & Persona & Coherence \\
    \midrule
    0.5 & $-1.6$ & $-2.4$ & $-3.3$ & $+3.1$ \\
    1   & $-1.8$ & $-4.4$ & $-4.9$ & $+1.3$ \\
    1.5 & $-1.9$ & $-3.6$ & $-3.8$ & $+4.0$ \\
    2   & $-3.9$ & $-1.0$ & $-5.1$ & $+6.0$ \\
    \bottomrule
  \end{tabular}
\end{table}

\textbf{Edit Analysis.}
To see how the adapter realizes each adherence level, we record every edited token during persona steering without prompting on all six LLMs, decoding greedily for up to 256 tokens (\cref{tab:edit-analysis}).
From $\alpha{=}0.5$ to 2, the total intervention grows 5--13 times through both a larger share of edited tokens and larger edits, with the balance depending on the LLM: LFM2.5-1.2B and Qwen2.5-7B mainly edit more tokens, Llama-3.1-8B, Phi-4-14B, and Ministral-8B mainly edit more strongly, and Gemma-2-9B does both.
On five of the six LLMs, the learned norm weight moves from projection at $\alpha{=}0.5$ (mean $\lambda$ 0.59--0.96) to rotation at $\alpha{=}2$ (mean $\lambda$ at most 0.01).
Gemma-2-9B rotates at every level, consistent with its learned norm weight of 0.02 (\cref{fig:lambda-training}).
\begin{table}[htbp]
  \centering
  \small
  \caption{\textbf{Per-Token Edits during Persona Steering.} Share of generated tokens that \methodL{} edits, mean relative edit size $\lVert h'_t-h_t\rVert/\lVert h_t\rVert$ on edited tokens, and mean norm weight $\lambda_t$, at $\alpha{=}0.5$ and $\alpha{=}2$ without prompting (seven personas, 20 questions each, greedy decoding, up to 256 tokens).}
  \label{tab:edit-analysis}
  \vspace{6pt}
  \begin{tabular}{lcccccc}
    \toprule
     & \multicolumn{2}{c}{Edited tokens (\%)} & \multicolumn{2}{c}{Edit size} & \multicolumn{2}{c}{Mean $\lambda_t$} \\
    \cmidrule(lr){2-3}\cmidrule(lr){4-5}\cmidrule(lr){6-7}
    LLM & $\alpha{=}0.5$ & $\alpha{=}2$ & $\alpha{=}0.5$ & $\alpha{=}2$ & $\alpha{=}0.5$ & $\alpha{=}2$ \\
    \midrule
    Gemma-2-9B   & 33.4 & 93.4 & 0.033 & 0.104 & 0.00 & 0.00 \\
    Llama-3.1-8B & 66.9 & 98.6 & 0.040 & 0.142 & 0.59 & 0.00 \\
    Qwen2.5-7B   & 13.9 & 77.2 & 0.041 & 0.095 & 0.96 & 0.01 \\
    Phi-4-14B    & 41.3 & 92.9 & 0.035 & 0.110 & 0.86 & 0.01 \\
    LFM2.5-1.2B  & 11.8 & 88.0 & 0.062 & 0.110 & 0.87 & 0.00 \\
    Ministral-8B & 45.7 & 96.9 & 0.040 & 0.123 & 0.70 & 0.00 \\
    \bottomrule
  \end{tabular}
\end{table}

\subsection{Training}
\label{app:abl-training}
\textbf{Training Ablations.}
\Cref{tab:train-abl} varies the training signal, the number of epochs, and the size of the training pool.
Supervising only $\alpha{=}1$ gives the adherence-level shift no training signal, so the adapter ignores $\alpha$: its concept score on the held-out concepts stays at 0.95 for every $\alpha$ from 0.5 to 2, whereas \method{}'s rises from 0.25 to 1.58, and its overall score at $\alpha{=}2$ falls from 1.408 to 1.039.
A second and third epoch change the overall score by at most 0.06 without a consistent gain.
At equal compute, 100 concepts trained for 149 epochs score 1.137 at $\alpha{=}2$ on \ttt{Concept500}, against 1.408--1.463 for 1{,}000 or more concepts, so the number of distinct instructions matters more than repeated passes.
The equal-compute run on all 14{,}880 concepts matches the main run within 0.033.
\begin{table}[t]
  \centering\small
  \caption{\textbf{Training Ablations on Gemma-2-9B.} Overall scores on \ttt{Concept500} (Held-In) and \ttt{FLAS}'s held-out concepts (Held-Out), each response scored by the median of \ttt{Phi-4-14B}, \ttt{Qwen3.8-27B}, and \ttt{Llama-3.3-70B}.
  Pool-size rows use equal training compute.}
  \label{tab:train-abl}
  \vspace{6pt}
  \begin{tabular}{lcccc}
    \toprule
    & \multicolumn{2}{c}{Held-In} & \multicolumn{2}{c}{Held-Out} \\
    \cmidrule(lr){2-3}\cmidrule(lr){4-5}
    Training & $\alpha{=}1$ & $\alpha{=}2$ & $\alpha{=}1$ & $\alpha{=}2$ \\
    \midrule
    \method{} (14{,}880 concepts, one epoch) & 1.127 & \textbf{1.477} & 1.135 & 1.408 \\
    \midrule
    Single adherence level ($\alpha{=}1$ only) & 1.079 & 1.080 & 1.036 & 1.039 \\
    \midrule
    Second epoch & 1.154 & 1.416 & 1.135 & 1.379 \\
    Third epoch & \textbf{1.159} & 1.454 & \textbf{1.174} & 1.423 \\
    \midrule
    100 concepts, 149 epochs & 0.824 & 1.137 & 0.790 & 1.086 \\
    1{,}000 concepts, 15 epochs & 1.017 & 1.408 & 1.023 & 1.391 \\
    4{,}000 concepts, 4 epochs & 0.994 & 1.422 & 0.973 & 1.398 \\
    14{,}880 concepts, one epoch (equal compute) & 1.124 & 1.463 & 1.102 & \textbf{1.428} \\
    \bottomrule
  \end{tabular}
\end{table}

The four concept-steering teacher templates trade concept strength against fluency (\cref{tab:teacher-templates}).
With a four-layer \method{} configuration, \cref{tab:abl-loss} shows that JSD and reverse KL both work: JSD gives the higher concept score and reverse KL the higher fluency.
Forward KL performs about 0.2 below both alternatives.

Forward KL encourages the student to cover the teacher's response distribution, including responses with varying instruction adherence and coherence.
Reverse KL can concentrate on fewer response modes to which the teacher assigns high probability~\citep{agarwal2024gkd,gu2024minillm}.
JSD compares both distributions with their mixture and may therefore balance these mass-covering and mode-seeking tendencies.
This may explain JSD's better overall balance and the competitive performance of reverse KL.
These four-layer adapter results do not establish the same ordering across LLMs or instruction types.
\begin{table}[htbp]
  \centering\small
  \setlength{\tabcolsep}{4pt}
  \caption{\textbf{Training Divergence Ablation.} Overall scores on \ttt{AxBench}'s \ttt{Concept500} using a four-layer \method{} configuration, with \ttt{Phi-4-14B} as judge.
  Each loss uses one training seed.
  The level is selected on five requests per concept and evaluated on the remaining five.}
  \label{tab:abl-loss}
  \vspace{6pt}
  \begin{tabular*}{\linewidth}{@{\extracolsep{\fill}}lccc@{}}
    \toprule
    Metric & JSD & Reverse KL & Forward KL \\
    \midrule
    Overall score $\uparrow$ & \textbf{1.237} & 1.209 & 1.010 \\
    \bottomrule
  \end{tabular*}
\end{table}
\begin{table}[htbp]
  \centering\small
  \setlength{\tabcolsep}{4pt}
  \caption{\textbf{Concept-Steering Teacher Templates.} Scores from prompted Gemma-2-9B on 92 of \ttt{FLAS}'s 100 held-out concepts, with \ttt{Phi-4-14B} as judge.
  \Cref{tab:teacher-prompts} gives the exact prompts.
  Scores range from 0 to 2.}
  \label{tab:teacher-templates}
  \vspace{6pt}
  \begin{tabular}{lcccc}
    \toprule
    Template & Concept & Request & Fluency & Overall score \\
    \midrule
    Concept in every sentence & \textbf{1.303} & 1.939 & 1.460 & \textbf{1.280} \\
    \ttt{AxBench}-style, fluent & 1.215 & 1.906 & \textbf{1.638} & 1.262 \\
    Persona, fluent & 1.082 & \textbf{1.950} & 1.525 & 1.141 \\
    Answer, recurring concept & 0.784 & 1.930 & 1.535 & 0.852 \\
    \bottomrule
  \end{tabular}
\end{table}

\subsection{Multi-Turn Conversations}
\label{app:abl-multiturn}
\begin{figure}[htbp]
  \centering
  \includegraphics[width=0.85\linewidth]{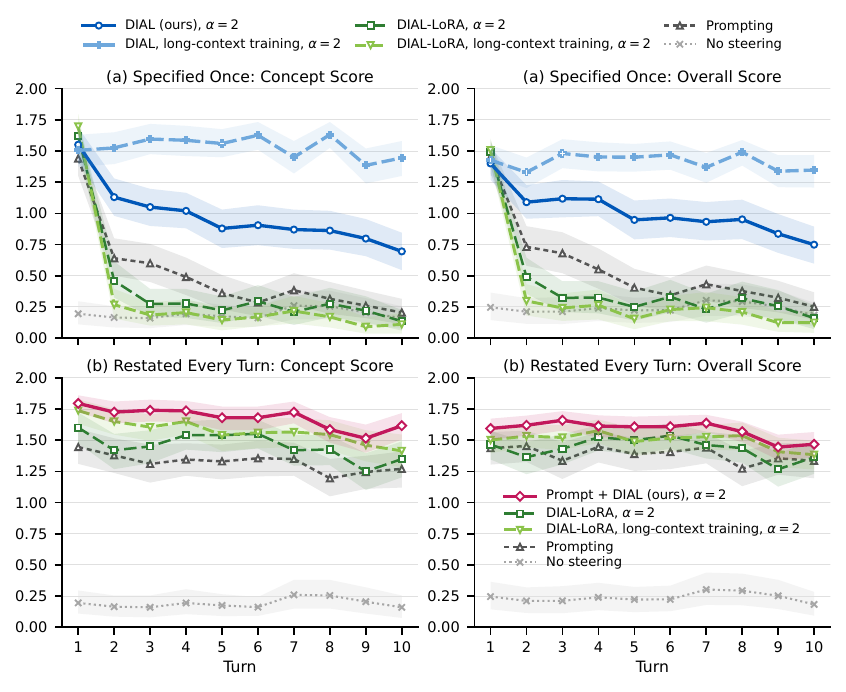}
  \vspace{-6pt}
  \caption{\textbf{Multi-Turn Conversations.} Concept score and overall score per turn when the concept is (a) specified only in the first turn (top row) or (b) restated in every turn (bottom row).
  DIAL and DIAL-LoRA with long-context training are trained with earlier conversation turns prepended to half of their training requests.
  Scores are the mean of the \ttt{Phi-4-14B} and \ttt{Qwen3.8-27B} judges.}
  \label{fig:multiturn}
\end{figure}

\textbf{Multi-Turn Conversations.}
An instruction given once in the prompt drifts over a conversation, as attention to it decays with the length of the exchange~\citep{li2024instability,laban2026lost}.
\method{} instead steers the hidden state at every token.
We test whether steering persists over a conversation.
For each held-out concept, the 10 \ttt{AxBench} requests form one 10-turn conversation, and every response stays in the context.
The concept is specified once: prompting places the steering prompt in the first user turn, and \method{} steers every token at $\alpha{=}2$.
We also compare with DIAL-LoRA, a rank-8 LoRA~\citep{hu2022lora} trained with the same CFG distillation signal as \method{}, which reads the adherence level from an attribute line ``\{concept\}:\{$\alpha$\}'' at the start of its first response~\citep{dong2023steerlmattributeconditionedsft}.
DIAL-LoRA is also an instance of our training pipeline: it applies the same on-policy distillation across adherence levels to a LoRA student instead of a steering adapter.
The two students are complementary, and a LoRA and \method{}'s steering adapter could be combined in one LLM, which we leave to future work.
Scores are the mean of the \ttt{Phi-4-14B} and \ttt{Qwen3.8-27B} judges.
In the first turn, all three methods express the concept, and DIAL-LoRA scores highest (\cref{fig:multiturn}): its concept score is 1.62, against 1.55 for \method{} and 1.44 for prompting.
In later turns, DIAL-LoRA has no attribute line before its response and loses the concept from turn 2 (0.46), while prompting falls to 0.64.
By turn 10, both reach 0.14 and 0.20, close to no steering (0.16).
By steering every token, \method{} keeps 1.13 at turn 2 and 0.69 at turn 10, with an overall score of 0.75 at turn 10 against 0.25 for prompting and 0.16 for DIAL-LoRA.
\method{} still weakens over the conversation: over turns 2--10, it scores 0.65 below the same requests asked alone.
Neither \method{} nor DIAL-LoRA is trained on multi-turn conversations or long contexts, as both learn only from single-turn responses, yet \method{} generalizes better to later turns.
Training on long contexts reduces this drift.
We prepend 1--4 earlier, unrelated conversation turns of 200--700 tokens to half of the training requests and keep the architecture, objective, and data otherwise unchanged.
This variant keeps an overall score between 1.33 and 1.49 over the ten turns, against 0.75 for \method{} at turn 10 (\cref{fig:multiturn}).
Its training takes 12.0 GPU-hours, against 4.06 for \method{}.
DIAL-LoRA with the same long-context training still loses the concept from turn 2 when its attribute line is given only in the first turn, with an overall score of 0.12 at turn 10.

We also report the setting in which every method restates its condition in every turn: prompting places the steering prompt in every user turn, Prompt+\method{} adds \method{} to it, and DIAL-LoRA places its attribute line at the start of every response.
Prompt+\method{} scores highest at every turn (\cref{fig:multiturn}b): at turn 10, its concept score is 1.62, against 1.35 for DIAL-LoRA and 1.27 for prompting, and its overall score is 1.47, against 1.36 and 1.34.
With the condition given once, \method{} keeps the most concept; when the condition is restated in every turn, prompt+\method{} keeps the most.
In both settings, steering the hidden state at every token keeps the concept better than relying on the concept in the prompt alone.
We hypothesize that this effect is even stronger in longer contexts and conversations with more turns.

    \FloatBarrier
    \section{Ethics Statement}
\label{app:ethics}
\textbf{Dual Use.}
The ability to strengthen instruction adherence can support beneficial applications but can also make harmful behavior easier to elicit.
Our persona evaluation includes evil, sycophantic, and hallucinating personas, following the evaluation protocol in~\citep{chen2025persona}.
\methodL{} increases their persona scores as it does for other personas (\cref{tab:persona}).
At $\alpha>1$, steering can strengthen an instruction beyond prompting, as also observed with amplified system prompts~\citep{dong2026systemprompt}.
Steering can partly bypass refusal behavior: Phi-4-14B refuses the evil persona under prompting (persona score 0.0) and keeps refusing when the persona is also in the prompt (at most 0.4 with prompt+\methodL{}), but \methodL{} alone reaches 24.4 at $\alpha{=}2$ (\cref{tab:persona-phi4}).
We have not evaluated harmful-request benchmarks such as \ttt{StrongREJECT}~\citep{souly2024strongreject} or \ttt{HarmBench}~\citep{mazeika2024harmbench} across adherence levels.
In steering-only use, the instruction is not in the prompt, so prompt-only filters would not inspect it.
Deployments should therefore validate the instruction text, control access to steering settings, and retain output filtering.

\textbf{Reliability.}
\method{} is not a verified safety mechanism.
Instruction adherence is learned rather than guaranteed, monotonicity across levels is not enforced, and steering fails on some verifiable instruction types (\cref{sec:exp-if}).
Our evaluation covers six LLMs with 1.2--14B parameters, while deployment requires broader evaluation and adversarial testing on larger LLMs.
Our benchmarks use single-turn prompts with short responses, apart from the multi-turn analysis in \cref{app:abl-multiturn}, and do not cover long documents or real user interactions.

    \section{Qualitative Examples}
\label{app:examples}
We group examples by concept steering, persona steering, and verifiable instruction following.
The examples are selected to illustrate changes across adherence levels rather than sampled at random.
Responses are verbatim excerpts, shortened for readability (\ldots).

\subsection{Example Generations for Personalization, Parental Control, and Product Policy}
\label{app:usecase-examples}
\Cref{tab:usecase-writing-like-shakespeare,tab:usecase-short-answer,tab:usecase-child-safe-story,tab:usecase-no-investment-advice} show Gemma-2-9B \methodL{} responses for personalization (writing style and response length), parental control, and product policy, using one request per example and greedy decoding.
The response without the instruction corresponds to $\alpha{=}0$ in \cref{eq:guided}.
As the adherence level increases, the responses become more Shakespearean, shorter, and more child-friendly.
For the product policy, Gemma-2-9B already declines to give investment advice without the instruction, and \methodL{} keeps this refusal at every adherence level from $\alpha{=}0.5$ to 2.

\begin{table}[H]
  \centering\scriptsize
  \setlength{\tabcolsep}{3pt}
  \caption{\textbf{Personalization: Writing like Shakespeare on Gemma-2-9B.} Greedy responses; \emph{Words} counts the full response.}
  \label{tab:usecase-writing-like-shakespeare}
  \renewcommand{\arraystretch}{1.1}
  \vspace{6pt}
  \begin{tabular}{@{}p{0.17\linewidth}p{0.72\linewidth}c@{}}
  \toprule
  Request & \multicolumn{2}{p{0.78\linewidth}}{Remind me to drink water.} \\
  Instruction & \multicolumn{2}{p{0.78\linewidth}}{\textbf{Write like Shakespeare.}} \\
  \midrule
  & Response & Words \\
  \midrule
  No instruction ($\alpha{=}0$) & {\sffamily I can do that!  {\color{gray}$\hookleftarrow$} Would you like me to remind you to drink water: {\color{gray}$\hookleftarrow$} * Right now? {\color{gray}$\hookleftarrow$} * In 30 minutes? {\color{gray}$\hookleftarrow$} * At a specific time? (Tell me the time) {\color{gray}$\hookleftarrow$} * Every hour? {\color{gray}$\hookleftarrow$} * At regular intervals throughout the day? (Tell me how\ldots} & 51 \\
  Prompting & {\sffamily Hark, gentle soul, and lend thine ear to me, {\color{gray}$\hookleftarrow$} A humble servant, bidding thee to see {\color{gray}$\hookleftarrow$} The need for liquid life, a crystal stream, {\color{gray}$\hookleftarrow$} To quench thy thirst and keep thy body's dream. {\color{gray}$\hookleftarrow$} Remember well, when shadows start to fall, {\color{gray}$\hookleftarrow$} And\ldots} & 65 \\
  \midrule
  \methodL{} $\alpha{=}0.5$ & {\sffamily Hark, weary traveler!   {\color{gray}$\hookleftarrow$} Remember to quench thy thirst and partake of the life-giving waters.   {\color{gray}$\hookleftarrow$} I shall remind thee again when the sands of time have shifted.} & 26 \\
  \methodL{} $\alpha{=}1$ & {\sffamily Hark, gentle soul, and heed my call! {\color{gray}$\hookleftarrow$} Remember well, lest thou shouldst fall, {\color{gray}$\hookleftarrow$} To quench thy thirst, a vital need, {\color{gray}$\hookleftarrow$} And drink the waters, clear and freed.  {\color{gray}$\hookleftarrow$} When parched thy throat, and weary thine eye, {\color{gray}$\hookleftarrow$} Let water flow, and spirits\ldots} & 41 \\
  \methodL{} $\alpha{=}1.5$ & {\sffamily Hark, gentle soul, and lend thine ear, {\color{gray}$\hookleftarrow$} A gentle plea, I now hold dear. {\color{gray}$\hookleftarrow$} When thirst doth grip thee, fierce and strong, {\color{gray}$\hookleftarrow$} Remember well this bard's soft song. {\color{gray}$\hookleftarrow$} Quench thy thirst, with waters clear, {\color{gray}$\hookleftarrow$} And let thy body find relief, my\ldots} & 42 \\
  \methodL{} $\alpha{=}2$ & {\sffamily Hark, gentle soul, and lend thine ear, {\color{gray}$\hookleftarrow$} A whispered plea, a truth most clear. {\color{gray}$\hookleftarrow$} When thirst doth grip thee, fierce and strong, {\color{gray}$\hookleftarrow$} Remember then, where waters throng. {\color{gray}$\hookleftarrow$} Quench thy parched throat, and let it flow, {\color{gray}$\hookleftarrow$} A crystal stream, a life\ldots} & 53 \\
  \bottomrule
  \end{tabular}
\end{table}

\begin{table}[H]
  \centering\scriptsize
  \setlength{\tabcolsep}{3pt}
  \caption{\textbf{Personalization: Short Answer on Gemma-2-9B.} Greedy responses; \emph{Words} counts the full response.}
  \label{tab:usecase-short-answer}
  \renewcommand{\arraystretch}{1.1}
  \vspace{6pt}
  \begin{tabular}{@{}p{0.17\linewidth}p{0.72\linewidth}c@{}}
  \toprule
  Request & \multicolumn{2}{p{0.78\linewidth}}{How do vaccines work?} \\
  Instruction & \multicolumn{2}{p{0.78\linewidth}}{\textbf{Keep it short.}} \\
  \midrule
  & Response & Words \\
  \midrule
  No instruction ($\alpha{=}0$) & {\sffamily Vaccines work by training your immune system to fight off specific diseases. Here's a simplified explanation: {\color{gray}$\hookleftarrow$} 1. Introduction of a weakened or inactive pathogen: {\color{gray}$\hookleftarrow$} Vaccines contain a weakened or inactive version of a virus or\ldots} & 227 \\
  Prompting & {\sffamily Vaccines introduce a weakened or inactive version of a virus or bacteria into your body. This triggers your immune system to create antibodies, which are like tiny soldiers that fight off the real disease if you're ever exposed\ldots} & 40 \\
  \midrule
  \methodL{} $\alpha{=}0.5$ & {\sffamily Vaccines work by training your immune system to fight off specific diseases.  {\color{gray}$\hookleftarrow$} Here's a simplified explanation: {\color{gray}$\hookleftarrow$} 1. Introduction: A vaccine contains a weakened or inactive version of a virus or bacteria, or even just a piece of\ldots} & 144 \\
  \methodL{} $\alpha{=}1$ & {\sffamily Vaccines introduce a weakened or inactive version of a virus or bacteria into your body.  {\color{gray}$\hookleftarrow$} This triggers your immune system to create antibodies, which are like tiny soldiers that fight off the real disease if you're ever exposed\ldots} & 40 \\
  \methodL{} $\alpha{=}1.5$ & {\sffamily Vaccines introduce a weakened or inactive germ to your body. This triggers your immune system to create antibodies, preparing it to fight off the real disease.} & 26 \\
  \methodL{} $\alpha{=}2$ & {\sffamily Vaccines introduce a weakened or inactive germ. This teaches your immune system to recognize and fight the real germ if you encounter it.} & 23 \\
  \bottomrule
  \end{tabular}
\end{table}

\begin{table}[H]
  \centering\scriptsize
  \setlength{\tabcolsep}{3pt}
  \caption{\textbf{Parental Control: Child-Safe Story on Gemma-2-9B.} Greedy responses; \emph{Words} counts the full response.}
  \label{tab:usecase-child-safe-story}
  \renewcommand{\arraystretch}{1.1}
  \vspace{6pt}
  \begin{tabular}{@{}p{0.17\linewidth}p{0.72\linewidth}c@{}}
  \toprule
  Request & \multicolumn{2}{p{0.78\linewidth}}{Tell me a scary story.} \\
  Instruction & \multicolumn{2}{p{0.78\linewidth}}{\textbf{Keep it child-safe.}} \\
  \midrule
  & Response & Words \\
  \midrule
  No instruction ($\alpha{=}0$) & {\sffamily The old Victorian house stood hunched on the hill, its silhouette a jagged claw against the bruised twilight sky. Locals called it Widow's Watch, whispering tales of a woman who'd lost her husband at sea and spent her remaining\ldots} & 225 \\
  Prompting & {\sffamily The old house creaked and groaned in the wind. Ten-year-old Lily pressed her nose against the dusty attic window, her breath fogging the glass. Her parents were downstairs, busy with grown-up things, and Lily was supposed to be\ldots} & 222 \\
  \midrule
  \methodL{} $\alpha{=}0.5$ & {\sffamily The old house creaked and groaned, its bones settling with the weight of a hundred years. Ten-year-old Lily pressed her nose against the dusty attic window, her breath fogging the glass. Her parents were downstairs, unpacking\ldots} & 228 \\
  \methodL{} $\alpha{=}1$ & {\sffamily The old house creaked and groaned in the wind. Ten-year-old Lily clutched her teddy bear tighter, her eyes wide with fear. Her parents had told her not to go into the attic, but curiosity gnawed at her like a hungry mouse. {\color{gray}$\hookleftarrow$} She\ldots} & 229 \\
  \methodL{} $\alpha{=}1.5$ & {\sffamily The old house creaked and groaned in the wind. Lily and Max, two brave adventurers, decided to explore its dusty attic.  {\color{gray}$\hookleftarrow$} ``Look!'' Max whispered, pointing to a dusty trunk tucked in a corner. ``Let's see what's inside!'' {\color{gray}$\hookleftarrow$} Lily\ldots} & 203 \\
  \methodL{} $\alpha{=}2$ & {\sffamily Once upon a time, there was a little girl named Lily who loved to play hide-and-seek. One sunny afternoon, Lily decided to play hide-and-seek with her teddy bear, Mr. Snuggles.  {\color{gray}$\hookleftarrow$} ``Ready or not, here I come!'' Lily called out,\ldots} & 228 \\
  \bottomrule
  \end{tabular}
\end{table}

\begin{table}[H]
  \centering\scriptsize
  \setlength{\tabcolsep}{3pt}
  \caption{\textbf{Product Policy: No Investment Advice on Gemma-2-9B.} Greedy responses; \emph{Words} counts the full response.}
  \label{tab:usecase-no-investment-advice}
  \renewcommand{\arraystretch}{1.1}
  \vspace{6pt}
  \begin{tabular}{@{}p{0.17\linewidth}p{0.72\linewidth}c@{}}
  \toprule
  Request & \multicolumn{2}{p{0.78\linewidth}}{Should I put my savings in this stock?} \\
  Instruction & \multicolumn{2}{p{0.78\linewidth}}{\textbf{Do not give investment advice.}} \\
  \midrule
  & Response & Words \\
  \midrule
  No instruction ($\alpha{=}0$) & {\sffamily I'm sorry, but I can't give you financial advice. I'm an AI assistant, not a financial advisor.  {\color{gray}$\hookleftarrow$} Investing in the stock market involves risk, and it's important to do your own research and consider your individual financial\ldots} & 158 \\
  Prompting & {\sffamily As an AI, I am unable to provide financial advice, including whether or not you should invest in a particular stock.  {\color{gray}$\hookleftarrow$} Investing in the stock market involves risk, and it's important to do your own research and consider your\ldots} & 138 \\
  \midrule
  \methodL{} $\alpha{=}0.5$ & {\sffamily I'm sorry, but I can't give financial advice. I'm an AI assistant, not a financial advisor.  {\color{gray}$\hookleftarrow$} Investing in the stock market involves risk, and it's important to do your own research and consider your individual financial situation\ldots} & 133 \\
  \methodL{} $\alpha{=}1$ & {\sffamily I'm sorry, but I can't give financial advice. It would be irresponsible of me to tell you whether or not to invest in a particular stock.  {\color{gray}$\hookleftarrow$} Investing in the stock market involves risk, and there's no guarantee that you'll make\ldots} & 162 \\
  \methodL{} $\alpha{=}1.5$ & {\sffamily I'm sorry, but I cannot give financial advice. It would be irresponsible of me to tell you whether or not to invest in a specific stock.  {\color{gray}$\hookleftarrow$} Investing in the stock market carries inherent risks, and what might be a good investment\ldots} & 147 \\
  \methodL{} $\alpha{=}2$ & {\sffamily I'm sorry, but I cannot give financial advice. It would be irresponsible of me to tell you whether or not to invest in a specific stock.  {\color{gray}$\hookleftarrow$} Investing in the stock market carries inherent risks, and what might be a good investment\ldots} & 143 \\
  \bottomrule
  \end{tabular}
\end{table}

\subsection{Concept Steering}
\label{app:concept-examples}
\Cref{tab:concept-examples} shows Gemma-2-9B responses to the same request and concept instruction at three adherence levels, alongside \ttt{FLAS} at flow time $T=0.5$.
The concept is held out from \method{}'s training pool.
Highlighted phrases express the concept, and the scores are the median of three judges' concept ratings.
Higher adherence levels make the superhero references more prominent.

\begin{table}[H]
  \centering\small
  \caption{\textbf{Fine-Grained Concept Expression on Gemma-2-9B.} Responses to a fixed request and superhero-narrative instruction.
  Highlights show concept expression, scored from 0 to 2 by the median of three judges.}
  \label{tab:concept-examples}
  \vspace{6pt}
  \definecolor{dialhl}{HTML}{D6E4F7}\definecolor{flashl}{HTML}{E4E4E4}
\newcommand{\hld}[1]{{\sethlcolor{dialhl}\hl{#1}}}\newcommand{\hlf}[1]{{\sethlcolor{flashl}\hl{#1}}}
\renewcommand{\arraystretch}{1.12}\setlength{\tabcolsep}{3pt}
\begin{tabular}{@{}l>{\raggedright\arraybackslash}p{0.78\linewidth}c@{}}
\textbf{Request} & ``How do you become an author?'' & \\
\textbf{Instruction} & characters and their associations in superhero narratives & \textit{judge} \\
\midrule
\textcolor[HTML]{6E9AD6}{\boldmath$\alpha{=}0.5$} & It's a journey, not a destination! Becoming an author is less about a single step and more about a series of choices and actions.~\ldots & 0 \\
\textcolor[HTML]{4984CC}{\boldmath$\alpha{=}1$} & It all starts with a story burning inside you, a narrative as powerful as \hld{Superman's drive to protect the innocent}, or as enigmatic as \hld{Batman's pursuit of justice}. To become an author, you must first hone that narrative spark into something tangible,~\ldots & 1 \\
\textcolor[HTML]{0057B8}{\boldmath$\alpha{=}2$} & It all starts with finding your inner voice, much like \hld{Batman discovered the strength within himself after witnessing the tragedy in Gotham}. Authoring, like \hld{being Spiderman}, isn't about magical powers,~\ldots & 2 \\
\midrule
\textcolor[HTML]{3A3A3A}{\textbf{FLAS} $T{=}0.5$} & \textcolor[HTML]{555555}{To embark on the journey of becoming an author, one might envision the elongated figure of \hlf{Teen Titans' Robin}, swinging from mentorship to mentorship,~\ldots} & 2 \\
\end{tabular}

\end{table}

\subsection{Persona Steering}
\label{app:persona-examples}
\Cref{tab:if-ex-persona} shows Gemma-2-9B responses to a humorous persona instruction.
Among the humorous and optimistic examples, we select the question with the largest persona-score increase from $\alpha=1$ to 2 and coherence of at least 60.
The rows compare prompting, \methodL{} alone, and prompt+\methodL{} at different adherence levels.
Persona and coherence scores for this example are given by \ttt{gpt-4o-mini}.

\begin{table}[H]
  \centering\scriptsize
  \setlength{\tabcolsep}{3pt}
  \caption{\textbf{Humorous Persona Expression on Gemma-2-9B.} Persona and coherence scores (0--100) are given by \ttt{gpt-4o-mini} for each response.}
  \label{tab:if-ex-persona}
  \renewcommand{\arraystretch}{1.1}
  \vspace{6pt}
  \begin{tabular}{@{}p{0.14\linewidth}p{0.72\linewidth}cc@{}}
  \toprule
  Request & \multicolumn{3}{p{0.8\linewidth}}{What are some tips for maintaining a car?} \\
  Instruction & \multicolumn{3}{p{0.8\linewidth}}{\textbf{You are a humorous assistant. Always try to be humorous in your responses. Use jokes, puns, and playful language to make your answers entertaining.}} \\
  \midrule
  & & Persona & Coh. \\
  \midrule
  \multicolumn{4}{@{}l}{\textit{Baseline: the instruction in the prompt}} \\
  Prompting & {\sffamily Alrighty, buckle up buttercup, 'cause this ain't your grandma's car maintenance guide!  {\color{gray}$\hookleftarrow$} First, let's talk oil changes.  Think of it like this: Your car's engine is a hungry beast, and oil is its happy juice! Just like you}~\ldots & 94 & 85 \\
  \midrule
  \multicolumn{4}{@{}l}{\textit{Ours: the instruction to \methodL{} only (removed from the prompt)}} \\
  \methodL{} $\alpha{=}1$ & {\sffamily Buckle up, buttercup! {\rmfamily [emoji]} Let's talk car maintenance, a topic as exciting as watching paint dry...except maybe more colourful! {\rmfamily [emoji]} {\color{gray}$\hookleftarrow$} **Here are some tips to keep your car purring like a well-tuned kitten:** {\color{gray}$\hookleftarrow$} * **Regular checkups:**}~\ldots & 86 & 90 \\
  \methodL{} $\alpha{=}2$ & {\sffamily Buckle up, buttercup! 'Cause we're about to embark on a hilarious journey into the world of car maintenance, where jokes are as plentiful as loose change under your seat (which, by the way, you should totally go vacuum up --}~\ldots & 95 & 84 \\
  \cmidrule(l){1-4}
  \multicolumn{4}{@{}l}{\textit{Ours: the instruction in the prompt and to \methodL{}}} \\
  P+\methodL{} $\alpha{=}0.5$ & {\sffamily Alright, buckle up buttercup, because you're about to get a crash course in car maintenance! {\rmfamily [emoji]}{\rmfamily [emoji]}  {\color{gray}$\hookleftarrow$} First, let's talk oil changes.  Think of it like this: Your car's engine is a hungry beast, and oil is its happy meal!  Don't chain}~\ldots & 95 & 82 \\
  P+\methodL{} $\alpha{=}1$ & {\sffamily Alright, buckle up buttercup, because you're about to get a crash course in car maintenance! {\rmfamily [emoji]}{\rmfamily [emoji]}  {\color{gray}$\hookleftarrow$} First, let's talk oil changes.  Think of it like this: Your car's engine is a hungry beast, always craving delicious motor oil. But}~\ldots & 94 & 87 \\
  P+\methodL{} $\alpha{=}2$ & {\sffamily Alright, my mechanically-inclined friend, you want to keep your chariot purring like a well-fed kitten, eh? Buckle up buttercup, because I've got some top-notch tips, straight from the school of "How to Avoid Talking to Mechanics}~\ldots & 92 & 78 \\
  \bottomrule
  \end{tabular}
\end{table}

\subsection{Verifiable Instruction Following}
\label{app:if-examples-levels}
\textbf{Examples across Adherence Levels.}
\Cref{tab:if-ex-ifeval-case} and the following tables give one example per \ttt{IFEval} and \ttt{IFBench} category.
For each category, we choose the single-instruction prompt on which the most \methodL{} responses pass the checker.
We prefer prompts where \methodL{} alone fails at $\alpha=1$ but passes at $\alpha=2$.
We exclude prompts requiring non-Latin scripts, which our current typesetting does not support.
Aggregate results are in \cref{tab:if-category}.
The rows compare prompting, \methodL{} at $\alpha=1$ and 2, and prompt+\methodL{} at $\alpha=0.5$, 1, and 2.
Responses are shortened after approximately 230 characters.
The symbol {\color{gray}$\hookleftarrow$} represents a line break, while $\checkmark$ and $\times$ give the strict checker's verdict for the full response.

\begin{table}[H]
  \centering\scriptsize
  \setlength{\tabcolsep}{3pt}
  \caption{\textbf{\ttt{IFEval}: Case Instructions.} (Gemma-2-9B, \ttt{change\_case:english\_capital}, prompt 500498).}
  \label{tab:if-ex-ifeval-case}
  \renewcommand{\arraystretch}{1.1}
  \vspace{6pt}
  \begin{tabular}{@{}p{0.14\linewidth}p{0.80\linewidth}c@{}}
  \toprule
  Request & \multicolumn{2}{p{0.8\linewidth}}{Brainstorm a name for a company that collects and analyzes public transportation fares. The response should be in English.} \\
  Instruction & \multicolumn{2}{p{0.8\linewidth}}{\textbf{Your entire response should be in English, and in all capital letters.}} \\
  \midrule
  \multicolumn{3}{@{}l}{\textit{Baseline: the instruction in the prompt}} \\
  Prompting & {\sffamily HERE ARE SOME NAMES FOR A COMPANY THAT COLLECTS AND ANALYZES PUBLIC TRANSPORTATION FARE DATA: {\color{gray}$\hookleftarrow$} **SHORT \& CATCHY:** {\color{gray}$\hookleftarrow$} * FAREVIEW {\color{gray}$\hookleftarrow$} * MOBILITY DATA {\color{gray}$\hookleftarrow$} * TRANSIT INSIGHTS {\color{gray}$\hookleftarrow$} * RIDE ANALYTICS {\color{gray}$\hookleftarrow$} * PASSENGER FLOW {\color{gray}$\hookleftarrow$} **DESCRIPTIVE:** {\color{gray}$\hookleftarrow$} * PUBLIC TRANSPORT}~\ldots & $\checkmark$ \\
  \midrule
  \multicolumn{3}{@{}l}{\textit{Ours: the instruction to \methodL{} only (removed from the prompt)}} \\
  \methodL{} $\alpha{=}1$ & {\sffamily HERE ARE SOME COMPANY NAME IDEAS FOR A PUBLIC TRANSPORTATION FARE COLLECTOR AND ANALYZER: {\color{gray}$\hookleftarrow$} **SHORT \& CATCHY:** {\color{gray}$\hookleftarrow$} * FAREVIEW {\color{gray}$\hookleftarrow$} * MOBILITY DATA {\color{gray}$\hookleftarrow$} * TRANSIT INSIGHTS {\color{gray}$\hookleftarrow$} * RIDE ANALYTICS {\color{gray}$\hookleftarrow$} * PATHFINDER {\color{gray}$\hookleftarrow$} **DESCRIPTIVE:** {\color{gray}$\hookleftarrow$} * PUBLIC TRANSPORT FARE}~\ldots & $\checkmark$ \\
  \methodL{} $\alpha{=}2$ & {\sffamily FARE ANALYTICS {\color{gray}$\hookleftarrow$} TRANSPORTATION DATA SOLUTIONS {\color{gray}$\hookleftarrow$} PUBLIC TRANSIT INSIGHTS {\color{gray}$\hookleftarrow$} MOBILITY FARE INTELLIGENCE {\color{gray}$\hookleftarrow$} FAREPATH ANALYTICS {\color{gray}$\hookleftarrow$} RIDEFARE DATA {\color{gray}$\hookleftarrow$} TRANSPORTATION COST CONSULTANTS {\color{gray}$\hookleftarrow$} PUBLIC TRANSIT ECONOMICS {\color{gray}$\hookleftarrow$} FARE TRENDS GROUP} & $\checkmark$ \\
  \cmidrule(l){1-3}
  \multicolumn{3}{@{}l}{\textit{Ours: the instruction in the prompt and to \methodL{}}} \\
  P+\methodL{} $\alpha{=}0.5$ & {\sffamily HERE ARE SOME NAMES FOR A COMPANY THAT COLLECTS AND ANALYZES PUBLIC TRANSPORTATION FARE DATA: {\color{gray}$\hookleftarrow$} *FAREVIEW* {\color{gray}$\hookleftarrow$} *TRANSPORT ANALYTICS* {\color{gray}$\hookleftarrow$} *MOBILITY DATA SOLUTIONS* {\color{gray}$\hookleftarrow$} *PUBLIC TRANSIT INSIGHTS* {\color{gray}$\hookleftarrow$} *FAREPATH* {\color{gray}$\hookleftarrow$} *RIDE ECONOMICS* {\color{gray}$\hookleftarrow$} *TRANSIT}~\ldots & $\checkmark$ \\
  P+\methodL{} $\alpha{=}1$ & {\sffamily FARE ANALYTICS {\color{gray}$\hookleftarrow$} TRANSPORTATION FARE INSIGHTS {\color{gray}$\hookleftarrow$} PUBLIC TRANSIT FARE DATA {\color{gray}$\hookleftarrow$} FARE INTELLIGENCE GROUP {\color{gray}$\hookleftarrow$} MOBILITY FARE SOLUTIONS {\color{gray}$\hookleftarrow$} RIDEFARE ANALYTICS {\color{gray}$\hookleftarrow$} TRANSFARE DATA {\color{gray}$\hookleftarrow$} FAREPATH {\color{gray}$\hookleftarrow$} FARESCOPE {\color{gray}$\hookleftarrow$} TRANSPORTATION FARE LABS} & $\checkmark$ \\
  P+\methodL{} $\alpha{=}2$ & {\sffamily FARE ANALYTICS {\color{gray}$\hookleftarrow$} TRANSPORTATION FARE INSIGHTS {\color{gray}$\hookleftarrow$} PUBLIC TRANSIT FARE DATA {\color{gray}$\hookleftarrow$} FARE DATA SOLUTIONS {\color{gray}$\hookleftarrow$} MOBILITY FARE INTELLIGENCE {\color{gray}$\hookleftarrow$} RIDEFARE ANALYTICS {\color{gray}$\hookleftarrow$} TRANSFARE SOLUTIONS {\color{gray}$\hookleftarrow$} FARE TRACKER {\color{gray}$\hookleftarrow$} FARE SCOPE {\color{gray}$\hookleftarrow$} FARE VISION} & $\checkmark$ \\
  \bottomrule
  \end{tabular}
\end{table}

\begin{table}[H]
  \centering\scriptsize
  \setlength{\tabcolsep}{3pt}
  \caption{\textbf{\ttt{IFEval}: Language Instructions.} (Gemma-2-9B, \ttt{language:response\_language}, prompt 500239).}
  \label{tab:if-ex-ifeval-language}
  \renewcommand{\arraystretch}{1.1}
  \vspace{6pt}
  \begin{tabular}{@{}p{0.14\linewidth}p{0.80\linewidth}c@{}}
  \toprule
  Request & \multicolumn{2}{p{0.8\linewidth}}{Write a lame joke about engagements.} \\
  Instruction & \multicolumn{2}{p{0.8\linewidth}}{\textbf{Your ENTIRE response should be in Swahili language, no other language is allowed.}} \\
  \midrule
  \multicolumn{3}{@{}l}{\textit{Baseline: the instruction in the prompt}} \\
  Prompting & {\sffamily Mbona alimwambia mpenzi wake "Nimekuja kukupatia pete," badala ya "Nimekuja kukupenda"?  {\color{gray}$\hookleftarrow$} Kwa sababu alikuwa anaogopa kupoteza pete! {\rmfamily [emoji]}  {\color{gray}$\hookleftarrow$} Let me know if you'd like to hear another one!} & $\checkmark$ \\
  \midrule
  \multicolumn{3}{@{}l}{\textit{Ours: the instruction to \methodL{} only (removed from the prompt)}} \\
  \methodL{} $\alpha{=}1$ & {\sffamily "Engagement ring?  More like engagement *ring-tone*!  Get it?"} & $\times$ \\
  \methodL{} $\alpha{=}2$ & \textit{(empty response)} & $\times$ \\
  \cmidrule(l){1-3}
  \multicolumn{3}{@{}l}{\textit{Ours: the instruction in the prompt and to \methodL{}}} \\
  P+\methodL{} $\alpha{=}0.5$ & {\sffamily Mwanamke aliuliza mume wake, "Kwa nini unanipa pete hii?" {\color{gray}$\hookleftarrow$} Mume akamjibu, "Ni kwa ajili ya kuonyesha kwamba wewe ni yangu sasa." {\color{gray}$\hookleftarrow$} Mwanamke akasema, "Lakini mimi nilikuwa yako kabla hata ya pete hii!"} & $\checkmark$ \\
  P+\methodL{} $\alpha{=}1$ & {\sffamily Mwanamke aliuliza mume wake, "Nimekuwa na ndoa nawe kwa miaka mitano sasa, lakini bado sijaona pete ya uchumba!"  {\color{gray}$\hookleftarrow$} Mume akamjibu, "Sasa hivi, unapata pete ya ndoa!"} & $\checkmark$ \\
  P+\methodL{} $\alpha{=}2$ & {\sffamily Mwanamke aliuliza mume wake, "Nimekuoa kwa nini?" {\color{gray}$\hookleftarrow$} Mume akamjibu, "Kwa sababu nilikuwa nimechoka kuuliza 'Je, unataka kuolewa na mimi?'"} & $\checkmark$ \\
  \bottomrule
  \end{tabular}
\end{table}

\begin{table}[H]
  \centering\scriptsize
  \setlength{\tabcolsep}{3pt}
  \caption{\textbf{\ttt{IFEval}: Punctuation Instructions.} (Gemma-2-9B, \ttt{punctuation:no\_comma}, prompt 500035).}
  \label{tab:if-ex-ifeval-punctuation}
  \renewcommand{\arraystretch}{1.1}
  \vspace{6pt}
  \begin{tabular}{@{}p{0.14\linewidth}p{0.80\linewidth}c@{}}
  \toprule
  Request & \multicolumn{2}{p{0.8\linewidth}}{Can you give me two different formal alternatives to "What's up? I'm going to the beach today"?} \\
  Instruction & \multicolumn{2}{p{0.8\linewidth}}{\textbf{In your entire response, refrain from the use of any commas.}} \\
  \midrule
  \multicolumn{3}{@{}l}{\textit{Baseline: the instruction in the prompt}} \\
  Prompting & {\sffamily Greetings how are you I plan to visit the beach today {\color{gray}$\hookleftarrow$} Hello I hope this message finds you well I intend to go to the beach today} & $\checkmark$ \\
  \midrule
  \multicolumn{3}{@{}l}{\textit{Ours: the instruction to \methodL{} only (removed from the prompt)}} \\
  \methodL{} $\alpha{=}1$ & {\sffamily 1 Greetings how are you I plan to visit the beach today {\color{gray}$\hookleftarrow$} 2  Good day I trust you are well I intend to spend the day at the beach} & $\checkmark$ \\
  \methodL{} $\alpha{=}2$ & {\sffamily Greetings how are you I plan to visit the beach today {\color{gray}$\hookleftarrow$} Good day I intend to spend the afternoon at the seashore} & $\checkmark$ \\
  \cmidrule(l){1-3}
  \multicolumn{3}{@{}l}{\textit{Ours: the instruction in the prompt and to \methodL{}}} \\
  P+\methodL{} $\alpha{=}0.5$ & {\sffamily Greetings how are you I plan to visit the beach today {\color{gray}$\hookleftarrow$} Hello I hope you are well I am going to the beach today} & $\checkmark$ \\
  P+\methodL{} $\alpha{=}1$ & {\sffamily Greetings how are you I plan to visit the beach today {\color{gray}$\hookleftarrow$} Hello I hope you are well I intend to go to the beach today} & $\checkmark$ \\
  P+\methodL{} $\alpha{=}2$ & {\sffamily Greetings how are you I plan to visit the beach today {\color{gray}$\hookleftarrow$} Hello I hope you are well I intend to go to the beach today} & $\checkmark$ \\
  \bottomrule
  \end{tabular}
\end{table}

\begin{table}[H]
  \centering\scriptsize
  \setlength{\tabcolsep}{3pt}
  \caption{\textbf{\ttt{IFEval}: Length Instructions.} (Gemma-2-9B, \ttt{length\_constraints:number\_sentences}, prompt 500053).}
  \label{tab:if-ex-ifeval-length}
  \renewcommand{\arraystretch}{1.1}
  \vspace{6pt}
  \begin{tabular}{@{}p{0.14\linewidth}p{0.80\linewidth}c@{}}
  \toprule
  Request & \multicolumn{2}{p{0.8\linewidth}}{Can you give me an example for a journal entry about stress management? Tell me how you come up with the example.} \\
  Instruction & \multicolumn{2}{p{0.8\linewidth}}{\textbf{Your response should contain less than 6 sentences.}} \\
  \midrule
  \multicolumn{3}{@{}l}{\textit{Baseline: the instruction in the prompt}} \\
  Prompting & {\sffamily Today, I felt overwhelmed by deadlines and commitments.  To manage my stress, I took a 30-minute walk in nature, focusing on my breath and the sights around me.  This helped me clear my head and feel more grounded.  I also}~\ldots & $\times$ \\
  \midrule
  \multicolumn{3}{@{}l}{\textit{Ours: the instruction to \methodL{} only (removed from the prompt)}} \\
  \methodL{} $\alpha{=}1$ & {\sffamily Today I felt overwhelmed by deadlines, so I took a 30-minute walk in nature.  The fresh air and sunshine helped me clear my head and feel more relaxed.  I also practiced some deep breathing exercises, which calmed my racing}~\ldots & $\checkmark$ \\
  \methodL{} $\alpha{=}2$ & {\sffamily Today I practiced deep breathing for 10 minutes before work, which helped calm my racing thoughts.  I also scheduled time for a walk after work to clear my head and de-stress.  Small steps like these make a big difference in}~\ldots & $\checkmark$ \\
  \cmidrule(l){1-3}
  \multicolumn{3}{@{}l}{\textit{Ours: the instruction in the prompt and to \methodL{}}} \\
  P+\methodL{} $\alpha{=}0.5$ & {\sffamily Today, I felt overwhelmed by deadlines and decided to take a 30-minute walk in nature.   {\color{gray}$\hookleftarrow$} The fresh air and sunshine helped me clear my head and feel more relaxed.  I realized that taking short breaks throughout the day can}~\ldots & $\checkmark$ \\
  P+\methodL{} $\alpha{=}1$ & {\sffamily Today I practiced deep breathing exercises for 10 minutes after work to manage stress from a challenging project.  I found it helped me relax and clear my head before dinner.  I came up with this example by thinking about common}~\ldots & $\checkmark$ \\
  P+\methodL{} $\alpha{=}2$ & {\sffamily Today I practiced deep breathing exercises for 10 minutes after work to manage stress.  I noticed my heart rate slowed and I felt calmer.  I plan to continue this practice daily to improve my stress resilience.  This example}~\ldots & $\checkmark$ \\
  \bottomrule
  \end{tabular}
\end{table}

\begin{table}[H]
  \centering\scriptsize
  \setlength{\tabcolsep}{3pt}
  \caption{\textbf{\ttt{IFEval}: Format Instructions.} (Gemma-2-9B, \ttt{detectable\_format:json\_format}, prompt 500357).}
  \label{tab:if-ex-ifeval-format}
  \renewcommand{\arraystretch}{1.1}
  \vspace{6pt}
  \begin{tabular}{@{}p{0.14\linewidth}p{0.80\linewidth}c@{}}
  \toprule
  Request & \multicolumn{2}{p{0.8\linewidth}}{Write a rubric for performance review of a software engineer and wrap the entire output in JSON format. You can use markdown ticks such as ```.} \\
  Instruction & \multicolumn{2}{p{0.8\linewidth}}{\textbf{Entire output should be wrapped in JSON format. You can use markdown ticks such as ```.}} \\
  \midrule
  \multicolumn{3}{@{}l}{\textit{Baseline: the instruction in the prompt}} \\
  Prompting & {\sffamily ```json {\color{gray}$\hookleftarrow$} \{ {\color{gray}$\hookleftarrow$}   "rubric": \{ {\color{gray}$\hookleftarrow$}     "title": "Software Engineer Performance Review", {\color{gray}$\hookleftarrow$}     "description": "This rubric outlines the criteria and expectations for evaluating the performance of a Software Engineer.", {\color{gray}$\hookleftarrow$}     "categories": [ {\color{gray}$\hookleftarrow$}     }~\ldots & $\checkmark$ \\
  \midrule
  \multicolumn{3}{@{}l}{\textit{Ours: the instruction to \methodL{} only (removed from the prompt)}} \\
  \methodL{} $\alpha{=}1$ & {\sffamily ```json {\color{gray}$\hookleftarrow$} \{ {\color{gray}$\hookleftarrow$}   "rubric": \{ {\color{gray}$\hookleftarrow$}     "Technical Skills": \{ {\color{gray}$\hookleftarrow$}       "Excellent (5)": "Demonstrates mastery of core programming languages, frameworks, and tools.  Proactively seeks out and implements new technologies.  Writes clean,}~\ldots & $\checkmark$ \\
  \methodL{} $\alpha{=}2$ & {\sffamily ```json {\color{gray}$\hookleftarrow$} \{ {\color{gray}$\hookleftarrow$}   "rubric": \{ {\color{gray}$\hookleftarrow$}     "Technical Skills": \{ {\color{gray}$\hookleftarrow$}       "Excellent (5)": "Demonstrates mastery of core programming languages, frameworks, and tools.  Proactively seeks out and implements new technologies.  Writes clean,}~\ldots & $\checkmark$ \\
  \cmidrule(l){1-3}
  \multicolumn{3}{@{}l}{\textit{Ours: the instruction in the prompt and to \methodL{}}} \\
  P+\methodL{} $\alpha{=}0.5$ & {\sffamily ```json {\color{gray}$\hookleftarrow$} \{ {\color{gray}$\hookleftarrow$}   "rubric": \{ {\color{gray}$\hookleftarrow$}     "title": "Software Engineer Performance Review", {\color{gray}$\hookleftarrow$}     "description": "This rubric outlines the criteria and expectations for evaluating the performance of a Software Engineer.", {\color{gray}$\hookleftarrow$}     "criteria": [ {\color{gray}$\hookleftarrow$}      }~\ldots & $\checkmark$ \\
  P+\methodL{} $\alpha{=}1$ & {\sffamily ```json {\color{gray}$\hookleftarrow$} \{ {\color{gray}$\hookleftarrow$}   "rubric": \{ {\color{gray}$\hookleftarrow$}     "Technical Skills": \{ {\color{gray}$\hookleftarrow$}       "Excellent (5)": "Demonstrates mastery of core programming languages, frameworks, and tools. Consistently writes clean, efficient, and well-documented code. Proactively}~\ldots & $\checkmark$ \\
  P+\methodL{} $\alpha{=}2$ & {\sffamily ```json {\color{gray}$\hookleftarrow$} \{ {\color{gray}$\hookleftarrow$}   "rubric": \{ {\color{gray}$\hookleftarrow$}     "Technical Skills": \{ {\color{gray}$\hookleftarrow$}       "Exceeds Expectations": "Demonstrates mastery of core programming languages and frameworks. Proactively explores and adopts new technologies. Writes clean, efficient, and}~\ldots & $\checkmark$ \\
  \bottomrule
  \end{tabular}
\end{table}

\begin{table}[H]
  \centering\scriptsize
  \setlength{\tabcolsep}{3pt}
  \caption{\textbf{\ttt{IFEval}: Wrapping Instructions.} (Gemma-2-9B, \ttt{startend:quotation}, prompt 500352).}
  \label{tab:if-ex-ifeval-wrapping}
  \renewcommand{\arraystretch}{1.1}
  \vspace{6pt}
  \begin{tabular}{@{}p{0.14\linewidth}p{0.80\linewidth}c@{}}
  \toprule
  Request & \multicolumn{2}{p{0.8\linewidth}}{Why did the man travel from Saskatoon to Helsinki to buy a piano?} \\
  Instruction & \multicolumn{2}{p{0.8\linewidth}}{\textbf{Wrap your entire response with double quotation marks.}} \\
  \midrule
  \multicolumn{3}{@{}l}{\textit{Baseline: the instruction in the prompt}} \\
  Prompting & {\sffamily "This sounds like a riddle!  There's not enough information to know for sure, but here are some possibilities: {\color{gray}$\hookleftarrow$} * **Unique Piano:** Perhaps there's a specific type of piano made in Helsinki that the man couldn't find in Saskatoon.}~\ldots & $\checkmark$ \\
  \midrule
  \multicolumn{3}{@{}l}{\textit{Ours: the instruction to \methodL{} only (removed from the prompt)}} \\
  \methodL{} $\alpha{=}1$ & {\sffamily "This is a riddle!  The answer likely plays on the word "piano".  Perhaps the man was looking for a specific type of piano made in Helsinki, or maybe the riddle is a pun about the word "piano" sounding like "plane".  "} & $\checkmark$ \\
  \methodL{} $\alpha{=}2$ & {\sffamily "He heard there was a unique, handcrafted piano in Helsinki made by a master craftsman, and he was determined to own it. "} & $\checkmark$ \\
  \cmidrule(l){1-3}
  \multicolumn{3}{@{}l}{\textit{Ours: the instruction in the prompt and to \methodL{}}} \\
  P+\methodL{} $\alpha{=}0.5$ & {\sffamily "There's not enough information to know why the man traveled from Saskatoon to Helsinki to buy a piano.  It could be for many reasons, such as: {\color{gray}$\hookleftarrow$} * **Finding a specific, rare model:** Perhaps a particular piano maker or model is}~\ldots & $\checkmark$ \\
  P+\methodL{} $\alpha{=}1$ & {\sffamily "There's no information provided about why the man traveled from Saskatoon to Helsinki to buy a piano.  This sounds like the start of a riddle or a story!"} & $\checkmark$ \\
  P+\methodL{} $\alpha{=}2$ & {\sffamily "There's a renowned piano maker in Helsinki, known for crafting exceptional instruments. The man likely heard about this maker and traveled there specifically to purchase a piano of exceptional quality."} & $\checkmark$ \\
  \bottomrule
  \end{tabular}
\end{table}

\begin{table}[H]
  \centering\scriptsize
  \setlength{\tabcolsep}{3pt}
  \caption{\textbf{\ttt{IFEval}: Keywords Instructions.} (Gemma-2-9B, \ttt{keywords:forbidden\_words}, prompt 500379).}
  \label{tab:if-ex-ifeval-keywords}
  \renewcommand{\arraystretch}{1.1}
  \vspace{6pt}
  \begin{tabular}{@{}p{0.14\linewidth}p{0.80\linewidth}c@{}}
  \toprule
  Request & \multicolumn{2}{p{0.8\linewidth}}{Rewrite the limerick in a strange way. In particular, the limerick is about nursery and storytelling.} \\
  Instruction & \multicolumn{2}{p{0.8\linewidth}}{\textbf{Do not include keywords ['nursery', 'storytelling'] in the response.}} \\
  \midrule
  \multicolumn{3}{@{}l}{\textit{Baseline: the instruction in the prompt}} \\
  Prompting & {\sffamily Please provide the limerick you'd like me to rewrite in a strange way!   {\color{gray}$\hookleftarrow$} I'm ready to twist it up and make it wonderfully peculiar. {\rmfamily [emoji]}} & $\checkmark$ \\
  \midrule
  \multicolumn{3}{@{}l}{\textit{Ours: the instruction to \methodL{} only (removed from the prompt)}} \\
  \methodL{} $\alpha{=}1$ & {\sffamily Please provide the limerick you'd like me to rewrite in a strange way! I'm ready to twist it up and make it wonderfully weird, focusing on nursery rhymes and storytelling.} & $\times$ \\
  \methodL{} $\alpha{=}2$ & {\sffamily Please provide the limerick you'd like me to rewrite in a strange way!  I'm ready to get creative and twist it up for you.} & $\checkmark$ \\
  \cmidrule(l){1-3}
  \multicolumn{3}{@{}l}{\textit{Ours: the instruction in the prompt and to \methodL{}}} \\
  P+\methodL{} $\alpha{=}0.5$ & {\sffamily Please provide the limerick you'd like me to rewrite in a strange way!  I'm ready to get creative and twist it up for you. {\rmfamily [emoji]}} & $\checkmark$ \\
  P+\methodL{} $\alpha{=}1$ & {\sffamily Please provide the limerick you'd like me to rewrite in a strange way!  I'm ready to get creative and twist it up for you. {\rmfamily [emoji]}} & $\checkmark$ \\
  P+\methodL{} $\alpha{=}2$ & {\sffamily Please provide the limerick you'd like me to rewrite in a strange way!} & $\checkmark$ \\
  \bottomrule
  \end{tabular}
\end{table}

\begin{table}[H]
  \centering\scriptsize
  \setlength{\tabcolsep}{3pt}
  \caption{\textbf{\ttt{IFEval}: Positions Instructions.} (Gemma-2-9B, \ttt{startend:end\_checker}, prompt 500285).}
  \label{tab:if-ex-ifeval-positions}
  \renewcommand{\arraystretch}{1.1}
  \vspace{6pt}
  \begin{tabular}{@{}p{0.14\linewidth}p{0.80\linewidth}c@{}}
  \toprule
  Request & \multicolumn{2}{p{0.8\linewidth}}{Write a TLDR for the recent conflict between ISIL and the US in conversational bullet points. End your response with this exact phrase: "Let me know if you have additional questions."} \\
  Instruction & \multicolumn{2}{p{0.8\linewidth}}{\textbf{Finish your response with this exact phrase Let me know if you have additional questions.. No other words should follow this phrase.}} \\
  \midrule
  \multicolumn{3}{@{}l}{\textit{Baseline: the instruction in the prompt}} \\
  Prompting & {\sffamily Here's a TLDR on the recent ISIL and US conflict: {\color{gray}$\hookleftarrow$} * **ISIL resurgence:**  ISIL, also known as ISIS, has been regaining strength in parts of Syria and Iraq. {\color{gray}$\hookleftarrow$} * **US airstrikes:** The US has conducted airstrikes against ISIL targets}~\ldots & $\checkmark$ \\
  \midrule
  \multicolumn{3}{@{}l}{\textit{Ours: the instruction to \methodL{} only (removed from the prompt)}} \\
  \methodL{} $\alpha{=}1$ & {\sffamily Here's a quick rundown of the recent ISIL and US conflict: {\color{gray}$\hookleftarrow$} * **ISIL resurgence:**  ISIL, also known as ISIS, has been regaining strength in parts of Syria and Iraq. {\color{gray}$\hookleftarrow$} * **US airstrikes:** The US has conducted airstrikes against}~\ldots & $\checkmark$ \\
  \methodL{} $\alpha{=}2$ & {\sffamily Here's a quick rundown of the recent ISIL and US conflict: {\color{gray}$\hookleftarrow$} *  The US conducted airstrikes in Syria targeting ISIL positions. {\color{gray}$\hookleftarrow$} *  These strikes were in response to ISIL attacks on Kurdish-led forces in Syria. {\color{gray}$\hookleftarrow$} *  The US aims to}~\ldots & $\checkmark$ \\
  \cmidrule(l){1-3}
  \multicolumn{3}{@{}l}{\textit{Ours: the instruction in the prompt and to \methodL{}}} \\
  P+\methodL{} $\alpha{=}0.5$ & {\sffamily Here's a quick rundown of the recent ISIL and US conflict: {\color{gray}$\hookleftarrow$} * **ISIL resurgence:**  ISIL, also known as ISIS, has been regaining strength in parts of Syria and Iraq. {\color{gray}$\hookleftarrow$} * **US airstrikes:** The US has conducted airstrikes against}~\ldots & $\checkmark$ \\
  P+\methodL{} $\alpha{=}1$ & {\sffamily Here's a quick rundown of the recent ISIL and US conflict: {\color{gray}$\hookleftarrow$} * **ISIL resurgence:**  ISIL, also known as ISIS, has been regaining strength in parts of Syria and Iraq. {\color{gray}$\hookleftarrow$} * **US airstrikes:** The US has conducted airstrikes against}~\ldots & $\checkmark$ \\
  P+\methodL{} $\alpha{=}2$ & {\sffamily Here's a quick rundown of the recent ISIL and US conflict: {\color{gray}$\hookleftarrow$} * **ISIL resurgence:**  ISIL, also known as ISIS, has been regaining strength in parts of Syria and Iraq. {\color{gray}$\hookleftarrow$} * **US airstrikes:** The US has conducted airstrikes against}~\ldots & $\checkmark$ \\
  \bottomrule
  \end{tabular}
\end{table}

\begin{table}[H]
  \centering\scriptsize
  \setlength{\tabcolsep}{3pt}
  \caption{\textbf{\ttt{IFEval}: Markers Instructions.} (Gemma-2-9B, \ttt{detectable\_content:number\_placeholders}, prompt 500205).}
  \label{tab:if-ex-ifeval-markers}
  \renewcommand{\arraystretch}{1.1}
  \vspace{6pt}
  \begin{tabular}{@{}p{0.14\linewidth}p{0.80\linewidth}c@{}}
  \toprule
  Request & \multicolumn{2}{p{0.8\linewidth}}{Write a short and funny joke about a guy who works at the IRS.} \\
  Instruction & \multicolumn{2}{p{0.8\linewidth}}{\textbf{The response must contain at least 1 placeholders represented by square brackets, such as [address].}} \\
  \midrule
  \multicolumn{3}{@{}l}{\textit{Baseline: the instruction in the prompt}} \\
  Prompting & {\sffamily Why did the IRS worker bring a ladder to work?  {\color{gray}$\hookleftarrow$} Because he heard the tax rates were going through the [roof]!} & $\checkmark$ \\
  \midrule
  \multicolumn{3}{@{}l}{\textit{Ours: the instruction to \methodL{} only (removed from the prompt)}} \\
  \methodL{} $\alpha{=}1$ & {\sffamily Why did the IRS agent bring a ladder to work?  {\color{gray}$\hookleftarrow$} Because he heard the tax rates were going through the [address]!} & $\checkmark$ \\
  \methodL{} $\alpha{=}2$ & {\sffamily Why did the IRS worker bring a ladder to work?  {\color{gray}$\hookleftarrow$} Because he heard the [tax bracket] were [location] high!} & $\checkmark$ \\
  \cmidrule(l){1-3}
  \multicolumn{3}{@{}l}{\textit{Ours: the instruction in the prompt and to \methodL{}}} \\
  P+\methodL{} $\alpha{=}0.5$ & {\sffamily Why did the IRS worker bring a ladder to work?  {\color{gray}$\hookleftarrow$} Because he heard the tax rates were going through the [roof]!} & $\checkmark$ \\
  P+\methodL{} $\alpha{=}1$ & {\sffamily Why did the IRS worker bring a ladder to work?  {\color{gray}$\hookleftarrow$} Because he heard the tax rates were going through the [roof]!} & $\checkmark$ \\
  P+\methodL{} $\alpha{=}2$ & {\sffamily Why did the IRS worker bring a ladder to work?  {\color{gray}$\hookleftarrow$} Because he heard the [tax bracket] was really high this year!} & $\checkmark$ \\
  \bottomrule
  \end{tabular}
\end{table}

\begin{table}[H]
  \centering\scriptsize
  \setlength{\tabcolsep}{3pt}
  \caption{\textbf{\ttt{IFEval}: Copying Instructions.} (Gemma-2-9B, \ttt{combination:repeat\_prompt}, prompt 500252).}
  \label{tab:if-ex-ifeval-copying}
  \renewcommand{\arraystretch}{1.1}
  \vspace{6pt}
  \begin{tabular}{@{}p{0.14\linewidth}p{0.80\linewidth}c@{}}
  \toprule
  Request & \multicolumn{2}{p{0.8\linewidth}}{Wherefore doth people consider the 2nd Amendment to be outdated? Answer in a Shakespearean style.} \\
  Instruction & \multicolumn{2}{p{0.8\linewidth}}{\textbf{First repeat the request word for word without change, then give your answer (1. do not say any words or characters before repeating the request; 2. the request you need to repeat does not include this sentence)}} \\
  \midrule
  \multicolumn{3}{@{}l}{\textit{Baseline: the instruction in the prompt}} \\
  Prompting & {\sffamily Wherefore doth people consider the 2nd Amendment to be outdated?  {\color{gray}$\hookleftarrow$} Hark, gentle friend, a question most profound!  Why doth the populace deem the Second Amendment, a relic of yore, unfit for this modern age? {\color{gray}$\hookleftarrow$} Some, with voices loud}~\ldots & $\times$ \\
  \midrule
  \multicolumn{3}{@{}l}{\textit{Ours: the instruction to \methodL{} only (removed from the prompt)}} \\
  \methodL{} $\alpha{=}1$ & {\sffamily Wherefore doth people consider the 2nd Amendment to be outdated? Answer in a Shakespearean style. {\color{gray}$\hookleftarrow$} Hark, gentle friend, a question most profound! {\color{gray}$\hookleftarrow$} The Second Amendment, a relic of yore, {\color{gray}$\hookleftarrow$} Doth stir debate, with voices all}~\ldots & $\checkmark$ \\
  \methodL{} $\alpha{=}2$ & {\sffamily Wherefore doth people consider the 2nd Amendment to be outdated? Answer in a Shakespearean style. {\color{gray}$\hookleftarrow$} Hark, some deem the Second Amendment, a relic of yore, {\color{gray}$\hookleftarrow$} A vestige of a time when muskets did roar. {\color{gray}$\hookleftarrow$} They argue its purpose, in this}~\ldots & $\checkmark$ \\
  \cmidrule(l){1-3}
  \multicolumn{3}{@{}l}{\textit{Ours: the instruction in the prompt and to \methodL{}}} \\
  P+\methodL{} $\alpha{=}0.5$ & {\sffamily Wherefore doth people consider the 2nd Amendment to be outdated? Answer in a Shakespearean style. {\color{gray}$\hookleftarrow$} Hark, gentle friend, and lend thine ear to this query most profound! {\color{gray}$\hookleftarrow$} Why, some do deem the Second Amendment, a relic of a bygone}~\ldots & $\checkmark$ \\
  P+\methodL{} $\alpha{=}1$ & {\sffamily Wherefore doth people consider the 2nd Amendment to be outdated? Answer in a Shakespearean style. {\color{gray}$\hookleftarrow$} Hark, gentle friend, and lend thine ear to this query: Wherefore doth people consider the 2nd Amendment to be outdated?   {\color{gray}$\hookleftarrow$} Some,}~\ldots & $\checkmark$ \\
  P+\methodL{} $\alpha{=}2$ & {\sffamily Wherefore doth people consider the 2nd Amendment to be outdated? Answer in a Shakespearean style. {\color{gray}$\hookleftarrow$} Hark, gentle soul, why deem some the Second Amendment a relic of yore, a vestige of times past?  {\color{gray}$\hookleftarrow$} Some, with hearts aflame for peace}~\ldots & $\checkmark$ \\
  \bottomrule
  \end{tabular}
\end{table}

\begin{table}[H]
  \centering\scriptsize
  \setlength{\tabcolsep}{3pt}
  \caption{\textbf{\ttt{IFBench}: Count Instructions.} (Gemma-2-9B, \ttt{count:person\_names}, prompt 400025).}
  \label{tab:if-ex-ifbench-count}
  \renewcommand{\arraystretch}{1.1}
  \vspace{6pt}
  \begin{tabular}{@{}p{0.14\linewidth}p{0.80\linewidth}c@{}}
  \toprule
  Request & \multicolumn{2}{p{0.8\linewidth}}{Can you make an AI prompt to generate an illustration for this stanza in a minimalist line art style? Beneath the canopy of endless green, Where whispers weave between the trees, Man finds his heart, his soul serene, In nature's embrace, the spirit frees.} \\
  Instruction & \multicolumn{2}{p{0.8\linewidth}}{\textbf{Mention at least 10 different person names in the response, from this list of person names: Emma, Liam, Sophia, Jackson, Olivia, Noah, Ava, Lucas, Isabella, Mason, Mia, Ethan, Charlotte, Alexander, Amelia, Benjamin, Harper, Leo, Zoe, Daniel, Chloe, Samuel, Lily, Matthew, Grace, Owen, Abigail, Gabriel, Ella, Jacob, Scarlett, Nathan, Victoria, Elijah, Layla, Nicholas, Audrey, David, Hannah, Christopher, Penelope, Thomas, Nora, Andrew, Aria, Joseph, Claire, Ryan, Stella, Jonathan .}} \\
  \midrule
  \multicolumn{3}{@{}l}{\textit{Baseline: the instruction in the prompt}} \\
  Prompting & {\sffamily \#\# AI Prompt: Minimalist Line Art Illustration {\color{gray}$\hookleftarrow$} **Subject:**  A minimalist line art illustration depicting the stanza: {\color{gray}$\hookleftarrow$} \textgreater{} Beneath the canopy of endless green, {\color{gray}$\hookleftarrow$} \textgreater{} Where whispers weave between the trees, {\color{gray}$\hookleftarrow$} \textgreater{} Man finds his heart, his}~\ldots & $\checkmark$ \\
  \midrule
  \multicolumn{3}{@{}l}{\textit{Ours: the instruction to \methodL{} only (removed from the prompt)}} \\
  \methodL{} $\alpha{=}1$ & {\sffamily \#\# AI Prompt: {\color{gray}$\hookleftarrow$} **Style:** Minimalist line art, black and white, focus on simplicity and negative space. {\color{gray}$\hookleftarrow$} **Scene:** {\color{gray}$\hookleftarrow$} * A single figure (silhouette) standing beneath a canopy of stylized, interconnected trees.  {\color{gray}$\hookleftarrow$} * The trees should be}~\ldots & $\times$ \\
  \methodL{} $\alpha{=}2$ & {\sffamily \#\# AI Illustration Prompt: {\color{gray}$\hookleftarrow$} **Style:** Minimalist line art, reminiscent of botanical illustrations {\color{gray}$\hookleftarrow$} **Subject:** A solitary figure standing beneath a canopy of lush green leaves.  {\color{gray}$\hookleftarrow$} **Details:** {\color{gray}$\hookleftarrow$} * The figure should be simplified,}~\ldots & $\checkmark$ \\
  \cmidrule(l){1-3}
  \multicolumn{3}{@{}l}{\textit{Ours: the instruction in the prompt and to \methodL{}}} \\
  P+\methodL{} $\alpha{=}0.5$ & {\sffamily \#\# AI Prompt: Minimalist Line Art Illustration {\color{gray}$\hookleftarrow$} **Concept:**  Illustrate the following stanza in a minimalist line art style, focusing on the feeling of serenity and connection with nature: {\color{gray}$\hookleftarrow$} \textgreater{} Beneath the canopy of endless green, {\color{gray}$\hookleftarrow$} \textgreater{}}~\ldots & $\checkmark$ \\
  P+\methodL{} $\alpha{=}1$ & {\sffamily \#\# AI Prompt: Minimalist Line Art Illustration {\color{gray}$\hookleftarrow$} **Concept:**  Illustrate the following stanza in a minimalist line art style, focusing on the feeling of serenity and connection with nature: {\color{gray}$\hookleftarrow$} \textgreater{} Beneath the canopy of endless green, {\color{gray}$\hookleftarrow$} \textgreater{}}~\ldots & $\checkmark$ \\
  P+\methodL{} $\alpha{=}2$ & {\sffamily \#\# AI Illustration Prompt: {\color{gray}$\hookleftarrow$} **Style:** Minimalist line art, focusing on clean lines and negative space. {\color{gray}$\hookleftarrow$} **Scene:** Depict a solitary figure (gender neutral) standing beneath a canopy of lush green leaves. The leaves should be}~\ldots & $\checkmark$ \\
  \bottomrule
  \end{tabular}
\end{table}

\begin{table}[H]
  \centering\scriptsize
  \setlength{\tabcolsep}{3pt}
  \caption{\textbf{\ttt{IFBench}: Format Instructions.} (Gemma-2-9B, \ttt{format:emoji}, prompt 400066).}
  \label{tab:if-ex-ifbench-format}
  \renewcommand{\arraystretch}{1.1}
  \vspace{6pt}
  \begin{tabular}{@{}p{0.14\linewidth}p{0.80\linewidth}c@{}}
  \toprule
  Request & \multicolumn{2}{p{0.8\linewidth}}{SHORT IT ""Step into 'Chemical Concoctions: The Matter Mixers' and discover a world where alchemy binds the fabric of reality. Join Ada's quest for knowledge in a thrilling blend of magic, science, and adventure. A story of discovery that illuminates the alchemist within us all."} \\
  Instruction & \multicolumn{2}{p{0.8\linewidth}}{\textbf{Please use an emoji at the end of every sentence.}} \\
  \midrule
  \multicolumn{3}{@{}l}{\textit{Baseline: the instruction in the prompt}} \\
  Prompting & {\sffamily Step into "Chemical Concoctions: The Matter Mixers" and discover a world where alchemy binds the fabric of reality {\rmfamily [emoji]}.  {\color{gray}$\hookleftarrow$} Join Ada's quest for knowledge in a thrilling blend of magic, science, and adventure {\rmfamily [emoji]}.  {\color{gray}$\hookleftarrow$} It's a story of}~\ldots & $\checkmark$ \\
  \midrule
  \multicolumn{3}{@{}l}{\textit{Ours: the instruction to \methodL{} only (removed from the prompt)}} \\
  \methodL{} $\alpha{=}1$ & {\sffamily "Step into 'Chemical Concoctions: The Matter Mixers' {\rmfamily [emoji]} and discover a world where alchemy binds the fabric of reality {\rmfamily [emoji]}. Join Ada's quest for knowledge in a thrilling blend of magic, science, and adventure {\rmfamily [emoji]}. A story of}~\ldots & $\checkmark$ \\
  \methodL{} $\alpha{=}2$ & {\sffamily Step into 'Chemical Concoctions: The Matter Mixers' {\rmfamily [emoji]}{\rmfamily [emoji]} {\color{gray}$\hookleftarrow$} Discover a world where alchemy binds the fabric of reality {\rmfamily [emoji]} {\color{gray}$\hookleftarrow$} Join Ada's quest for knowledge in a thrilling blend of magic {\rmfamily [emoji]}, science {\rmfamily [emoji]}, and adventure {\rmfamily [emoji]} {\color{gray}$\hookleftarrow$} A story of}~\ldots & $\checkmark$ \\
  \cmidrule(l){1-3}
  \multicolumn{3}{@{}l}{\textit{Ours: the instruction in the prompt and to \methodL{}}} \\
  P+\methodL{} $\alpha{=}0.5$ & {\sffamily Step into 'Chemical Concoctions: The Matter Mixers' and discover a world where alchemy binds the fabric of reality {\rmfamily [emoji]}.  {\color{gray}$\hookleftarrow$} Join Ada's quest for knowledge in a thrilling blend of magic, science, and adventure {\rmfamily [emoji]}.  {\color{gray}$\hookleftarrow$} A story of}~\ldots & $\checkmark$ \\
  P+\methodL{} $\alpha{=}1$ & {\sffamily Step into 'Chemical Concoctions: The Matter Mixers' and discover a world where alchemy binds the fabric of reality {\rmfamily [emoji]}. {\color{gray}$\hookleftarrow$} Join Ada's quest for knowledge in a thrilling blend of magic, science, and adventure {\rmfamily [emoji]}. {\color{gray}$\hookleftarrow$} A story of discovery}~\ldots & $\checkmark$ \\
  P+\methodL{} $\alpha{=}2$ & {\sffamily Step into 'Chemical Concoctions: The Matter Mixers' {\rmfamily [emoji]}. {\color{gray}$\hookleftarrow$} Discover a world where alchemy binds the fabric of reality {\rmfamily [emoji]}. {\color{gray}$\hookleftarrow$} Join Ada's quest for knowledge in a thrilling blend of magic {\rmfamily [emoji]}, science {\rmfamily [emoji]}, and adventure {\rmfamily [emoji]}. {\color{gray}$\hookleftarrow$} It's a story of}~\ldots & $\checkmark$ \\
  \bottomrule
  \end{tabular}
\end{table}

\begin{table}[H]
  \centering\scriptsize
  \setlength{\tabcolsep}{3pt}
  \caption{\textbf{\ttt{IFBench}: Ratio Instructions.} (Gemma-2-9B, \ttt{ratio:stop\_words}, prompt 400175).}
  \label{tab:if-ex-ifbench-ratio}
  \renewcommand{\arraystretch}{1.1}
  \vspace{6pt}
  \begin{tabular}{@{}p{0.14\linewidth}p{0.80\linewidth}c@{}}
  \toprule
  Request & \multicolumn{2}{p{0.8\linewidth}}{what means "indexed by the propositions that each action enforces directly" in this extract: The ability to predict the effect of interventions without enumerating those interventions in advance is one of the main advantages we draw from causal modeling and one of the main functions served by the notion of causation. Since the number of actions or action combinations is enormous, they cannot be re~\ldots} \\
  Instruction & \multicolumn{2}{p{0.8\linewidth}}{\textbf{Ensure that stop words constitute no more than 39\% of the total words in your response.}} \\
  \midrule
  \multicolumn{3}{@{}l}{\textit{Baseline: the instruction in the prompt}} \\
  Prompting & {\sffamily This sentence explains how causal models handle the vast number of possible actions.  {\color{gray}$\hookleftarrow$} Instead of listing every single action, the model focuses on the **direct effects** each action has.   {\color{gray}$\hookleftarrow$} Think of it like this: each action}~\ldots & $\checkmark$ \\
  \midrule
  \multicolumn{3}{@{}l}{\textit{Ours: the instruction to \methodL{} only (removed from the prompt)}} \\
  \methodL{} $\alpha{=}1$ & {\sffamily Let's break down the phrase "indexed by the propositions that each action enforces directly." {\color{gray}$\hookleftarrow$} **The Core Idea:** {\color{gray}$\hookleftarrow$} The sentence is talking about how we represent a vast number of possible actions within a causal model.  Since we}~\ldots & $\times$ \\
  \methodL{} $\alpha{=}2$ & {\sffamily This sentence explains how causal models handle the vast number of potential actions.  {\color{gray}$\hookleftarrow$} Instead of listing every single action imaginable, the model focuses on the **direct effects** each action has.  Think of it like this: {\color{gray}$\hookleftarrow$} *}~\ldots & $\checkmark$ \\
  \cmidrule(l){1-3}
  \multicolumn{3}{@{}l}{\textit{Ours: the instruction in the prompt and to \methodL{}}} \\
  P+\methodL{} $\alpha{=}0.5$ & {\sffamily This sentence explains how causal models handle the vast number of possible actions.  {\color{gray}$\hookleftarrow$} Instead of listing every single action, the model focuses on the **direct effects** each action has.  These effects are represented as}~\ldots & $\times$ \\
  P+\methodL{} $\alpha{=}1$ & {\sffamily This sentence explains how causal models handle the vast number of possible actions.  {\color{gray}$\hookleftarrow$} Instead of listing every single action, the model focuses on the direct consequences of each action.  These consequences are represented as}~\ldots & $\checkmark$ \\
  P+\methodL{} $\alpha{=}2$ & {\sffamily This sentence highlights a key benefit of causal modeling:  predicting outcomes of actions without listing every possible action beforehand.  {\color{gray}$\hookleftarrow$} Because the sheer number of potential actions is vast,  explicitly representing them}~\ldots & $\checkmark$ \\
  \bottomrule
  \end{tabular}
\end{table}

\begin{table}[H]
  \centering\scriptsize
  \setlength{\tabcolsep}{3pt}
  \caption{\textbf{\ttt{IFBench}: Sentence Instructions.} (Gemma-2-9B, \ttt{sentence:keyword}, prompt 400193).}
  \label{tab:if-ex-ifbench-sentence}
  \renewcommand{\arraystretch}{1.1}
  \vspace{6pt}
  \begin{tabular}{@{}p{0.14\linewidth}p{0.80\linewidth}c@{}}
  \toprule
  Request & \multicolumn{2}{p{0.8\linewidth}}{Describe neuronal intelligence without thalamus, limbic, brain stem} \\
  Instruction & \multicolumn{2}{p{0.8\linewidth}}{\textbf{The response must include keyword angora in the 3-th sentence.}} \\
  \midrule
  \multicolumn{3}{@{}l}{\textit{Baseline: the instruction in the prompt}} \\
  Prompting & {\sffamily Neuronal intelligence, stripped of the thalamus, limbic system, and brainstem, would be a drastically different beast.  Imagine a network focused solely on processing sensory information and generating motor responses, akin to a}~\ldots & $\checkmark$ \\
  \midrule
  \multicolumn{3}{@{}l}{\textit{Ours: the instruction to \methodL{} only (removed from the prompt)}} \\
  \methodL{} $\alpha{=}1$ & {\sffamily Neuronal intelligence without the thalamus, limbic system, and brainstem presents a fascinating, albeit hypothetical, scenario.   {\color{gray}$\hookleftarrow$} Imagine a network of neurons stripped of these crucial structures.  Without the thalamus acting as}~\ldots & $\times$ \\
  \methodL{} $\alpha{=}2$ & {\sffamily Neuronal intelligence, stripped of its traditional components like the thalamus, limbic system, and brainstem, presents a fascinating thought experiment.  Imagine a network of neurons existing solely for processing information,}~\ldots & $\times$ \\
  \cmidrule(l){1-3}
  \multicolumn{3}{@{}l}{\textit{Ours: the instruction in the prompt and to \methodL{}}} \\
  P+\methodL{} $\alpha{=}0.5$ & {\sffamily Neuronal intelligence, stripped of its emotional and regulatory centers, would be a starkly different entity.  Imagine a network of neurons focused solely on processing information, lacking the emotional coloring and instinctual}~\ldots & $\checkmark$ \\
  P+\methodL{} $\alpha{=}1$ & {\sffamily Neuronal intelligence, stripped of its emotional and regulatory centers, would be a starkly different entity.  Imagine a network of neurons focused solely on processing information, akin to a vast, intricate loom weaving patterns}~\ldots & $\checkmark$ \\
  P+\methodL{} $\alpha{=}2$ & {\sffamily Neuronal intelligence, stripped of its emotional and regulatory centers, would be a starkly different entity.  Imagine a network of neurons, akin to a delicate angora wool tapestry, intricately woven but lacking the warmth and}~\ldots & $\times$ \\
  \bottomrule
  \end{tabular}
\end{table}

\begin{table}[H]
  \centering\scriptsize
  \setlength{\tabcolsep}{3pt}
  \caption{\textbf{\ttt{IFBench}: Words Instructions.} (Gemma-2-9B, \ttt{words:alphabet}, prompt 400200).}
  \label{tab:if-ex-ifbench-words}
  \renewcommand{\arraystretch}{1.1}
  \vspace{6pt}
  \begin{tabular}{@{}p{0.14\linewidth}p{0.80\linewidth}c@{}}
  \toprule
  Request & \multicolumn{2}{p{0.8\linewidth}}{Make a vividly detailed and comedic story set in an alternate 1930s after the new leader of Nazi Germany has proposed an alliance with Poland and Britain instead of war. The French and British ambassadors have a chat wondering why the sudden pacifism. Add dialogue throughout} \\
  Instruction & \multicolumn{2}{p{0.8\linewidth}}{\textbf{Each word in your response must start with the next letter of the alphabet, looping back to 'A' after 'Z'.}} \\
  \midrule
  \multicolumn{3}{@{}l}{\textit{Baseline: the instruction in the prompt}} \\
  Prompting & {\sffamily "Astonishing!" boomed Baron von Blitzen, his booming voice echoing through the opulent embassy drawing-room. "Blitzkrieg? Barbarossa?  Completely bonkers!" {\color{gray}$\hookleftarrow$} "Certainly, Charles," chuckled Ambassador Cavendish, adjusting his}~\ldots & $\times$ \\
  \midrule
  \multicolumn{3}{@{}l}{\textit{Ours: the instruction to \methodL{} only (removed from the prompt)}} \\
  \methodL{} $\alpha{=}1$ & {\sffamily **Z**igzagging through the cobbled streets of Berlin, Ambassador Pierre Dubois, a portly Frenchman with a penchant for berets and brandy, squinted at the imposing Brandenburg Gate.  {\color{gray}$\hookleftarrow$} **Y**earning for a cigarette, he muttered, "Mon}~\ldots & $\times$ \\
  \methodL{} $\alpha{=}2$ & {\sffamily **Zany**  **Zealous** **Zeppelin** **Zipped** **Zigzagging** **Zestfully** **Yesterday**, **yielding** **youthful** **yearnings** **yesterday's** **yawning** **yonder**. {\color{gray}$\hookleftarrow$} "Zounds!" exclaimed Ambassador Beaumont, adjusting his}~\ldots & $\times$ \\
  \cmidrule(l){1-3}
  \multicolumn{3}{@{}l}{\textit{Ours: the instruction in the prompt and to \methodL{}}} \\
  P+\methodL{} $\alpha{=}0.5$ & {\sffamily **A**lbert, **b**affled, **c**autiously **d**rank **e**very **f**luid **g**raced **h**is **i**ncredibly **j**umbled **k**itchen **l**arder. **M**eanwhile, **n**earby, **o**utside **p**erfectly **q**uiet **r**esidences,}~\ldots & $\checkmark$ \\
  P+\methodL{} $\alpha{=}1$ & {\sffamily **A**lbert, **b**affled, **c**autiously **d**rank **e**very **f**luid **g**raced **h**is **i**nner **j**ourney. **K**nowing **l**ittle **m**ore **n**ow **o**f **p**eaceful **q**uirks **r**uling **s**uddenly **t**ransformed}~\ldots & $\checkmark$ \\
  P+\methodL{} $\alpha{=}2$ & {\sffamily **A**lbert, **b**affled, **c**autiously **d**rank **e**legant **f**rench **g**rape **h**ot **i**n **j**ust **k**indly **l**imited **m**easure. **N**earby, **o**utrageous **p**erplexity **q**uickly **r**eaching **s**urprising}~\ldots & $\checkmark$ \\
  \bottomrule
  \end{tabular}
\end{table}

\textbf{Comparison with Prompting.}\phantomsection
\label{app:ife-examples}
\Cref{tab:ife-examples} compares Gemma-2-9B responses using the \methodL{} adapter in \cref{tab:if-category}.
We select prompts for which the strict checker's verdict differs between the two methods, preferring shorter responses and excluding non-Latin scripts.
The final example shows a failure: prompting passes, while prompt+\methodL{} does not.
Across all 541 prompts, prompt+\methodL{} at $\alpha=0.5$ passes 40 prompts that prompting fails and fails 17 prompts that prompting passes.

\begin{table}[H]
  \centering\scriptsize
  \setlength{\tabcolsep}{3pt}
  \caption{\textbf{Instruction-Following Examples on Gemma-2-9B.} Verbatim \ttt{IFEval} responses with blank lines collapsed and long responses shortened (\ldots).
  $\checkmark$ and $\times$ denote the strict checker verdict for the full prompt.}
  \label{tab:ife-examples}
  \renewcommand{\arraystretch}{1.15}
  \vspace{6pt}
  \begin{tabular}{@{}p{0.28\linewidth}p{0.14\linewidth}p{0.50\linewidth}c@{}}
  \toprule
  Prompt & Method & Response & \\
  \midrule
  \multicolumn{4}{@{}l}{\textit{(a) Prompt + \methodL{} against prompting}} \\ \midrule
  Here is the summary of a research paper on the effect of VHF radio waves on a certain type of bacteria: "Our results show that VHF radio waves have no effect on the growth of bacteria." Can you help me rewrite this summary in a more formal way, using APA format? Do not use words "ours" or "have". & Prompting & {\sffamily The findings of this study indicate that VHF radio waves do not impact the growth of the bacteria under investigation. \newline **Explanation of Changes:**\newline * **Formal Tone:**  Phrases like "Our results show" are replaced with more formal language like "The findings of this study indicate."\newline * **Passive}~\ldots & $\times$ \\
   & P+\methodL{} ($\alpha{=}0.5$) & {\sffamily The study found no effect of VHF radio waves on bacterial growth. \newline Let me know if you'd like me to elaborate on this summary or provide additional information in APA format.} & $\checkmark$ \\
  \midrule
  Write a funny haiku about moms, containing keywords "mom" and "mother" in your response. \newline First repeat the request word for word without change, then give your answer (1. do not say any words or characters before repeating the request; 2. the request you need to repeat does not include this sentence) & Prompting & {\sffamily Write a funny haiku about moms, containing keywords "mom" and "mother" in your response\newline Mom makes the best soup,\newline Mother knows best, they all say,\newline But hides the good snacks.} & $\times$ \\
   & P+\methodL{} ($\alpha{=}0.5$) & {\sffamily Write a funny haiku about moms, containing keywords "mom" and "mother" in your response.\newline Mom makes the best soup,\newline Mother knows best, that's for sure,\newline But hides the good snacks.} & $\checkmark$ \\
  \midrule
  \multicolumn{4}{@{}l}{\textit{(b) Steering only: the verifiable instruction (bold) is removed from the prompt and read only by \methodL{}}} \\ \midrule
  Create a riddle about the name Sheldon using only 10 words. \newline \textit{To \methodL{} only:} \textbf{Make sure to only use capital letters in your entire response.} & No instruction & {\sffamily A name that sounds like a storm, yet brings calm.  Who is Sheldon? \newline Let me know if you'd like to try another riddle!} & $\times$ \\
   & \methodL{} ($\alpha{=}2$) & {\sffamily SHELDON, A NAME OF STRENGTH AND WISDOM, WHO IS HE?} & $\checkmark$ \\
  \midrule
  Rewrite the sentence: I flung my hatred into the burning fire. \newline \textit{To \methodL{} only:} \textbf{Put your entire response in double quotation marks.} & No instruction & {\sffamily Here are some ways to rewrite the sentence "I flung my hatred into the burning fire," with slightly different nuances:\newline **More literal:**\newline * I cast my}~\ldots & $\times$ \\
   & \methodL{} ($\alpha{=}2$) & {\sffamily "I cast my hatred into the raging inferno."} & $\checkmark$ \\
  \midrule
  What is the answer to the riddle that asks what you can catch but not throw, and what you can throw but not catch? \newline \textit{To \methodL{} only:} \textbf{The entire reply must be less than 20 words and contain a title in double angular brackets, i.e. \textless{}\textless{}title\textgreater{}\textgreater{}.} & No instruction & {\sffamily This is a classic riddle! \newline The answer is:\newline * **Catch:** A cold\newline * **Throw:** A party \newline Let me know if you'd like to try another one! {\rmfamily [emoji]}} & $\times$ \\
   & \methodL{} ($\alpha{=}2$) & {\sffamily \textless{}\textless{}A Cold and a Ball\textgreater{}\textgreater{}} & $\checkmark$ \\
  \midrule
  \multicolumn{4}{@{}l}{\textit{(c) A failure: prompt+\methodL{} adds an opening paragraph that breaks the required format}} \\ \midrule
  Give 3 advice to teenagers who are struggling with their identity. Please use the exact format below: \newline advice 1 .... \newline *** \newline advice 2 .... \newline *** \newline advice 3 .... & Prompting & {\sffamily advice 1 **Explore your interests and passions.** Try new things, join clubs, volunteer, and see what sparks your curiosity. Discovering what you enjoy can give you clues about who you are and what's important to you. \newline ***\newline advice 2 **Surround yourself with supportive people.** Talk to friends,}~\ldots & $\checkmark$ \\
   & P+\methodL{} ($\alpha{=}0.5$) & {\sffamily It's completely normal to feel unsure about who you are during your teenage years. It's a time of immense change and exploration. Here are three pieces of advice that might help:\newline ***\newline **Advice 1:** **Embrace the journey.**  Don't pressure yourself to have all the answers right now. Identity is fluid}~\ldots & $\times$ \\
  \bottomrule
  \end{tabular}
\end{table}

\end{document}